\documentclass[10pt,twocolumn,letterpaper]{article}

\usepackage[utf8]{inputenc}
\usepackage{cvpr}
\usepackage{times}
\usepackage{epsfig}
\usepackage{graphicx}
\usepackage{float}
\usepackage{amsmath}
\usepackage{amssymb}
\usepackage{booktabs}
\usepackage{caption}
\usepackage{subcaption}
\usepackage{multirow}
\usepackage[pagebackref=true,breaklinks=true,letterpaper=true,colorlinks,bookmarks=false]{hyperref}

\newcommand{\bc}{\mathbf{c}}

\newcommand{\bh}{\mathbf{h}}
\newcommand{\bI}{\mathbf{I}}

\newcommand{\bR}{\mathbf{R}}

\newcommand{\bt}{\mathbf{t}}

\newcommand{\bx}{\mathbf{x}}
\newcommand{\by}{\mathbf{y}}

\newcommand{\cH}{\mathcal{H}}

\newcommand{\cP}{\mathcal{P}}

\newcommand{\cT}{\mathcal{T}}

\newcommand{\cX}{\mathcal{X}}
\newcommand{\cY}{\mathcal{Y}}

\newcommand{\figref}[1]{Fig.~\ref{#1}}
\newcommand{\secref}[1]{Section~\ref{#1}}

\newcommand{\tabref}[1]{Table~\ref{#1}}

\makeatletter
\DeclareRobustCommand\onedot{\futurelet\@let@token\@onedot}
\def\@onedot{\ifx\@let@token.\else.\null\fi\xspace}
\def\eg{e.g\onedot} 
\def\ie{i.e\onedot}

\def\wrt{wrt\onedot}

\def\etal{et~al\onedot}

\makeatother

\newcommand{\boldparagraph}[1]{\vspace{0.2cm}\noindent{\bf #1:} }

\usepackage{xcolor}
\definecolor{darkgreen}{rgb}{0,0.7,0}
\definecolor{orange}{RGB}{255,127,0}

\cvprfinalcopy 
\ifcvprfinal\fi

\begin{document}

\title{Learning Unsupervised Hierarchical Part Decomposition of 3D Objects\\ from a Single RGB Image}

\author{Despoina Paschalidou$^{1,3,5}$ \quad Luc van Gool$^{3,4,5}$ \quad Andreas Geiger$^{1,2,5}$\\
$^1$Max Planck Institute for Intelligent Systems T{\"u}bingen\\
$^2$University of T{\"u}bingen\quad
$^3$Computer Vision Lab, ETH Z{\"u}rich\quad
$^4$KU Leuven \\
$^5$Max Planck ETH Center for Learning Systems\\
{\tt\small \{firstname.lastname\}@tue.mpg.de \quad vangool@vision.ee.ethz.ch}}

\maketitle
\begin{abstract}
Humans perceive the 3D world as a set of distinct objects that are characterized by various low-level (geometry, reflectance) and high-level (connectivity, adjacency, symmetry) properties. Recent methods based on convolutional neural networks (CNNs) demonstrated impressive progress in 3D reconstruction, even when using a single 2D image as input. However, the majority of these methods focuses on recovering the local 3D geometry of an object without considering its part-based decomposition or relations between parts. We address this challenging problem by proposing a novel formulation that allows to jointly recover the geometry of a 3D object as a set of primitives as well as their latent hierarchical structure without part-level supervision. Our model recovers the higher level structural decomposition of various objects in the form of a binary tree of primitives, where simple parts are represented with fewer primitives and more complex parts are modeled with more components. Our experiments on the ShapeNet and D-FAUST datasets demonstrate that considering the organization of parts indeed facilitates reasoning about 3D geometry.
\end{abstract}

\section{Introduction} \label{sec:intro}

\begin{figure}[t!]
    \centering
    \includegraphics[width=1.0\linewidth]{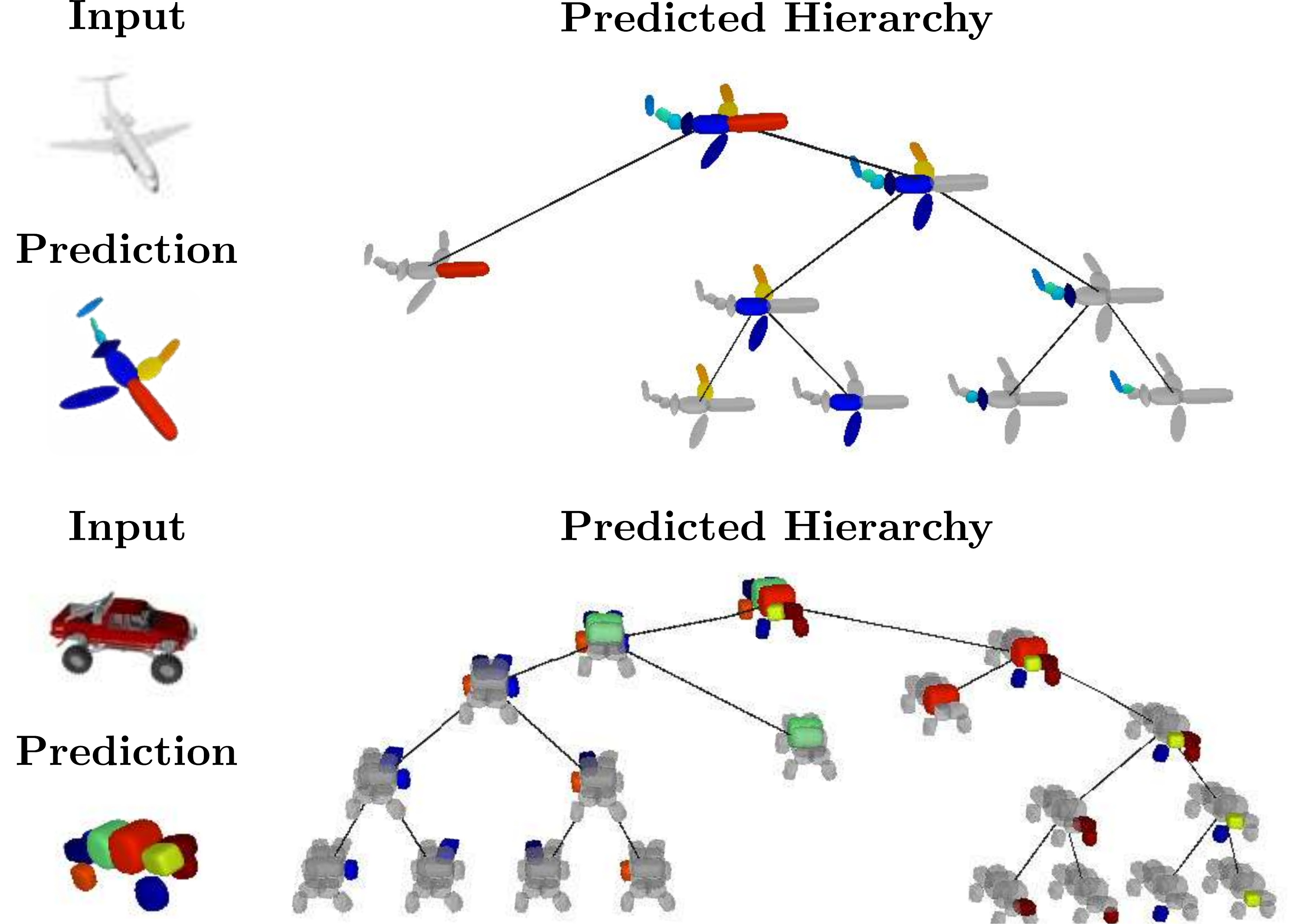}
    \caption{\textbf{Hierarchical Part Decomposition.} We consider the problem of learning structure-aware representations that go beyond part-level geometry and focus on part-level relationships. Here, we show our reconstruction as an unbalanced binary tree of primitives, given a single RGB image as input. Note that our model does not require any supervision on object parts or the hierarchical structure of the 3D object. We show that our representation is able to model different parts of an object with different levels of abstraction, leading to improved reconstruction quality.}
    \label{fig:teaser}
    \vspace{-0.8em}
\end{figure}

Within the first year of their life, humans develop a common-sense understanding of the physical behavior of the world \cite{Baillargeon2004}. This understanding relies heavily on the ability to properly reason about the arrangement of objects in a scene. Early works in cognitive science \cite{Hoffman1984Cognition, Biederman1986Human, Kinzler2007ProgressInBrainResearch} stipulate that the human visual system perceives objects as a hierarchical decomposition of parts. Interestingly, while this seems to be a fairly easy task for the human brain,
computer vision algorithms struggle to form such a high-level reasoning, particularly in the absence of supervision.

\begin{figure}
    \centering
    \includegraphics[width=1.0\linewidth]{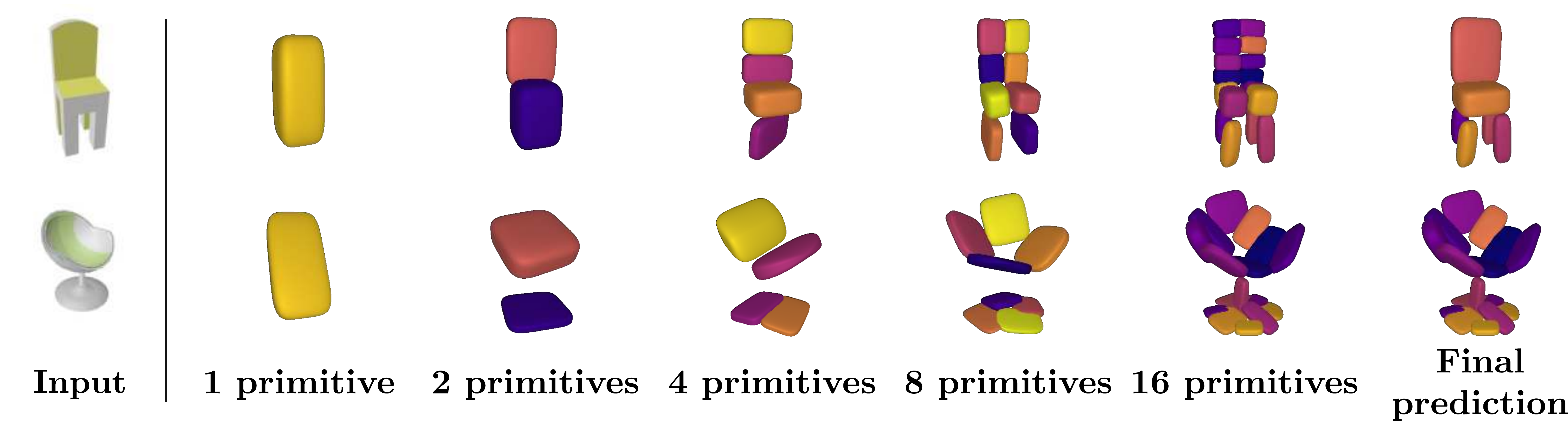}
        \caption{{\bf{Level of Detail.}} Our network represents an object as a tree of primitives. At each depth level $d$, the target object is reconstructed with $2^d$ primitives, This results in a representation with various levels of detail. Naturally, reconstructions from deeper depth levels are more detailed. We associate each primitive with a unique color, thus primitives illustrated with the same color correspond to the same object part. Note that the above reconstructions are derived from the \textit{same} model, trained with a maximum number of $2^4=16$ primitives. During inference, the network dynamically combines representations from different depth levels to recover the final prediction (last column).}
    \label{fig:levels_of_detail}
    \vspace{-0.8em}
\end{figure}

The structure of a scene is tightly related to the inherent hierarchical organization of its parts. At a coarse level, a scene can be decomposed into objects and at a finer level each object can be represented with parts and these parts with finer parts. Structure-aware representations go beyond part-level geometry and focus on global relationships between objects and object parts. In this work, we propose a structure-aware representation that considers part relationships (\figref{fig:teaser}) and models object parts with multiple levels of abstraction, namely geometrically complex parts are modeled with more components and simple parts are modeled with fewer components. Such a multi-scale representation can be efficiently stored at the required level of detail, namely with less parameters (\figref{fig:levels_of_detail}).

Recent breakthroughs in deep learning led to impressive progress in 3D shape extraction by learning a parametric function, implemented as a neural network, that maps an input image to a 3D shape represented as a mesh \cite{Liao2018CVPR, Groueix2018CVPR, Kanazawa2018ECCV, Wang2018ECCV, Yang2019ICCV, Pan2019ICCV}, a pointcloud \cite{Fan2017CVPR, Qi2017NIPS, Achlioptas2018ICML, Jiang2018ECCV, Thomas2019ICCV, Yang2019ICCV}, a voxel grid \cite{Brock2016ARXIV, Choy2016ECCV, Gadelha2017THREEDV, Rezende2016NIPS, Riegler2017CVPR, Stutz2018CVPR, Xie2019ICCV}, 2.5D depth maps \cite{Kar2017NIPS, Hartmann2017ICCV, Paschalidou2018CVPR, Donne2019CVPR} or an implicit surface \cite{Mescheder2019CVPR, Chen2019CVPR, Park2019CVPR, Saito2019ICCV, Xu2019NIPS, Michalkiewicz2019ICCV}. These approaches are mainly focused on reconstructing the geometry of an object, without taking into consideration its constituent parts. This results in non-interpretable reconstructions.
To address the lack of interpretability, researchers shifted their attention to representations that employ shape primitives \cite{Tulsiani2017CVPRa, Paschalidou2019CVPR, Li2019CVPR, Deprelle2019NIPS}.
While these methods yield meaningful semantic shape abstractions, part relationships do not explicitly manifest in their representations.

Instead of representing 3D objects as an unstructured collection of parts, we propose a novel neural network architecture that recovers the latent hierarchical layout of an object without structure supervision.
In particular, we employ a neural network that learns to recursively partition an object into its constituent parts by building a latent space that encodes both the part-level hierarchy and the part geometries. The predicted hierarchical decomposition is represented as an unbalanced binary tree of primitives.
More importantly, this is learned without any supervision neither on the object parts nor their structure. Instead, our model jointly infers these latent variables \textit{during training}.

In summary, we make the following {\bf{contributions}}:
We jointly learn to predict part relationships and per-part geometry without any part-level supervision.
The only supervision required for training our model is a watertight mesh of the 3D object.
Our structure-aware representation yields semantic shape reconstructions that compare favorably to the state-of-the-art 3D reconstruction approach of \cite{Mescheder2019CVPR}, using significantly less parameters and without any additional post-processing. 
Moreover, our learned hierarchies have a semantic interpretation, as the same node in the learned tree is consistently used for representing the same object part.
Experiments on the ShapeNet \cite{Chang2015ARXIV} and the Dynamic FAUST (D-FAUST) dataset \cite{Bogo2017CVPR} demonstrate the ability of our model to parse objects into structure-aware representations that are more expressive and geometrically accurate compared to approaches that only consider the 3D geometry of the object parts \cite{Tulsiani2017CVPRa, Paschalidou2019CVPR, Genova2019ICCV, Deng2019ARXIV}. Code and data is publicly available\footnote{\url{https://github.com/paschalidoud/hierarchical_primitives}}.

\section{Related Work} \label{sec:related}
\begin{figure*}
    \centering
    \includegraphics[width=1.0\textwidth]{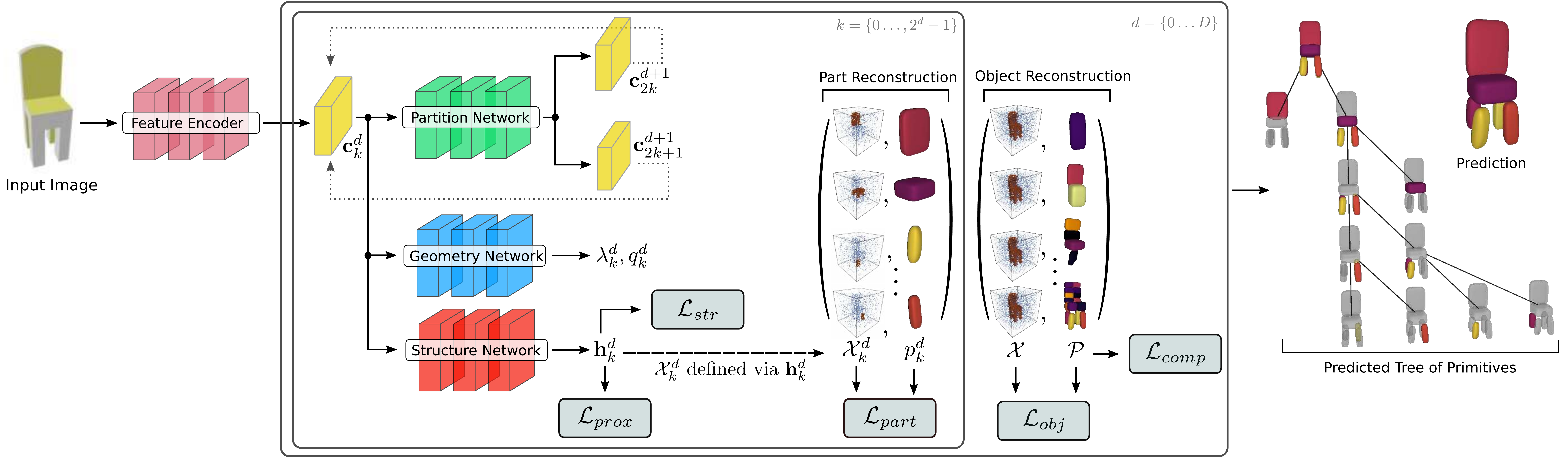}
    \caption{{\bf{Overview.}} Given an input $\bI$ (\eg, image, voxel grid), our network predicts a binary tree of primitives $\cP$ of maximum depth $D$. The \emph{feature encoder} maps the input $\bI$ into a feature vector $\bc_0^0$. Subsequently, the \emph{partition network} splits each feature representation $\bc_k^d$ in two $\{\bc_{2k}^{d+1}, \bc_{2k+1}^{d+1}\}$, resulting in feature representations for $\{1,2,4,\dots, 2^d\}$ primitives where $\bc_k^d$ denotes the feature representation for the $k$-th primitive at depth $d$. Each $\bc_k^d$ is passed to the \emph{structure network} that "assigns" a part of the object to a specific primitive $p_k^d$. As a result, each $p_k^d$ is responsible for representing a specific part of the target shape, denoted as the set of points $\cX_k^d$. Finally, the \emph{geometry network} predicts the primitive parameters $\lambda_k^d$ and the reconstruction quality $q_k^d$ for each primitive.
    To compute the reconstruction loss, we measure how well the predicted primitives match the target object (\emph{Object Reconstruction}) and the assigned parts (\emph{Part Reconstruction}).
            We use plate notation to denote repetition over all nodes $k$ at each depth level $d$.
        The final reconstruction is shown on the right.}
    \label{fig:network_architecture}
\end{figure*}

We now discuss the most related primitive-based and structure-aware shape representations.

\boldparagraph{Supervised Structure-Aware Representations}Our work is related to methods that learn structure-aware shape representations that go beyond mere enumeration of object's parts and recover the higher level structural decomposition of objects based on part-level relations \cite{Mitra2014SIGGRAPH}.
Li \etal \cite{Li2017SIGGRAPH} represent 3D shapes using a symmetry hierarchy \cite{Wang2011EUROGRAPHICS} and train a recursive neural network to predict its hierarchical structure. Their network learns a hierarchical organization of bounding boxes and then fills them with voxelized parts. Note that, this model considers supervision in terms of segmentation of objects into their primitive parts. Closely related to \cite{Li2017SIGGRAPH} is StructureNet \cite{Mo2019SIGGRAPH} which leverages a graph neural network to represent shapes as n-ary graphs. StructureNet considers supervision both in terms of the primitive parameters and the hierarchies. Likewise, Hu \etal \cite{Hu2019ARXIV} propose a supervised model that recovers the 3D structure of a cable-stayed bridge as a binary parsing tree. In contrast our model is unsupervised, \ie, it does not require supervision neither on the primitive parts nor the part relations. 

\boldparagraph{Physics-Based Structure-Aware Representations}The task of inferring higher-level relationships among parts has also been investigated in different settings. Xu \etal \cite{Xu2019ICLR} recover the object parts, their hierarchical structure and each part's dynamics by observing how objects are expected to move in the future. In particular, each part inherits the motion of its parent and the hierarchy emerges by minimizing the norm of these local displacement vectors. Kipf \etal \cite{Kipf2018ICML} explore the use of variational autoencoders for learning the underlying interaction among various moving particles.
Steenkiste \etal \cite{Steenkiste2018ICLR} extend the work of \cite{Greff2017NIPS} on perceptual grouping of pixels and learn an interaction function that models whether objects interact with each other at multiple frames. For both \cite{Kipf2018ICML, Steenkiste2018ICLR}, the hierarchical structure emerges from interactions at
multiple timestamps.  In contrast to \cite{Xu2019ICLR, Kipf2018ICML, Steenkiste2018ICLR}, our model does not relate hierarchies to motion, thus we do not require multiple frames for discovering the hierarchical structure.

\boldparagraph{Supervised Primitive-Based Representations}
Zou \etal \cite{Zou2017ICCV} exploit LSTMs in combination with a Mixture Density Network (MDN) to learn a cuboid representation from depth maps. Similarly, Niu \etal \cite{Niu2018CVPR} employ an RNN that iteratively predicts cuboid primitives as well as their symmetry and connectivity relationships from RGB images. More recently, Li \etal \cite{Li2019CVPR} utilize PointNet++ \cite{Qi2017NIPS} for predicting per-point properties that are subsequently used for estimating the primitive parameters, by solving a series of linear least-squares problems. In contrast to \cite{Zou2017ICCV, Niu2018CVPR, Qi2017NIPS}, which require supervision in terms of the primitive parameters, our model is learned in an unsupervised fashion. In addition, modelling primitives with superquadrics, allows us to exploit a larger shape vocabulary that is not limited to cubes as in \cite{Zou2017ICCV, Niu2018CVPR} or spheres, cones, cylinders and planes as in \cite{Li2019CVPR}.
Another line of work, complementary to ours, incorporates the principles of constructive solid geometry (CSG) \cite{Laidlaw1986SIGGRAPH} in a learning framework for shape modeling \cite{Sharma2018CVPR, Ellis2018NIPS, Tian2019ICLR, Liu2019ICLR}. These works require rich annotations for the primitive parameters and the sequence of predictions.

\boldparagraph{Unsupervised Shape Abstraction}Closely related to our model are the works of \cite{Tulsiani2017CVPRa, Paschalidou2019CVPR} that employ a convolutional neural network (CNN) to regress the parameters of the primitives that best describe the target object, in an unsupervised manner. Primitives can be cuboids \cite{Tulsiani2017CVPRa} or superquadrics \cite{Paschalidou2019CVPR} and are learned by minimizing the discrepancy between the target and the predicted shape, by either computing the truncated bi-directional distance \cite{Tulsiani2017CVPRa} or the Chamfer-distance between points on the target and the predicted shape \cite{Paschalidou2019CVPR}.
While these methods learn a flat arrangement of parts, our structure-aware representation decomposes the depicted object into a hierarchical layout of semantic parts. This results in part geometries with different levels of granularity. Our model differs from \cite{Tulsiani2017CVPRa, Paschalidou2019CVPR} also \wrt the optimization objective. We empirically observe that for both \cite{Tulsiani2017CVPRa, Paschalidou2019CVPR}, the proposed loss formulations suffer from various local minima that stem from the nature of their optimization objective. To mitigate this, we use the more robust classification loss proposed in \cite{Mescheder2019CVPR, Chen2019CVPR, Park2019CVPR} and train our network by learning to classify whether points lie inside or outside the target object.
Very recently, \cite{Genova2019ICCV, Deng2019ARXIV} explored such a loss for recovering shape elements from 3D objects. Genova \etal \cite{Genova2019ICCV} leverage a CNN to learn to predict the parameters of a set of axis-aligned 3D Gaussians from a set of depth maps rendered at different viewpoints. Similarly, Deng \etal \cite{Deng2019ARXIV} employ an autoencoder to recover the geometry of an object as a collection of smooth convexes. In contrast to \cite{Genova2019ICCV, Deng2019ARXIV}, our model goes beyond the local geometry of parts and attempts to recover the underlying hierarchical structure of the object parts.

\section{Method} \label{sec:method}

In this section, we describe our novel neural network architecture for inferring structure-aware representations. Given an input $\bI$ (\eg, RGB image, voxel grid) our goal is to learn a neural network $\phi_{\theta}$, which maps the input to a set of primitives that best describe the target object.
The target object is represented as a set of pairs $\cX = \{\left(\bx_i, o_i\right)\}_{i=1}^N$, where $\bx_i$ corresponds to the location of the $i$-th point and $o_i$ denotes its label, namely whether $\bx_i$ lies inside $(o_i=1)$ or outside $(o_i=0)$ the target object. We acquire these $N$ pairs by sampling points inside the bounding box of the target mesh and determine their labels using a watertight mesh of the target object.
During training, our network learns to predict shapes that contain all internal points from the target mesh $(o_i=1)$ and none of the external $(o_i=0)$.
We discuss our sampling strategy in our supplementary.

Instead of predicting an unstructured set of primitives, we recover a hierarchical decomposition over parts in the form of a \emph{binary tree} of maximum depth $D$ as
\begin{equation}
    \cP=\{\{p_k^d\}_{k=0}^{2^d-1} \,\mid \, d=\{0 \dots D\}\}
\end{equation}
where $p_k^d$ is the $k$-th primitive at depth $d$.
Note that for the $k$-th node at depth $d$, its parent is defined as $p_{\lfloor \frac{k}{2}\rfloor}^{d-1}$ and its two children as $p_{2k}^{d+1}$ and $p_{2k+1}^{d+1}$.

At every depth level, $\cP$ reconstructs the target object with $\{1,2,\dots, $M$\}$ primitives. $M$ is an upper limit to the maximum number of primitives and is equal to $2^D$.
More specifically, $\cP$ is constructed as follows: the root node is associated with the root primitive that represents the entire shape and
is recursively split into two nodes (its children) until reaching the maximum depth $D$.
This recursive partition yields reconstructions that recover the geometry of the target shape using $2^d$ primitives, where $d$ denotes the depth level (see \figref{fig:levels_of_detail}).
Throughout this paper, the term \emph{node} is used interchangeably with \emph{primitive} and always refers to the primitive associated with this particular node.

Every primitive is fully described by a set of parameters $\lambda_k^d$ that define its shape, size and position in 3D space.
Since not all objects require the same number of primitives, we enable our model to predict unbalanced trees, \ie stop recursive partitioning if the reconstruction quality is sufficient. To achieve this our network also regresses a \emph{reconstruction quality} for each primitive denoted as $q_k^d$. Based on the value of each $q_k^d$ the network dynamically stops the recursive partitioning process resulting in parsimonious representations as illustrated in \figref{fig:teaser}.

\subsection{Network Architecture}
\label{sec:architecture}

Our network comprises three main components: (i) the \emph{partition network} that recursively splits the shape representation into representations of parts, (ii) the \emph{structure network} that focuses on learning the hierarchical arrangement of primitives, namely assigning parts of the object to the primitives at each depth level and (iii) the \emph{geometry network} that recovers the primitive parameters. An overview of the proposed pipeline is illustrated in \figref{fig:network_architecture}.
The first part of our pipeline is a \emph{feature encoder}, implemented with a ResNet-18 \cite{He2016CVPR}, ignoring the final fully connected layer. Instead, we only keep the feature vector of length $F=512$ after average pooling.

\boldparagraph{Partition Network}The feature encoder maps the input $\bI$ to an intermediate feature representation $\bc_0^0 \in \mathbb{R}^F$ that describes the root node $p_0^0$.
The partition network implements a function $p_{\theta}: \mathbb{R}^F \to \mathbb{R}^{2F}$ that recursively partitions the feature representation $\bc_k^d$ of node $p_k^d$ into two feature representations, one for each children  $\{p_{2k}^{d+1}, p_{2k+1}^{d+1}\}$:
\begin{equation}
    p_{\theta}(\bc_k^d) = \{\bc_{2k}^{d+1}, \bc_{2k+1}^{d+1}\}.
    \label{eq:partitioner}
\end{equation}
Each primitive $p_k^d$ is directly predicted from $\bc_k^d$ without considering the other intermediate features. This implies that the necessary information for predicting the primitive parameterization is entirely encapsulated in $\bc_k^d$ and not in any other intermediate feature representation.

\boldparagraph{Structure Network}Due to the lack of ground-truth supervision in terms of the tree structure, we introduce the structure network that seeks to learn a pseudo-ground truth part-decomposition of the target object.
More formally, it learns a function $s_{\theta}:\mathbb{R}^F \rightarrow \mathbb{R}^3$ that maps each feature representation $\bc_k^d$ to $\bh_k^d$ a spatial location in $\mathbb{R}^3$.

One can think of each $\bh_k^d$ as the (geometric) centroid of a specific part of the target object.
We define
\begin{equation}
\cH=\{\{\bh_k^d\}_{k=0}^{2^d-1} \,| \, d=\{0 \dots D\}\}
\end{equation}
the set of centroids of all parts of the object at all depth levels. From $\cH$ and $\cX$, we are now able to derive the part decomposition of the target object as the set of points $\cX_k^d$ that are internal to a part with centroid $\bh_k^d$.

Note that, in order to learn $\cP$, we need to be able to partition the target object into $2^d$ parts at each depth level.
At the root level $(d=0)$, $\bh_0^0$ is the centroid of the target object and $\cX_0^0$ is equal to $\cX$.
For $d=1$, $\bh_0^1$ and $\bh_1^1$ are the centroids of the two parts representing the target object. $\cX_0^1$ and $\cX_1^1$ comprise the same points as $\cX_0^0$.
For the external points, the labels remain the same.
For the internal points, however, the labels are distributed between $\cX_0^1$ and $\cX_1^1$
based on whether $\bh_0^1$ or $\bh_1^1$ is closer.
That is, $\cX_0^1$ and $\cX_1^1$ each contain more external labels and less internal labels compared to $\cX_0^0$.
The same process is repeated until we reach the maximum depth.

\begin{figure}
    \centering
    \begin{subfigure}[b]{\linewidth}
    \centering
    \includegraphics[width=1.0\linewidth]{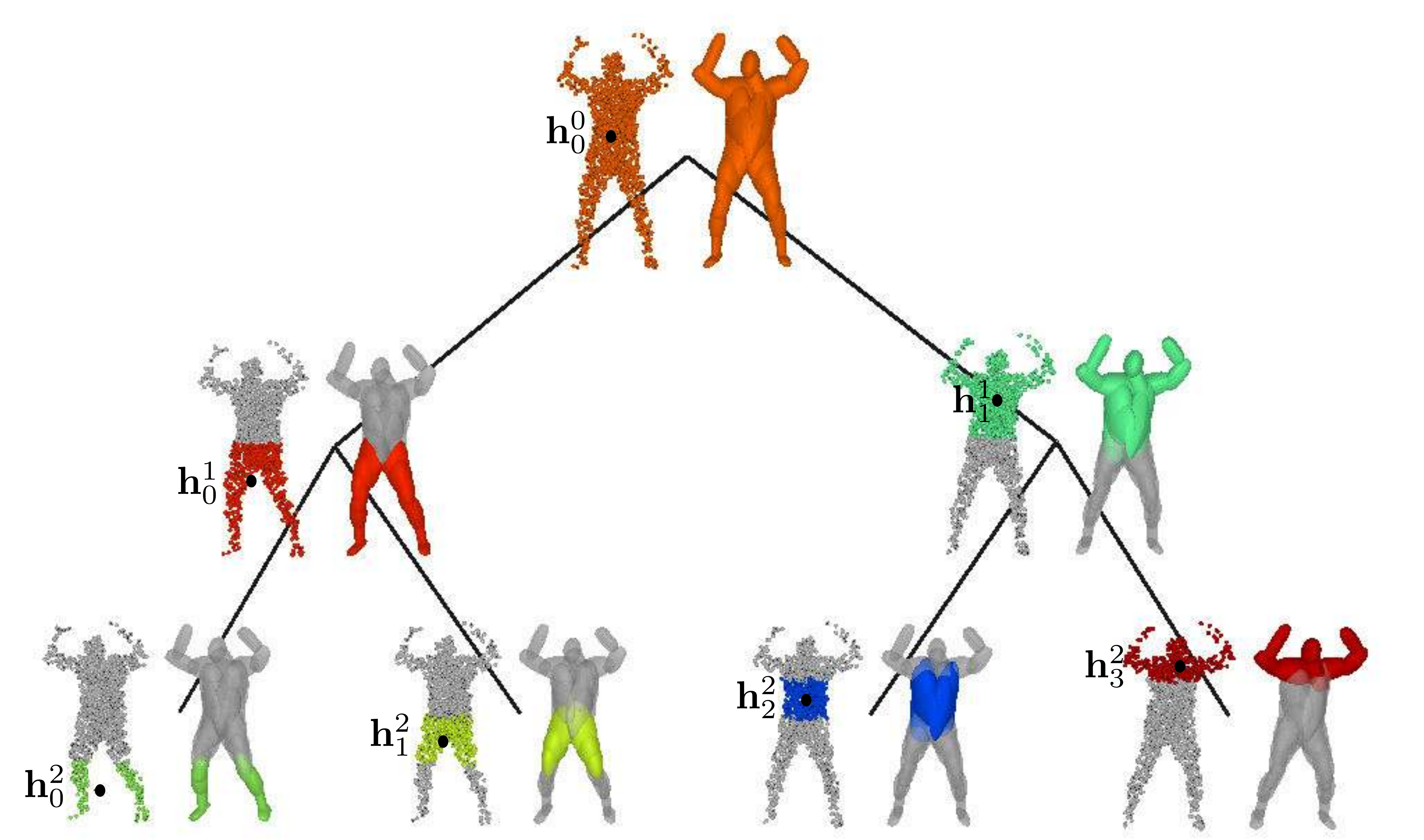}
    \vspace{-2.2em}
    \caption{}
    \label{fig:part_hkd_association}
    \end{subfigure}
    \begin{subfigure}[b]{\linewidth}
    \centering
    \includegraphics[width=1.0\linewidth]{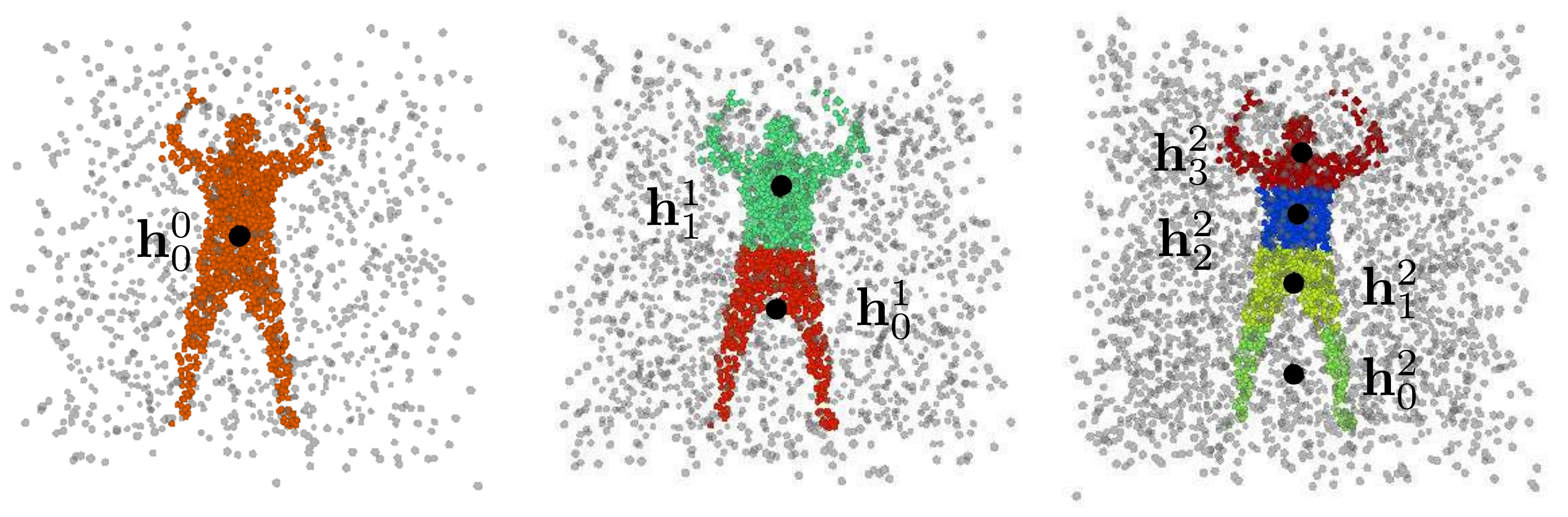}
    \vspace{-2.2em}
    \caption{}
    \label{fig:labels_hkd_association}
    \end{subfigure}
    \caption{{\bf{Structure Network.}} We visualize the centroids $\bh_k^d$ and the 3D points $\cX_k^d$ that correspond to the estimated part $p_k^d$ for the first three levels of the tree.
    	  	            \figref{fig:labels_hkd_association} explains visually Eq. \eqref{eq:split}. We color points based on their closest centroid $\bh_k^d$. Points illustrated with the color associated to a part are labeled ``internal'' ($o=1$). Points illustrated with gray are labeled ``external'' ($o=0$).
    }
    \label{fig:structure_network}
    \vspace{-1.2em}
\end{figure}

More formally, we define the set of points $\cX_k^d$ corresponding to primitive $p_k^d$ implicitly via its centroid $\bh_k^d$:
\begin{equation}
    \cX_k^d = \left\{N_k(\bx, o) \quad \forall (\bx, o) \in \cX_{\lfloor \frac{k}{2}\rfloor}^{d-1}\right\}
    \label{eq:split}
\end{equation}
Here, $\cX_{\lfloor \frac{k}{2}\rfloor}^{d-1}$ denotes the points of the parent.
The function $N_k(\bx, o)$ assigns each $(\bx,o) \in \cX_{\lfloor \frac{k}{2}\rfloor}^{d-1}$ to part $p_k^d$ if it is closer to $\bh_k^d$ than to $\bh_{s(k)}^d$ where $s(k)$ is the sibling of $k$:
\begin{equation}
\begin{aligned}
    N_k(\bx, o) &= \begin{cases}
            (\bx, 1) \quad \|\bh_k^d-\bx\| \leq \|\bh_{s(k)}^d-\bx\| \land o = 1\\
            (\bx, 0) \quad \text{otherwise}
        \end{cases}
\end{aligned}
\end{equation}
Intuitively, this process recursively associates points to the closest sibling at each level of the binary tree where the association is determined by the label $o$.
\figref{fig:structure_network} illustrates the part decomposition of the target shape using $\cH$. We visualize each part with a different color.

\boldparagraph{Geometry Network}
The geometry network learns a function $r_{\theta}:\mathbb{R}^F \rightarrow \mathbb{R}^K \times [0, 1]$ that maps the feature representation $\bc_k^d$ to its corresponding primitive parametrization $\lambda_k^d$ and the reconstruction quality prediction $q_k^d$:
\begin{equation}
    r_{\theta}(\bc_k^d) = \{\lambda_k^d, q_k^d\}.
    \label{eq:partitioner}
\end{equation}

\subsection{Primitive Parametrization}
\label{sec:primitives_parametrization}

For primitives, we use superquadric surfaces.
A detailed analysis of the use of superquadrics as geometric primitives is beyond the scope of this paper, thus we refer the reader to \cite{Jaklic2000, Paschalidou2019CVPR} for more details. Below, we focus on the properties most relevant to us.
For any point $\bx \in \mathbb{R}^3$, we can determine whether it lies inside or outside a superquadric using its implicit surface function which is commonly referred to as the \textit{inside-outside function}:
\begin{equation}
    f(\bx; \lambda) = \left(\left(\frac{x}{\alpha_{1}}\right)^{\frac{2}{\epsilon_{2}}}
        + \left(\frac{y}{\alpha_{2}}\right)^{\frac{2}{\epsilon_{2}}}\right)^{\frac{\epsilon_{2}}{\epsilon_{1}}}
        + \left(\frac{z}{\alpha_{3}}\right)^{\frac{2}{\epsilon_{1}}}
    \label{eq:implicit_sq}
\end{equation}
where $\mathbf{\alpha} = [\alpha_{1}, \alpha_{2}, \alpha_{3}]$ determine the size and $\mathbf{\epsilon} = [\epsilon_{1}, \epsilon_{2}]$ the shape of the superquadric.
If $f(\bx; \lambda) = 1.0$, the given point $\bx$ lies on the surface of the superquadric, if $f(\bx; \lambda) < 1.0$ the corresponding point lies inside and if $f(\bx; \lambda) > 1.0$ the point lies outside the superquadric.
To account for numerical instabilities that arise from the exponentiations in \eqref{eq:implicit_sq}, instead of directly using $f(\bx; \lambda)$, we follow \cite{Jaklic2000} and use $f(\bx; \lambda)^{\epsilon_1}$. Finally, we convert the inside-outside function to an \emph{occupancy function}, $g:\mathbb{R}^3 \rightarrow [0, 1]$:
\begin{equation}
    g(\bx; \lambda) = \sigma\left(s\left(1 - f(\bx; \lambda)^{\epsilon_1}\right)\right)
    \label{eq:implicit_sq_actual}
\end{equation}
that results in per-point predictions suitable for the classification problem we want to solve. $\sigma(\cdot)$ is the sigmoid function and $s$ controls the sharpness of the transition of the occupancy function.
To account for any rigid body motion transformations, we augment the primitive parameters with a translation vector $\mathbf{t} = [t_{x}, t_{y}, t_{z}]$ and a quaternion $\mathbf{q} = [q_{0}, q_{1}, q_{2}, q_{3}]$ \cite{Hamilton1848}, which determine the coordinate system transformation $\cT(\bx)=\bR(\lambda)\,\bx + \bt(\lambda)$. Note that in \eqref{eq:implicit_sq}, \eqref{eq:implicit_sq_actual} we omit the primitive indexes $k, d$ for clarity.
Visualizations of \eqref{eq:implicit_sq_actual} are given in our supplementary.

\subsection{Network Losses} \label{sec:losses}

Our optimization objective $\mathcal{L}(\cP, \cH; \cX)$ is a weighted sum over four terms:
\begin{equation}
\begin{aligned}
        \mathcal{L}(\cP, \cH; \cX) =\, &\mathcal{L}_{str}(\cH;\cX) + \mathcal{L}_{rec}(\cP;\cX) \\+ \, &\mathcal{L}_{comp}(\cP;\cX) + \mathcal{L}_{prox}(\cP)
    \label{eq:overall_loss}
\end{aligned}
\end{equation}

\boldparagraph{Structure Loss}
Using $\cH$ and $\cX$, we can decompose the target mesh into a hierarchy of disjoint parts. Namely, each $\bh_k^d$ implicitly defines a set of points $\cX_k^d$ that describe a specific part of the object as described in \eqref{eq:split}.
To quantify how well $\cH$ clusters the input shape $\cX$ we minimize the sum of squared distances, similar to classical k-means:
\begin{equation}
    \mathcal{L}_{str}(\cH; \cX) = \sum_{h_k^d \in \cH} \frac{1}{2^d-1}\sum_{\substack{(\bx, o) \in \cX_k^d}} \, o\,{\Vert\bx - \bh_k^d \Vert}_2
    \label{eq:structure_loss}
\end{equation}
Note that for the loss in \eqref{eq:structure_loss}, we only consider gradients with respect to $\cH$ as $\cX_k^d$ is implicitly defined via $\cH$. This results in a procedure resembling Expectation-Maximization (EM) for clustering point clouds, where computing $\cX_k^d$ is the expectation step and each gradient updated corresponds to the maximization step.
In contrast to EM, however, we minimize this loss across all instances of the training set, leading to parsimonious but consistent shape abstractions.
An example of this clustering process performed at training-time is shown in \figref{fig:structure_network}.

\boldparagraph{Reconstruction Loss}
The reconstruction loss measures how well the predicted primitives match the target shape. Similar to \cite{Genova2019ICCV, Deng2019ARXIV}, we formulate our reconstruction loss as a binary classification problem, where our network learns to predict the surface boundary of the predicted shape by classifying whether points in $\cX$ lie inside or outside the target object.
To do this, we first define the occupancy function of the predicted shape at each depth level.
Using the occupancy function of each primitive defined in \eqref{eq:implicit_sq_actual}, the occupancy function of the overall shape at depth $d$ becomes:
\begin{equation}
    G^d(\bx) = \max_{\substack{k \in 0\dots2^d-1}}g_k^d\left(\bx; \lambda_k^d\right)
    \label{eq:occupancy_predicted}
\end{equation}
Note that \eqref{eq:occupancy_predicted} is simply the union of the per-primitive occupancy functions.
We formulate our reconstruction loss \wrt the object and \wrt each part of the object as follows
\begin{align}
\mathcal{L}_{rec}(\cP;\cX) = &\sum_{\substack{(\bx, o) \in \cX}}\sum_{d=0}^{D} L\left(G^d(\bx), o\right) +
                     \label{eq:object_loss} \\
                     &\sum_{d=0}^{D}\sum_{k=0}^{2^d-1}\sum_{\substack{(\bx, o) \in \cX_k^d}}L\left(g_k^d\left(\bx; \lambda_k^d\right), o\right)
                     \label{eq:part_loss}
\end{align}
where $L(\cdot)$ is the binary cross entropy loss.
The first term is an \emph{object reconstruction loss} \eqref{eq:object_loss} and measures how well the predicted shape at each depth level matches the target shape.
The second term \eqref{eq:part_loss} which we refer to as \emph{part reconstruction loss} measures how accurately each primitive $p_k^d$ matches the part of the object it represents, defined as the point set $\cX_k^d$. Note that the \emph{part reconstruction loss} enforces non-overlapping primitives, as $\cX_k^d$ are non-overlapping by construction. We illustrate our reconstruction loss in \figref{fig:network_architecture}.

\boldparagraph{Compatibility Loss}
This loss measures how well our model is able to predict the expected \emph{reconstruction quality} $q_k^d$ of a primitive $p_k^d$.
A standard metric for measuring the reconstruction quality is the Intersection over Union (IoU). We therefore task our network to predict the \emph{reconstruction quality} of each primitive $p_k^d$ in terms of its IoU \wrt the part of the object $\cX_k^d$ it represents:
\begin{equation}
    \mathcal{L}_{comp}(\cP;\cX) = \sum_{d=0}^{\mathcal{D}}\sum_{k=0}^{2^d-1} \left(q_k^d - \text{IoU}(p_k^d, \cX_k^d)\right)^2
\end{equation}
During inference, $q_k^d$ allows for further partitioning primitives whose IoU is below a threshold $q_{th}$ and to stop if the reconstruction quality is high (the primitive fits the object part well). As a result, our model predicts an unbalanced tree of primitives where objects can be represented with various number of primitives from 1 to $2^D$. This results in parsimonious representations where simple parts are represented with fewer primitives. We empirically observe that the threshold value $q_{th}$ does not significantly affect our results, thus we empirically set it to $0.6$. During training, we do not use the predicted reconstruction quality $q_k^d$ to dynamically partition the nodes but instead predict the full tree.

\begin{figure}
    \begin{subfigure}[b]{.3\linewidth}
	\begin{subfigure}[b]{\linewidth}
		\centering
		\includegraphics[width=0.7\linewidth]{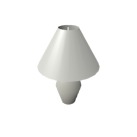}
        \vspace{-1.2em}
        \caption{\bf{Input}}
	\end{subfigure}
    \vskip\baselineskip
	\begin{subfigure}[b]{\linewidth}
		\centering
		\includegraphics[width=0.7\linewidth]{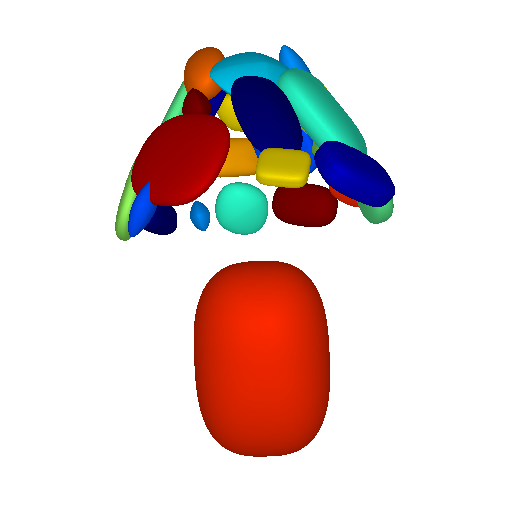}
        \vspace{-0.8em}
        \caption{\bf{Prediction}}
    \end{subfigure}
	\end{subfigure}    \begin{subfigure}[b]{.65\linewidth}
       \centering
	   \includegraphics[width=\linewidth]{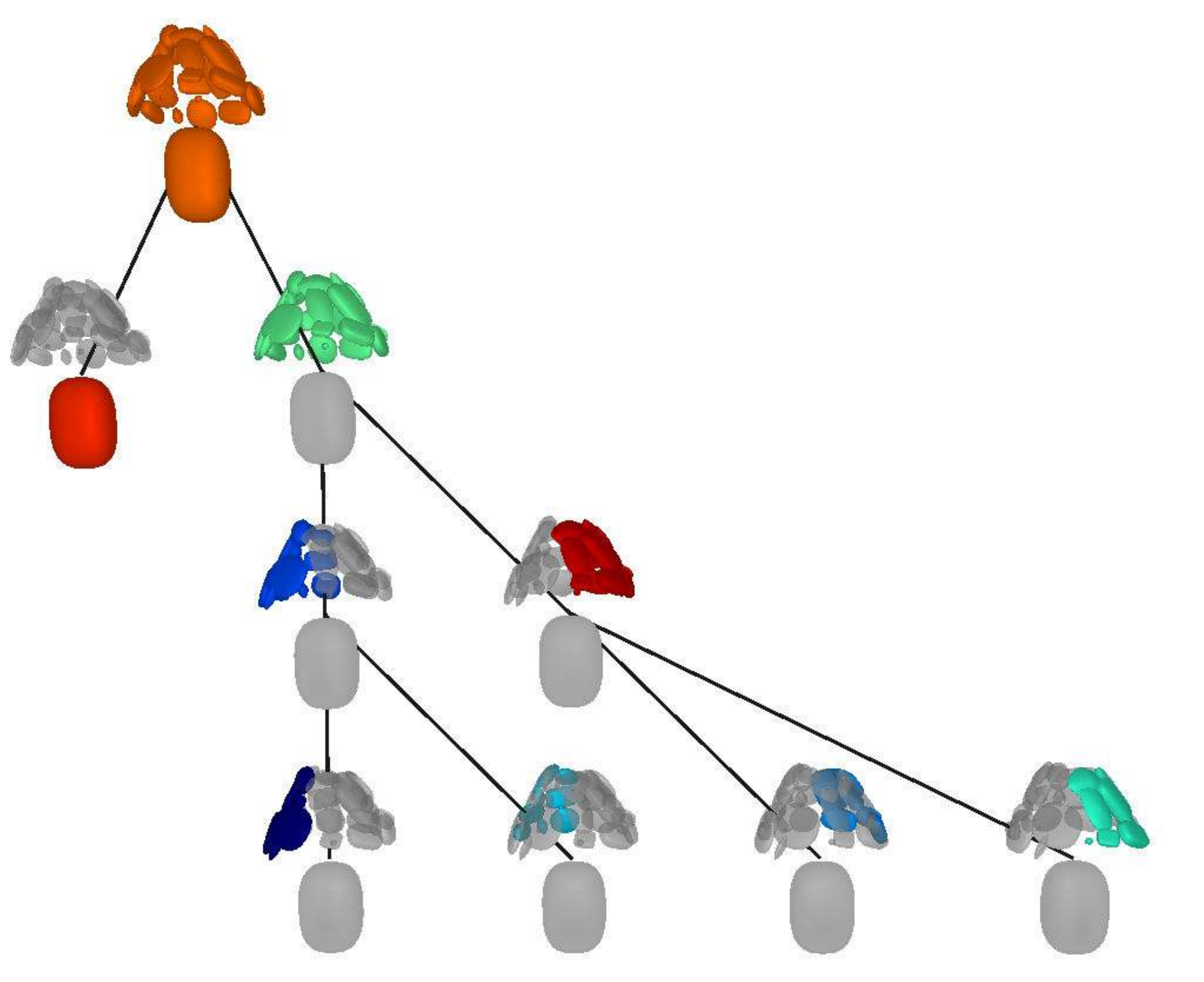}
        \vspace{-0.8em}
       \caption{\bf{Predicted Hierarchy}}
    \vspace{-1.0em}
	\end{subfigure}
    \begin{subfigure}[b]{.3\linewidth}
	\begin{subfigure}[b]{\linewidth}
		\centering
		\includegraphics[width=0.7\linewidth]{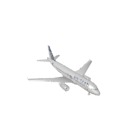}
        \vspace{-1.2em}
        \caption{\bf{Input}}
	\end{subfigure}
    \vskip\baselineskip
	\begin{subfigure}[b]{\linewidth}
		\centering
		\includegraphics[width=0.7\linewidth]{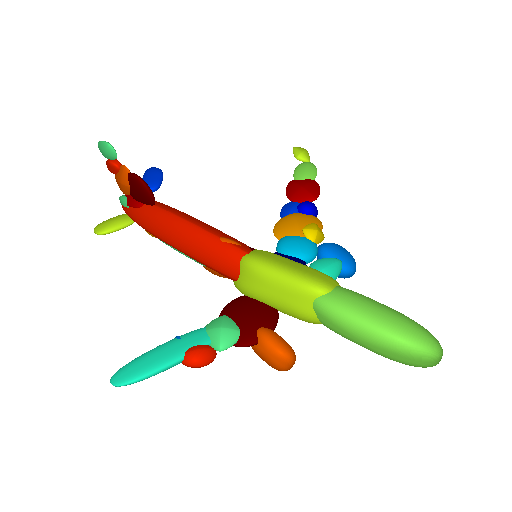}
        \vspace{-0.8em}
        \caption{\bf{Prediction}}
    \end{subfigure}
	\end{subfigure}    \begin{subfigure}[b]{.65\linewidth}
       \centering
	   \includegraphics[width=\linewidth]{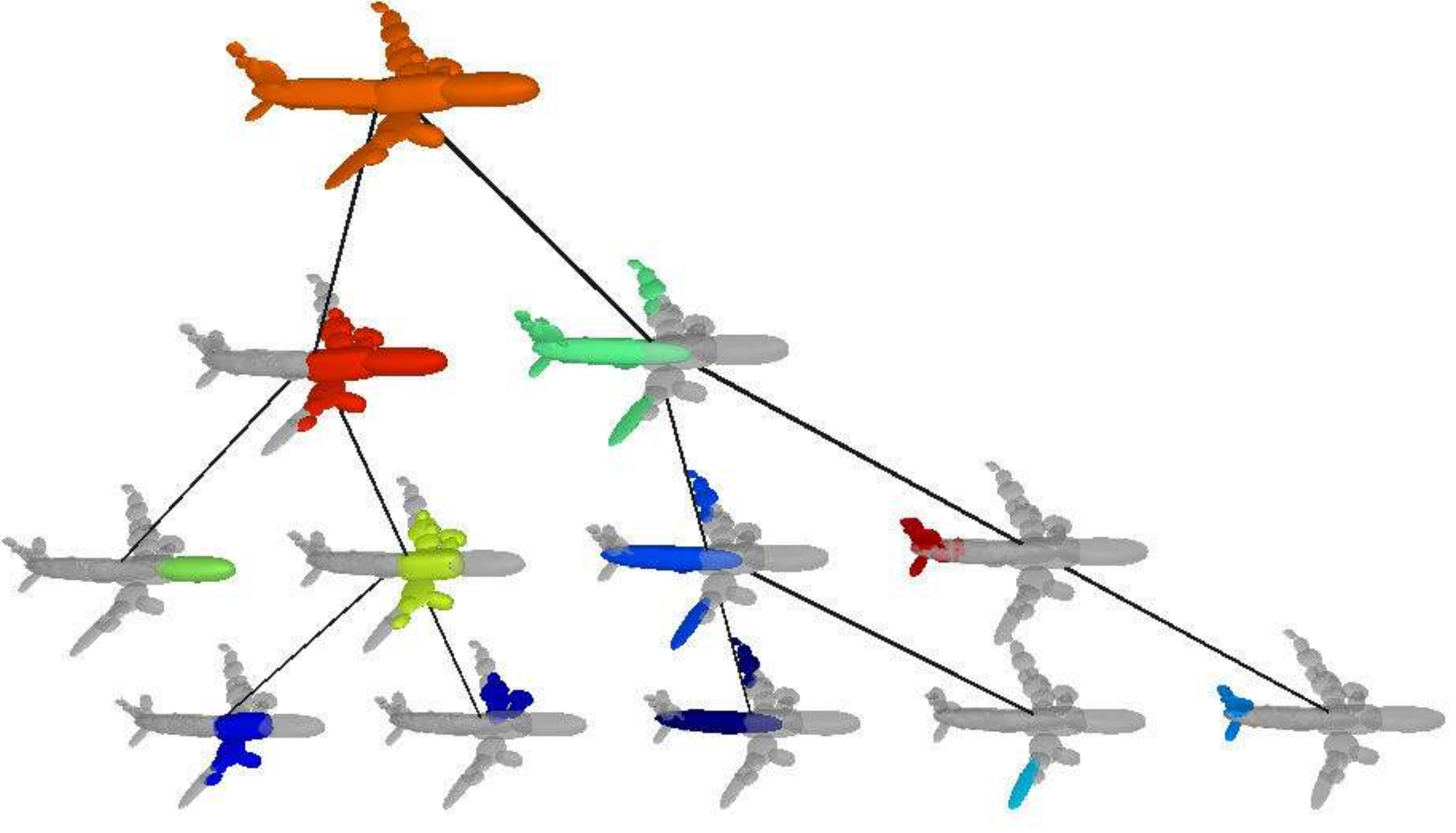}
       \vspace{-0.8em}
       \caption{\bf{Predicted Hierarchy}}
	\end{subfigure}
    \caption{{\bf Predicted Hierarchies on ShapeNet}. Our model recovers the geometry of an object as an unbalanced hierarchy over primitives, where simpler parts (\eg base of the lamp) are represented with few primitives and more complex parts (\eg wings of the plane) with more.}
    \label{fig:shapenet_hiearchy}
    \vspace{-1.1em}
\end{figure}

\boldparagraph{Proximity Loss}
This term is added to counteract vanishing gradients due to the sigmoid in \eqref{eq:implicit_sq_actual}.
For example, if the initial prediction of a primitive is far away from the target object, the reconstruction loss will be large while its gradients will be small. As a result, it is impossible to ``move'' this primitive to the right location. Thus, we introduce a proximity loss which encourages the center of each primitive $p_k^d$ to be close to the centroid of the part it represents:
\begin{equation}
    \mathcal{L}_{prox}(\cP) = \sum_{d=0}^{D}\sum_{k=0}^{2^d-1} \,{\Vert \bt(\lambda_k^d) - \bh_k^d \Vert}_2
\end{equation}
where $\bt(\lambda_k^d)$ is the translation vector of the primitive $p_k^d$ and $\bh_k^d$ is the centroid of the part it represents. We demonstrate the vanishing gradient problem in our supplementary.

{
\def\arraystretch{1.05}
\begin{table*}
\resizebox{\textwidth}{!}{    \centering
    \begin{tabular}{l||c|cccc||c|cccc}
        \toprule
        \multicolumn{1}{c}{\,} & \multicolumn{5}{c}{Chamfer-$L_1$} & \multicolumn{5}{c}{IoU} \\
        &  OccNet \cite{Mescheder2019CVPR} & SQs \cite{Paschalidou2019CVPR} & SIF \cite{Genova2019ICCV} & CvxNets \cite{Deng2019ARXIV} & Ours &
        OccNet \cite{Mescheder2019CVPR} & SQs \cite{Paschalidou2019CVPR} & SIF \cite{Genova2019ICCV} & CvxNets \cite{Deng2019ARXIV} & Ours\\
        Category & & & & & & \\
        \midrule
        airplane & 0.147    & 0.122       & \bf{0.065} & 0.093    & 0.175      & 0.571      & 0.456      & 0.530    & \bf{0.598}  & 0.529 \\
        bench    & 0.155    & \bf{0.114}  & 0.131      & 0.133    & 0.153      & 0.485      & 0.202      & 0.333    & \bf{0.461}  & 0.437 \\
        cabinet  & 0.167    & \bf{0.087}  & 0.102      & 0.160    & \bf{0.087} & 0.733      & 0.110      & 0.648    & \bf{0.709}  & 0.658 \\
        car      & 0.159    & 0.117       & \bf{0.056} & 0.103    & 0.141      & 0.737      & 0.650      & 0.657    & 0.675       & \bf{0.702} \\
        chair    & 0.228    & 0.138       & 0.192      & 0.337    & \bf{0.114} & 0.501      & 0.176      & 0.389    & 0.491       & \bf{0.526} \\
        display  & 0.278    & \bf{0.106}  & 0.208      & 0.223    & 0.137      & 0.471      & 0.200      & 0.491    & 0.576       & \bf{0.633} \\
        lamp     & 0.479    & 0.189       & 0.454      & 0.795    & \bf{0.169} & 0.371      & 0.189      & 0.260    & 0.311       & \bf{0.441} \\
        speaker  & 0.300    & 0.132       & 0.253      & 0.462    & \bf{0.108} & 0.647      & 0.136      & 0.577    & 0.620       & \bf{0.660} \\
        rifle    & 0.141    & 0.127       & \bf{0.069} & 0.106    & 0.203      & 0.474      & \bf{0.519} & 0.463    & 0.515       & 0.435 \\
        sofa     & 0.194    & \bf{0.106}  & 0.146      & 0.164    & 0.128      & 0.680      & 0.122      & 0.606    & 0.677       & \bf{0.693} \\
        table    & 0.189    & \bf{0.110}  & 0.264      & 0.358    & 0.122      & 0.506      & 0.180      & 0.372    & 0.473       & \bf{0.491} \\
        phone    & 0.140    & 0.112       & \bf{0.095} & 0.083    & 0.149      & 0.720      & 0.185      & 0.658    & 0.719       & \bf{0.770} \\
        vessel   & 0.218    & 0.125       & \bf{0.108} & 0.173    & 0.178      & 0.530      & 0.471      & 0.502    & 0.552       & \bf{0.570} \\
        \midrule                                                                                           
        mean     & 0.215    & \bf{0.122}   & 0.165      & 0.245    & 0.143     & 0.571 & 0.277  & 0.499    & 0.567  & \bf{0.580}\\
        \bottomrule
    \end{tabular}
    }
    \caption{{\bf Single Image Reconstruction on ShapeNet.} Quantitative evaluation of our method against OccNet
    \cite{Mescheder2019CVPR} and primitive-based methods with superquadrics
    \cite{Paschalidou2019CVPR} (SQs), SIF
    \cite{Genova2019ICCV} and CvxNets \cite{Deng2019ARXIV}. We report the volumeteric IoU (higher is better) and the
    Chamfer-$L_1$ distance (lower is better) \wrt the ground-truth mesh.
    }
    \label{tab:shapenet_v1_quantitative}
    \vspace{-0.4em}
\end{table*}
}

\section{Experiments} \label{sec:results}
\begin{figure}
    \centering
    \includegraphics[width=1.0\linewidth]{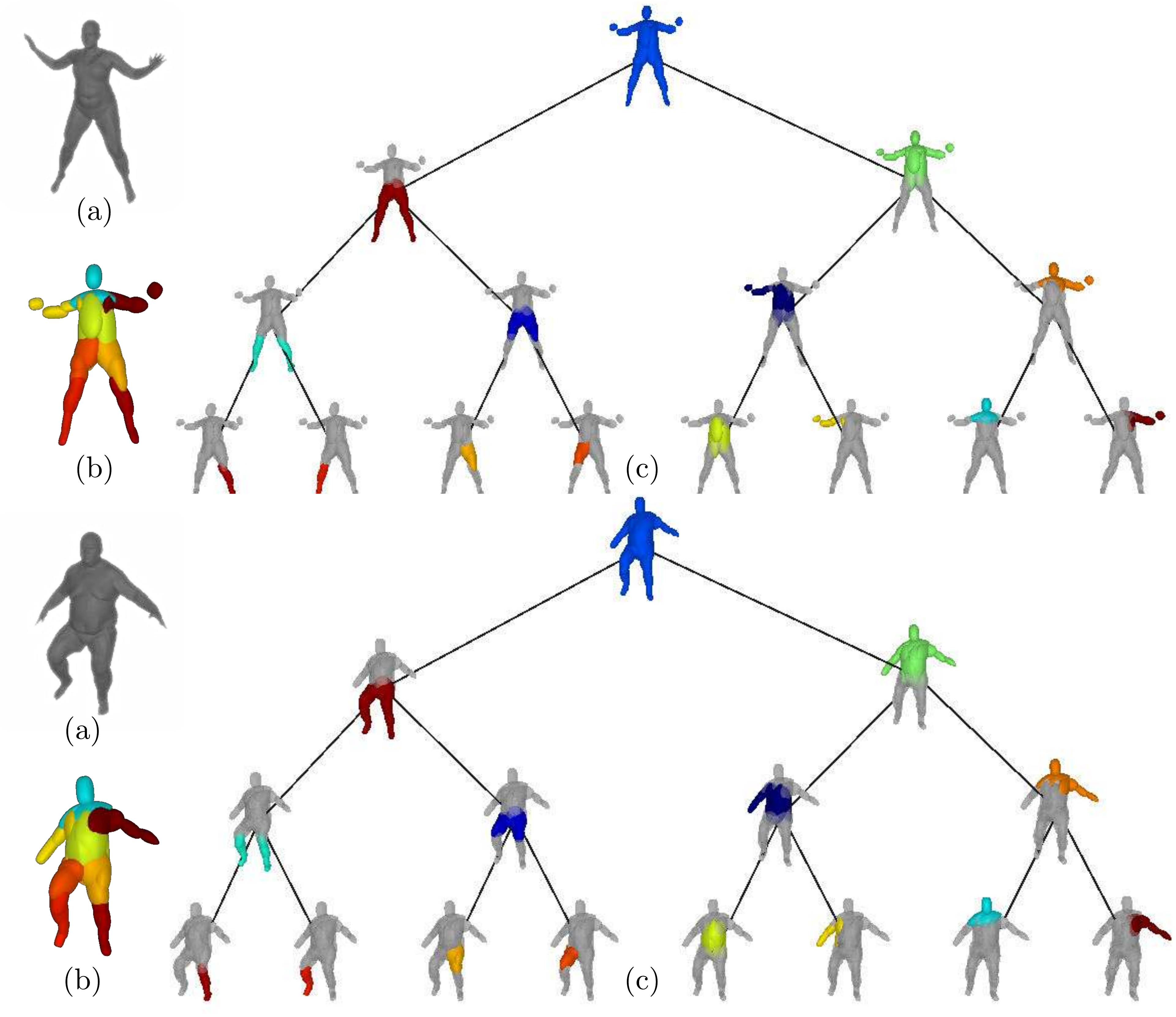}
    \caption{{\bf{Predicted Hierarchies on D-FAUST.}} We visualize the input RGB image (a), the prediction (b) and the predicted hierarchy (c). We associate each primitive with a color and we observe that our network learns semantic mappings of body parts across different articulations, \eg node $(3, 3)$ is used for representing the upper part of the left leg, whereas node $(1,1)$ is used for representing the upper body.
    }
    \label{fig:semantic_hierarchy_d_faust}
    \vspace{-1em}
\end{figure}

In this section, we provide evidence that our structure-aware representation yields semantic shape abstractions while achieving competitive (or even better results) than various state-of-the-art shape reconstruction methods, such as \cite{Mescheder2019CVPR}. Moreover, we also investigate the quality of the learned hierarchies and show that the use of our structure-aware representation yields semantic scene parsings. Implementation details and ablations on the impact of various components of our model are detailed in the supplementary.

\boldparagraph{Datasets}
First, we use the ShapeNet \cite{Chang2015ARXIV} subset of Choy \etal \cite{Choy2016ECCV}, training our model using the same image renderings and train/test splits as Choy \etal. Furthermore, we also experiment with the Dynamic FAUST (D-FAUST) dataset \cite{Bogo2017CVPR}, which contains meshes for 129 sequences of 10 humans performing various tasks, such as "running", "punching" or "shake arms".
We randomly divide these sequences into training (91), test (29) and validation (9).

\boldparagraph{Baselines}
Closely related to our work are the shape parsing methods of \cite{Tulsiani2017CVPRa} and \cite{Paschalidou2019CVPR} that employ cuboids and superquadric surfaces as primitives. We refer to \cite{Paschalidou2019CVPR} as SQs and we evaluate using their publicly available code\footnote{\url{https://superquadrics.com}}. Moreover, we also compare to the Structured Implicit Function (SIF) \cite{Genova2019ICCV} that represent the object's geometry as the isolevel of the sum of a set of Gaussians and to the CvxNets \cite{Deng2019ARXIV} that represent the object parts using smooth convex shapes. Finally, we also report results for OccNet \cite{Mescheder2019CVPR}, which is the state-of-the-art implicit shape reconstruction technique. Note that in contrast to us, \cite{Mescheder2019CVPR} does not consider part decomposition or any form of latent structure.

\boldparagraph{Evaluation Metrics}
Similar to \cite{Mescheder2019CVPR, Genova2019ICCV, Deng2019ARXIV}, we evaluate our model quantitatively and report the mean Volumetric IoU and the Chamfer-$L_1$ distance. Both metrics are discussed in detail in our supplementary.

\begin{figure}
    \centering
    		            		            		            		        		            		            \begin{subfigure}[b]{0.15\linewidth}
		\centering
		\includegraphics[width=\linewidth]{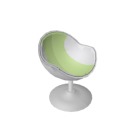}
    \end{subfigure}
    \hfill
    \begin{subfigure}[b]{0.15\linewidth}
		\centering
		\includegraphics[width=\linewidth]{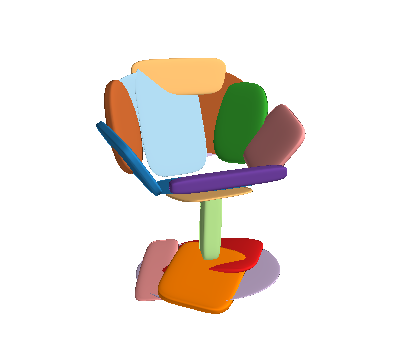}
    \end{subfigure}
    \hfill
    \begin{subfigure}[b]{0.15\linewidth}
		\centering
		\includegraphics[width=\linewidth]{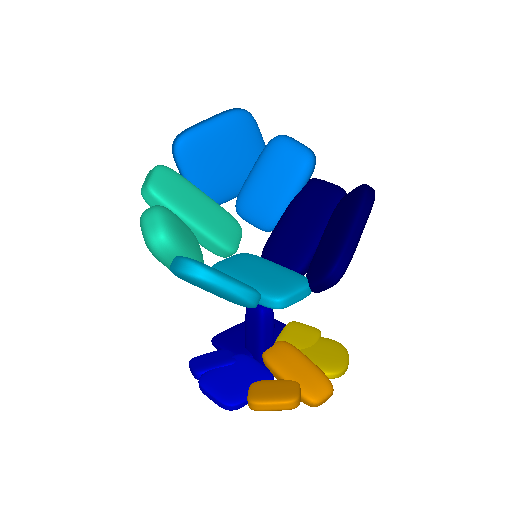}
    \end{subfigure}
    \hfill
    \begin{subfigure}[b]{0.15\linewidth}
		\centering
		\includegraphics[width=\linewidth]{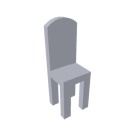}
    \end{subfigure}
    \begin{subfigure}[b]{0.15\linewidth}
		\centering
		\includegraphics[width=\linewidth]{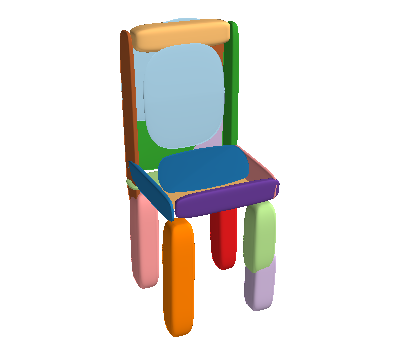}
    \end{subfigure}
    \hfill
    \begin{subfigure}[b]{0.15\linewidth}
		\centering
		\includegraphics[width=\linewidth]{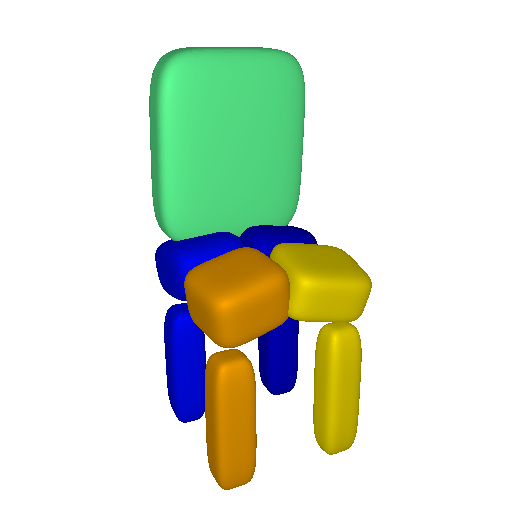}
    \end{subfigure}
    \vskip\baselineskip    \vspace{-1.4em}
    \begin{subfigure}[b]{0.15\linewidth}
		\centering
		\includegraphics[width=\linewidth]{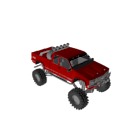}
    \end{subfigure}
    \hfill
    \begin{subfigure}[b]{0.15\linewidth}
		\centering
		\includegraphics[width=\linewidth]{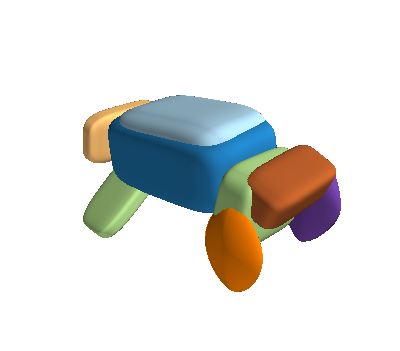}
    \end{subfigure}
    \hfill
    \begin{subfigure}[b]{0.15\linewidth}
		\centering
		\includegraphics[width=\linewidth]{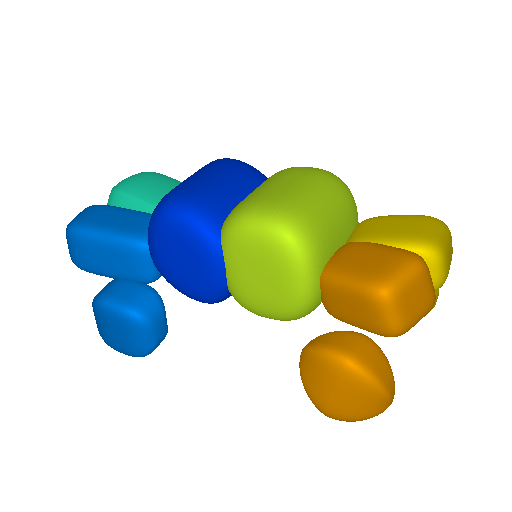}
    \end{subfigure}
    \hfill
    \begin{subfigure}[b]{0.15\linewidth}
		\centering
		\includegraphics[width=\linewidth]{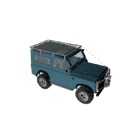}
    \end{subfigure}
    \begin{subfigure}[b]{0.15\linewidth}
		\centering
		\includegraphics[width=\linewidth]{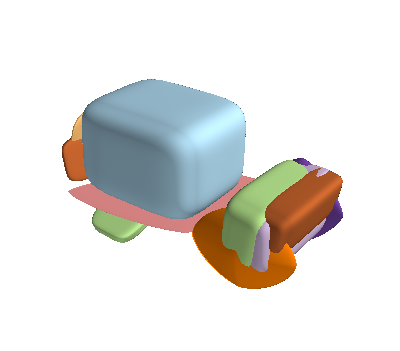}
    \end{subfigure}
    \hfill
    \begin{subfigure}[b]{0.15\linewidth}
		\centering
		\includegraphics[width=\linewidth]{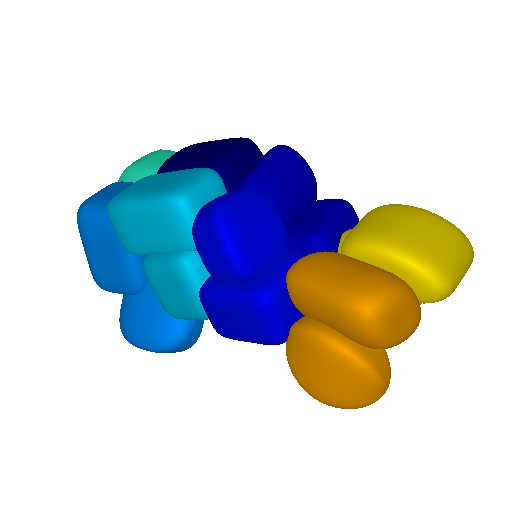}
    \end{subfigure}
    \vskip\baselineskip    \vspace{-1.4em}
    \begin{subfigure}[b]{0.15\linewidth}
		\centering
		\includegraphics[width=\linewidth]{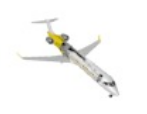}
    \end{subfigure}
    \hfill
    \begin{subfigure}[b]{0.15\linewidth}
		\centering
		\includegraphics[width=0.8\linewidth]{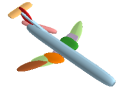}
    \end{subfigure}
    \hfill
    \begin{subfigure}[b]{0.15\linewidth}
		\centering
		\includegraphics[width=\linewidth]{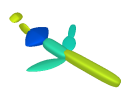}
    \end{subfigure}
    \hfill
    \begin{subfigure}[b]{0.15\linewidth}
		\centering
		\includegraphics[width=\linewidth]{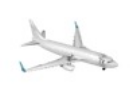}
    \end{subfigure}
    \begin{subfigure}[b]{0.15\linewidth}
		\centering
		\includegraphics[width=\linewidth]{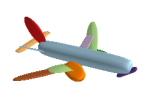}
    \end{subfigure}
    \hfill
    \begin{subfigure}[b]{0.15\linewidth}
		\centering
		\includegraphics[width=\linewidth]{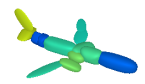}
    \end{subfigure}
    \vskip\baselineskip    \vspace{-1.2em}
    \begin{subfigure}[b]{0.15\linewidth}
		\centering
        \caption{\textbf{Input}}
    \end{subfigure}
    \hfill
    \begin{subfigure}[b]{0.15\linewidth}
		\centering
        \caption{\bf{SQs}}
    \end{subfigure}
    \hfill
    \begin{subfigure}[b]{0.15\linewidth}
		\centering
        \caption{\textbf{Ours}}
    \end{subfigure}
    \hfill
    \begin{subfigure}[b]{0.15\linewidth}
		\centering
        \caption{\textbf{Input}}
    \end{subfigure}
    \hfill
    \begin{subfigure}[b]{0.15\linewidth}
		\centering
        \caption{\bf{SQs}}
    \end{subfigure}
    \hfill
    \begin{subfigure}[b]{0.15\linewidth}
		\centering
        \caption{\textbf{Ours}}
    \end{subfigure}
    \caption{\textbf{Single Image 3D Reconstruction.} The input image is shown in (a, d), the other columns show the results of our method (c, f) compared to \cite{Paschalidou2019CVPR} (b, e). Additional qualitative results are provided in the supplementary.}
    \label{fig:qualitative_shapenet_rgb}
    \vspace{-1.0em}
\end{figure}

\subsection{Results on ShapeNet}

We evaluate our model on the single-view 3D reconstruction task and compare against various state-of-the-art methods. We follow the standard experimental setup and train a single model for the $13$ ShapeNet objects.
Both our model and $\cite{Paschalidou2019CVPR}$ are trained for a maximum number of $64$ primitives ($D=6$). For SIF \cite{Genova2019ICCV} and CvxNets \cite{Deng2019ARXIV} the reported results are computed using $50$ shape elements.
The quantitative results are reported in \tabref{tab:shapenet_v1_quantitative}. We observe that our model outperforms the primitive-based baselines in terms of the IoU as well as the OccNet \cite{Mescheder2019CVPR} for the majority of objects $(7/13)$.
Regarding Chamfer-$L_1$, our model is the second best amongst primitive representations, as \cite{Paschalidou2019CVPR} is optimized for this metric. This also justifies that \cite{Paschalidou2019CVPR} performs worse in terms of IoU. While our model performs on par with existing state-of-the-art primitive representations in terms of Chamfer-$L_1$, it also recovers hierarchies, which none of our baselines do.
A qualitative comparison of our model with SQs \cite{Paschalidou2019CVPR} is depicted in \figref{fig:qualitative_shapenet_rgb}. \figref{fig:shapenet_hiearchy} visualizes the learned hierarchy for this model. We observe that our model recovers unbalanced binary trees that decompose a 3D object into a set of parts.
Note that \cite{Tulsiani2017CVPRa, Paschalidou2019CVPR} were originally introduced for volumetric 3D reconstruction, thus we provide an experiment on this task in our supplementary.

\subsection{Results on D-FAUST}

\begin{figure}
    \centering
    \centering
    \begin{subfigure}[b]{0.15\linewidth}
		\centering
		\includegraphics[width=\linewidth]{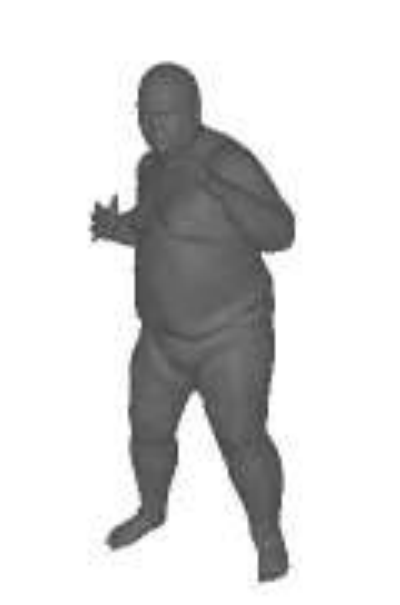}
    \end{subfigure}
    \hfill
    \begin{subfigure}[b]{0.15\linewidth}
		\centering
		\includegraphics[width=0.95\linewidth]{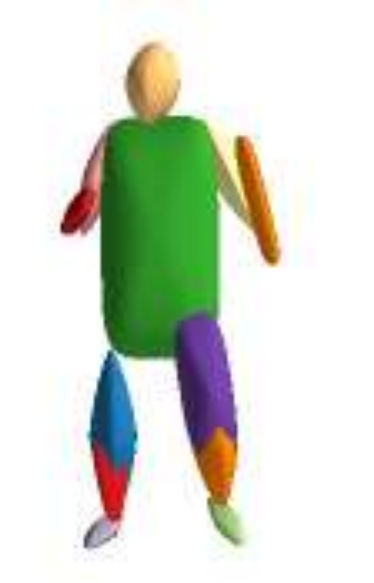}
    \end{subfigure}
    \hfill
    \begin{subfigure}[b]{0.15\linewidth}
		\centering
		\includegraphics[width=\linewidth]{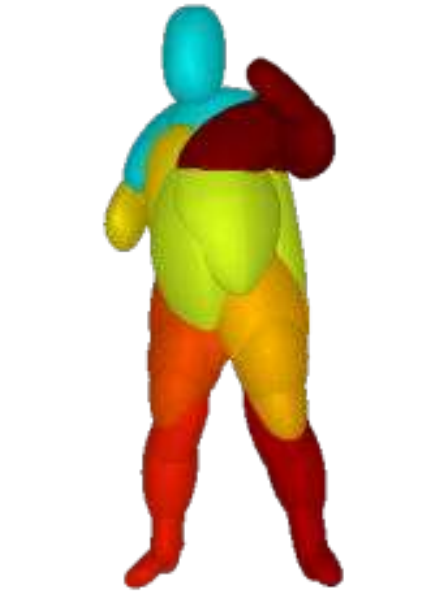}
    \end{subfigure}
    \hfill
    \begin{subfigure}[b]{0.15\linewidth}
		\centering
		\includegraphics[width=0.9\linewidth]{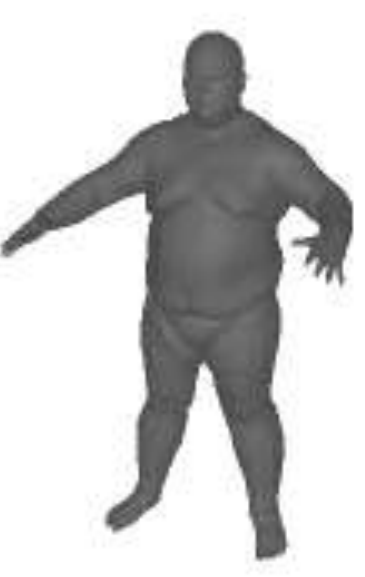}
    \end{subfigure}
    \hfill
    \begin{subfigure}[b]{0.15\linewidth}
		\centering
		\includegraphics[width=0.9\linewidth]{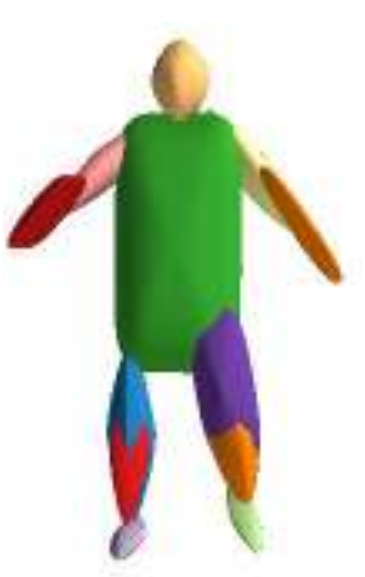}
    \end{subfigure}
    \hfill
    \begin{subfigure}[b]{0.15\linewidth}
		\centering
		\includegraphics[width=0.9\linewidth]{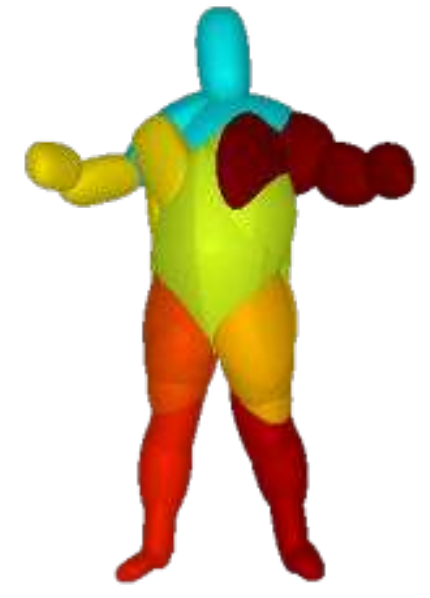}
    \end{subfigure}
    \vspace{-1.2em}
    \vskip\baselineskip    \begin{subfigure}[b]{0.15\linewidth}
		\centering
        \includegraphics[width=1.1\linewidth]{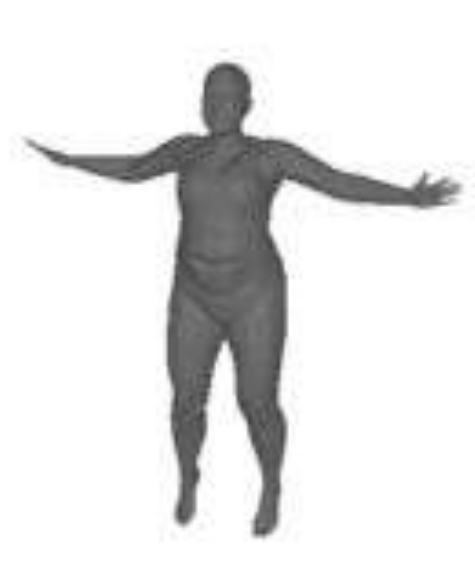}
		    \end{subfigure}
    \hfill
    \begin{subfigure}[b]{0.15\linewidth}
		\centering
        \includegraphics[width=1.1\linewidth]{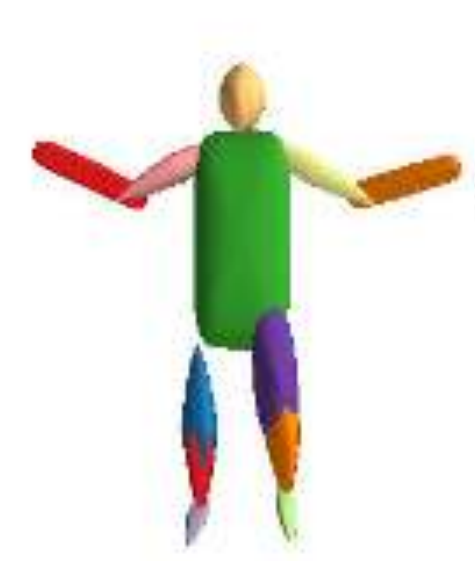}
		    \end{subfigure}
    \hfill
    \begin{subfigure}[b]{0.15\linewidth}
		\centering
        \includegraphics[width=1.0\linewidth]{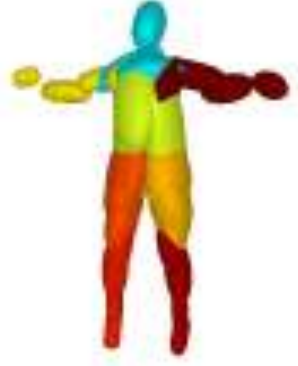}
		    \end{subfigure}
    \hfill
    \begin{subfigure}[b]{0.15\linewidth}
		\centering
        \includegraphics[width=0.9\linewidth]{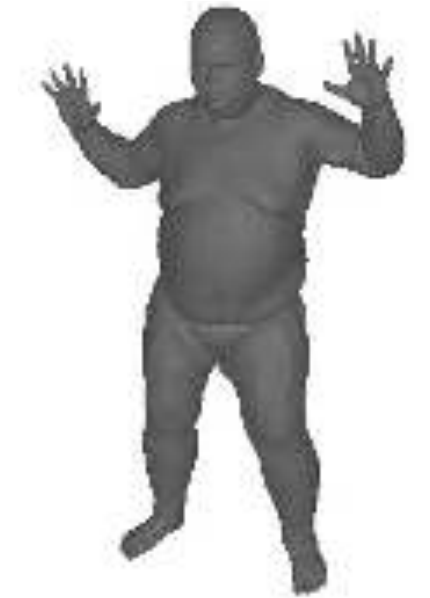}
		    \end{subfigure}
    \hfill
    \begin{subfigure}[b]{0.15\linewidth}
		\centering
        \includegraphics[width=0.9\linewidth]{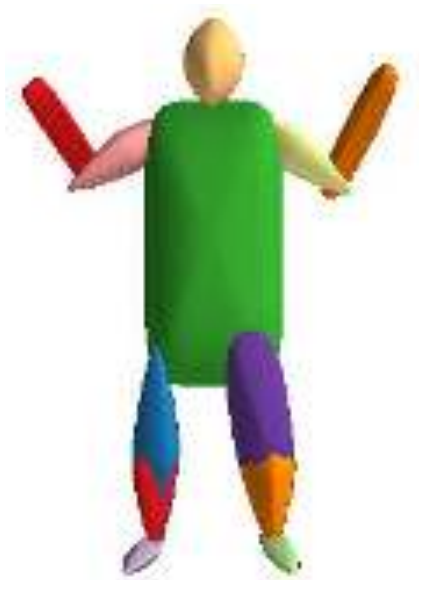}
		    \end{subfigure}
    \hfill
    \begin{subfigure}[b]{0.15\linewidth}
		\centering
        \includegraphics[width=0.9\linewidth]{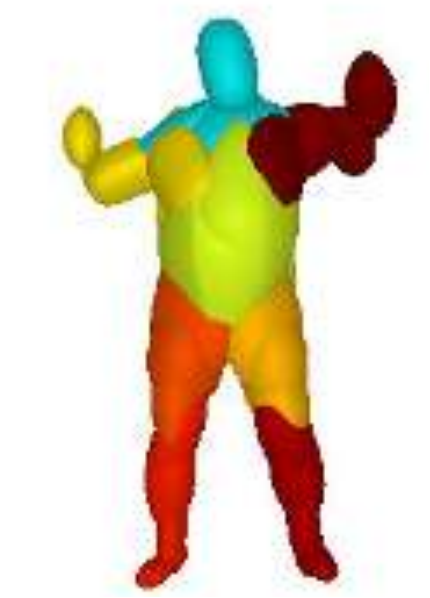}
		    \end{subfigure}
    \vskip\baselineskip    \vspace{-1.2em}
    \begin{subfigure}[b]{0.15\linewidth}
		\centering
        \caption{\textbf{Input}}
    \end{subfigure}
    \hfill
    \begin{subfigure}[b]{0.15\linewidth}
		\centering
        \caption{\bf{SQs}}
    \end{subfigure}
    \hfill
    \begin{subfigure}[b]{0.15\linewidth}
		\centering
        \caption{\textbf{Ours}}
    \end{subfigure}
    \hfill
    \begin{subfigure}[b]{0.15\linewidth}
		\centering
        \caption{\textbf{Input}}
    \end{subfigure}
    \hfill
    \begin{subfigure}[b]{0.15\linewidth}
		\centering
        \caption{\bf{SQs}}
    \end{subfigure}
    \hfill
    \begin{subfigure}[b]{0.15\linewidth}
		\centering
        \caption{\textbf{Ours}}
    \end{subfigure}
        \caption{{\bf Single Image 3D Reconstruction}. Qualitative comparison of our reconstructions (c, f), to \cite{Paschalidou2019CVPR} that does not consider any form of structure (b, e). The input RGB image is shown in (a, d). Note how our representation yields geometrically more accurate reconstructions, while being semantic, \eg, the primitive colored in blue consistently represents the head of the human while the primitive colored in orange captures the left thigh. Additional qualitative results are provided in the supplementary.}
    \label{fig:qualitative_dfaust}
    \vspace{-0.2em}
\end{figure}

We also demonstrate results on the Dynamic FAUST (D-FAUST) dataset \cite{Bogo2017CVPR}, which is very challenging due to the fine structure of the human body.
We evaluate our model on the single-view 3D reconstruction task and compare with \cite{Paschalidou2019CVPR}.
Both methods are trained for a maximum number of $32$ primitives ($D=5$). \figref{fig:semantic_hierarchy_d_faust} illustrates the predicted hierarchy on different humans from the test set. We note that the predicted hierarchies are indeed semantic, as the same nodes are used for modelling the same part of the human body. \figref{fig:qualitative_dfaust} compares the predictions of our model with SQs. We observe that while our baseline yields more parsimonious abstractions, their level of detail is limited. On the contrary, our model captures the geometry of the human body with more detail. This is also validated quantitatively, from Table \ref{tab:dfaust_quantitative}.
Note that in contrast to ShapeNet, D-FAUST does not contain long, thin (\eg legs of tables, chairs) or hollow parts (\eg cars), thus optimizing for either Chamfer-L1 or IoU leads to similar results. Hence, our method outperforms \cite{Paschalidou2019CVPR} also in terms of Chamfer-L1. Due to lack of space, we only illustrate the predicted hierarchies up to the fourth depth level. The full hierarchies are provided in the supplementary.

\def\arraystretch{0.95}
\begin{table}
    \centering
    \begin{tabular}{lcc}
        \toprule
        & IoU & Chamfer-$L_1$ \\
        \midrule
        SQs \cite{Paschalidou2019CVPR} & 0.608 & 0.189\\
        Ours & \bf{0.699} & \bf{0.098}\\
        \bottomrule
    \end{tabular}
    \caption{{\bf Single Image Reconstruction on D-FAUST.} We report the volumetric IoU and the Chamfer-L1 \wrt the ground-truth mesh for our model compared to \cite{Paschalidou2019CVPR}.}    \label{tab:dfaust_quantitative}
    \vspace{-1em}
\end{table}

\section{Conclusion}

We propose a learning-based approach that jointly predicts part relationships together with per-part geometries in the form of a binary tree without requiring any part-level annotations for training. Our model yields geometrically accurate primitive-based reconstructions that outperform existing shape abstraction techniques while performing competitively with more flexible implicit shape representations.
In future work, we plan to to extend our model and predict hierarchical structures that remain consistent in time, thus yielding kinematic trees of objects.
Another future direction, is to consider more flexible primitives such as general convex shapes and incorporate additional constraints \eg symmetry to further improve the reconstructions.

\section*{Acknowledgments}

This research was supported by the Max Planck ETH Center for Learning Systems
and a HUAWEI research gift. 

{\small
	\bibliographystyle{ieee_fullname}
	\bibliography{bibliography_long,bibliography,bibliography_custom}
}

\newpage
\clearpage
\appendix
\onecolumn

\begin{minipage}{\textwidth}
   \null
   \vskip .375in
   \begin{center}
      {\Large \bf Supplementary Material for\\Learning Unsupervised Hierarchical Part Decomposition of 3D Objects\\ from a Single RGB Image \par}
            \vspace*{24pt}
      {
      \large
      \lineskip .5em
      \begin{tabular}[t]{c}
        Despoina Paschalidou$^{1,3,5}$ \quad Luc van Gool$^{3,4,5}$ \quad Andreas Geiger$^{1,2,5}$\\
        $^1$Max Planck Institute for Intelligent Systems T{\"u}bingen\\
        $^2$University of T{\"u}bingen\quad
        $^3$Computer Vision Lab, ETH Z{\"u}rich\quad
        $^4$KU Leuven \\
        $^5$Max Planck ETH Center for Learning Systems\\
        {\tt\small \{firstname.lastname\}@tue.mpg.de \quad vangool@vision.ee.ethz.ch}
         \vspace*{1pt}\\      \end{tabular}
      \par
      }
            \vskip .5em
            \vspace*{12pt}
   \end{center}
\end{minipage}

\begin{abstract}
In this \textbf{supplementary document}, we first present examples of our occupancy function.
In addition, we present a detailed overview of our network architecture and the training procedure. We then discuss how various components influence the performance of our model on the single-view 3D reconstruction task.
Finally, we provide additional experimental results on more categories from the ShapeNet dataset \cite{Chang2015ARXIV} and on the D-FAUST dataset \cite{Bogo2017CVPR} together with the corresponding hierarchical structures.
The \textbf{supplementary video} shows 3D animations of the predicted structural hierarchy for various objects from the ShapeNet dataset as well as humans from the D-FAUST.
\end{abstract}

\section{Occupancy Function}
\begin{figure}[b]
    \centering
    \includegraphics[width=0.6\linewidth]{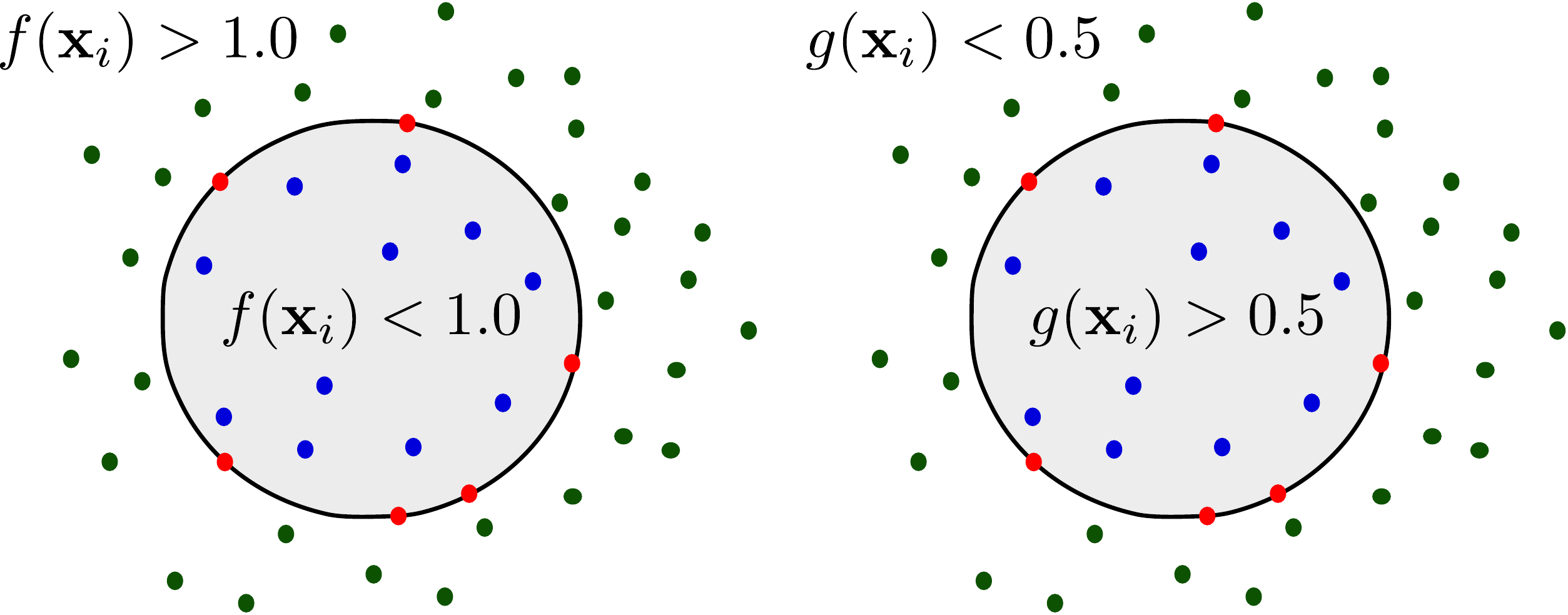}
    \caption{\textbf{Implicit surface function of superquadrics.} We visualize
    the 2D slice of $f(\bx_i)$ and $g(\bx_i)$ for  a superquadric with
    $\alpha_1=\alpha_2=\alpha_3=\epsilon_1=\epsilon_2=1$.}
    \label{fig:implicit_surface_sq}
\end{figure}

\begin{figure}
    \begin{subfigure}[b]{1.0\linewidth}
        \centering
        \includegraphics[width=1.0\linewidth]{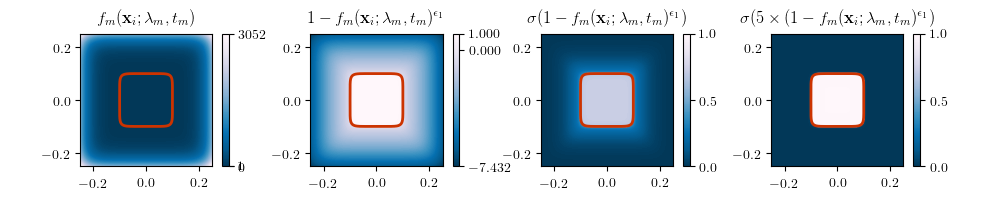}
        \caption{$\epsilon_1 = 0.25$, and $\epsilon_2=0.25$}
    \end{subfigure}
    \begin{subfigure}[b]{1.0\linewidth}
        \centering
        \includegraphics[width=1.0\linewidth]{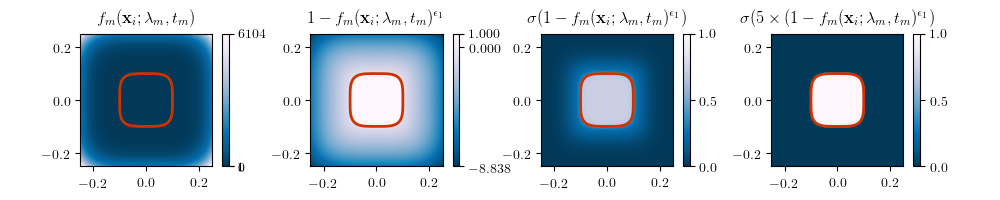}
        \caption{$\epsilon_1 = 0.25$, and $\epsilon_2=0.5$}
    \end{subfigure}
    \begin{subfigure}[b]{1.0\linewidth}
        \centering
        \includegraphics[width=1.0\linewidth]{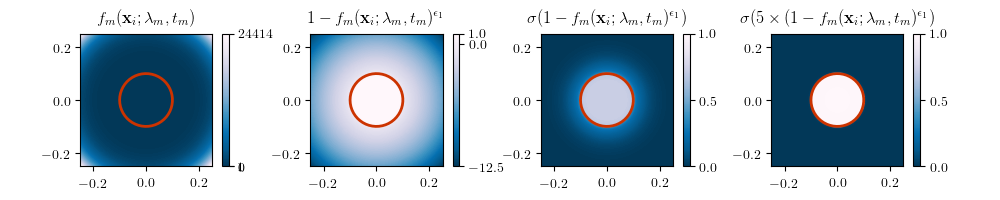}
        \caption{$\epsilon_1 = 0.25$, and $\epsilon_2=1.0$}
    \end{subfigure}
    \begin{subfigure}[b]{1.0\linewidth}
        \centering
        \includegraphics[width=1.0\linewidth]{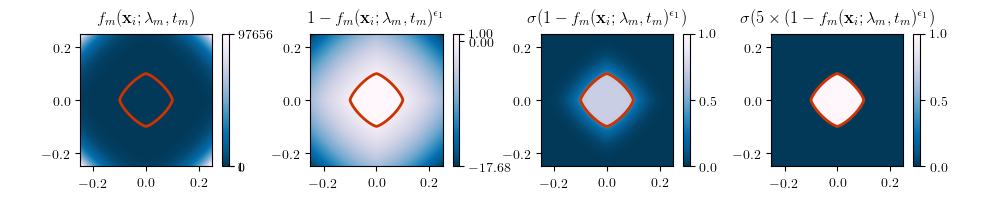}
        \caption{$\epsilon_1 = 0.25$, and $\epsilon_2=1.5$}
    \end{subfigure}
    \caption{\textbf{Implicit surface function} We visualize the implicit
    surface function for different primitive parameters and for different
    sharpness values. The surface boundary is drawn with {\color{red}{red}}.}
    \label{fig:sharpness_1}
\end{figure}

\begin{figure}
    \begin{subfigure}[b]{1.0\linewidth}
        \centering
        \includegraphics[width=1.0\linewidth]{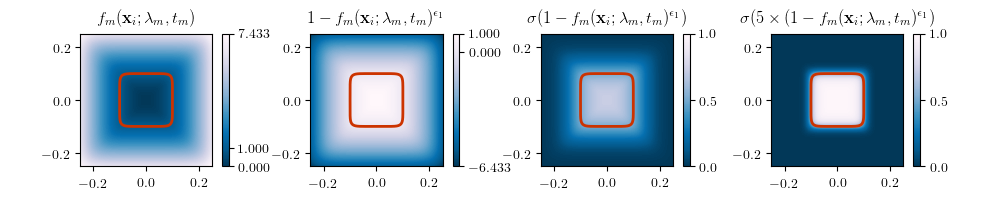}
        \caption{$\epsilon_1 = 1.0$, and $\epsilon_2=0.25$}
    \end{subfigure}
    \begin{subfigure}[b]{1.0\linewidth}
        \centering
        \includegraphics[width=1.0\linewidth]{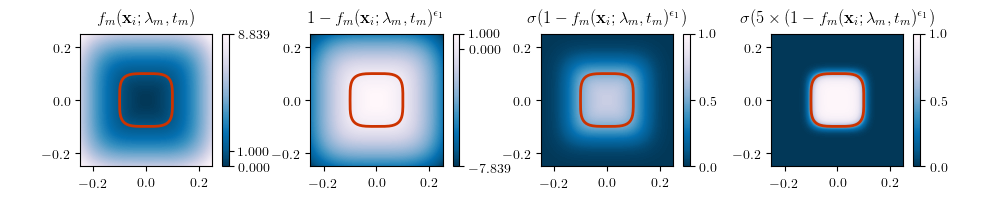}
        \caption{$\epsilon_1 = 1.0$, and $\epsilon_2=0.5$}
    \end{subfigure}
    \begin{subfigure}[b]{1.0\linewidth}
        \centering
        \includegraphics[width=1.0\linewidth]{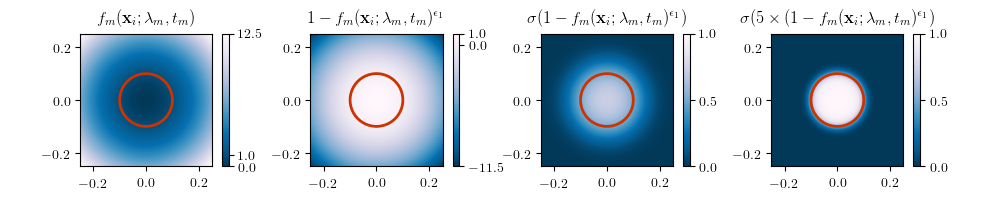}
        \caption{$\epsilon_1 = 1.0$, and $\epsilon_2=1.0$}
    \end{subfigure}
    \begin{subfigure}[b]{1.0\linewidth}
        \centering
        \includegraphics[width=1.0\linewidth]{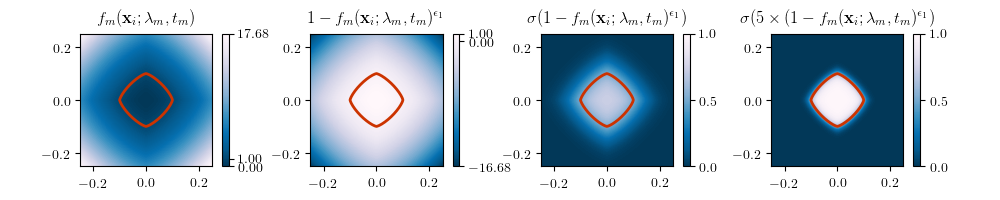}
        \caption{$\epsilon_1 = 1.0$, and $\epsilon_2=1.5$}
    \end{subfigure}
    \caption{\textbf{Implicit surface function} We visualize the implicit
    surface function for different primitive parameters and for different
    sharpness values. The surface boundary is drawn with {\color{red}{red}}.}
    \label{fig:sharpness_3}
\end{figure}

\begin{figure}
    \begin{subfigure}[b]{1.0\linewidth}
        \centering
        \includegraphics[width=1.0\linewidth]{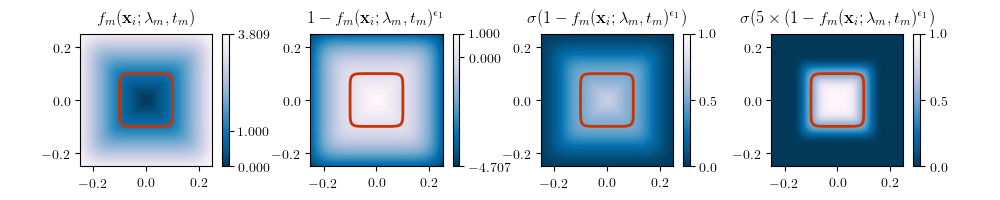}
        \caption{$\epsilon_1 = 1.5$, and $\epsilon_2=0.25$}
    \end{subfigure}
    \begin{subfigure}[b]{1.0\linewidth}
        \centering
        \includegraphics[width=1.0\linewidth]{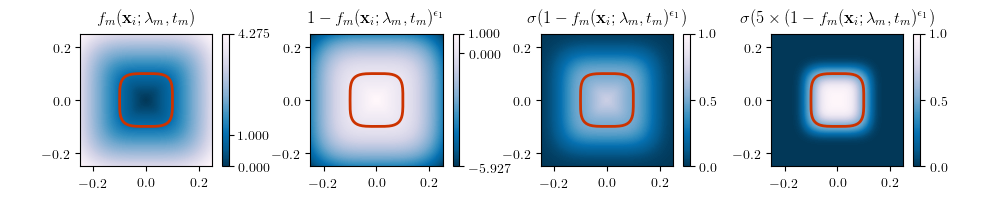}
        \caption{$\epsilon_1 = 1.5$, and $\epsilon_2=0.5$}
    \end{subfigure}
    \begin{subfigure}[b]{1.0\linewidth}
        \centering
        \includegraphics[width=1.0\linewidth]{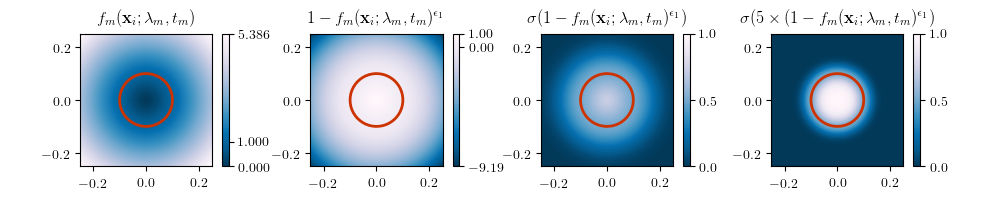}
        \caption{$\epsilon_1 = 1.5$, and $\epsilon_2=1.0$}
    \end{subfigure}
    \begin{subfigure}[b]{1.0\linewidth}
        \centering
        \includegraphics[width=1.0\linewidth]{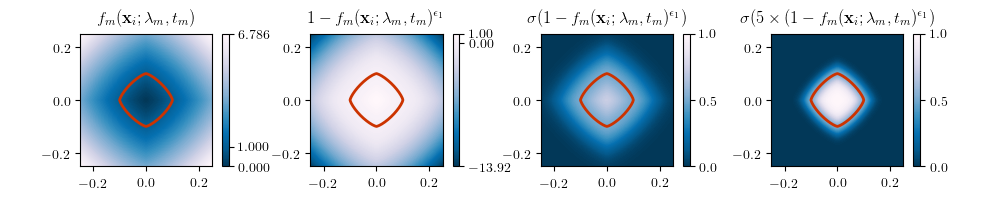}
        \caption{$\epsilon_1 = 1.5$, and $\epsilon_2=1.5$}
    \end{subfigure}
    \caption{\textbf{Implicit surface function} We visualize the implicit
    surface function for different primitive parameters and for different
    sharpness values. The surface boundary is drawn with {\color{red}{red}}.}
    \label{fig:sharpness_4}
\end{figure}

In this section, we provide illustrations of the occupancy function $g$ for different primitive parameters and for different sharpness values.
For any point $\bx \in \mathbb{R}^3$, we can determine whether it lies inside or outside a superquadric using its implicit surface function which is commonly referred to as the \textit{inside-outside function}:
\begin{equation}
    f(\bx; \lambda) = \left(\left(\frac{x}{\alpha_{1}}\right)^{\frac{2}{\epsilon_{2}}}
        + \left(\frac{y}{\alpha_{2}}\right)^{\frac{2}{\epsilon_{2}}}\right)^{\frac{\epsilon_{2}}{\epsilon_{1}}}
        + \left(\frac{z}{\alpha_{3}}\right)^{\frac{2}{\epsilon_{1}}}
    \label{eq:implicit_sq}
\end{equation}
where $\mathbf{\alpha} = [\alpha_{1}, \alpha_{2}, \alpha_{3}]$ determine the size and $\mathbf{\epsilon} = [\epsilon_{1}, \epsilon_{2}]$ determine the shape of the superquadric.
If $f(\bx; \lambda) = 1.0$, the given point $\bx$ lies on the surface of the superquadric, if $f(\bx; \lambda) < 1.0$ the corresponding point lies inside and if $f(\bx; \lambda) > 1.0$ the point lies outside the superquadric.
To account for numerical instabilities that arise from the exponentiations in \eqref{eq:implicit_sq}, instead of directly using $f(\bx; \lambda)$, we follow \cite{Jaklic2000} and use $f(\bx; \lambda)^{\epsilon_1}$.
In addition, we also convert the inside-outside function to an \emph{occupancy function}, $g:\mathbb{R}^3 \rightarrow [0, 1]$:
\begin{equation}
    g(\bx; \lambda) = \sigma\left(s\left(1 - f(\bx; \lambda)^{\epsilon_1}\right)\right)
    \label{eq:implicit_sq_actual}
\end{equation}
that results in per-point predictions suitable for the classification problem we want to solve. $\sigma(\cdot)$ is the sigmoid function and $s$ controls the sharpness of the transition of the occupancy function. As a result, if $g(\bx; \lambda) < 0.5$ the corresponding point lies outside and if $g(\bx; \lambda) > 0.5$ the point lies inside the superquadric. \figref{fig:implicit_surface_sq} visualizes the range of the implicit surface function of superquadrics of \eqref{eq:implicit_sq} and \eqref{eq:implicit_sq_actual}. \figref{fig:sharpness_1}$+$\ref{fig:sharpness_3}$+$\ref{fig:sharpness_4} visualize the implicit surface function for different values of $\epsilon_1$ and $\epsilon_2$ and different values of sharpness $s$. We observe that without applying the sigmoid to \eqref{eq:implicit_sq} the range of values of \eqref{eq:implicit_sq} varies significantly for different primitive parameters.

\clearpage
\newpage

\section{Implementation Details}

In this section, we provide a detailed description of our network architecture. We then describe our sampling strategy and provide details on the metrics we use both for training and testing. Finally, we show how various components influence the performance of our model on the single-view 3D reconstruction task.

\subsection{Network Architecture}
Here we describe the architecture of each individual component of our model, shown in Figure 3 of our main submission.

\begin{figure}[ht]
    \centering
    \begin{subfigure}[b]{\linewidth}
        \centering
        \includegraphics[width=0.6\linewidth]{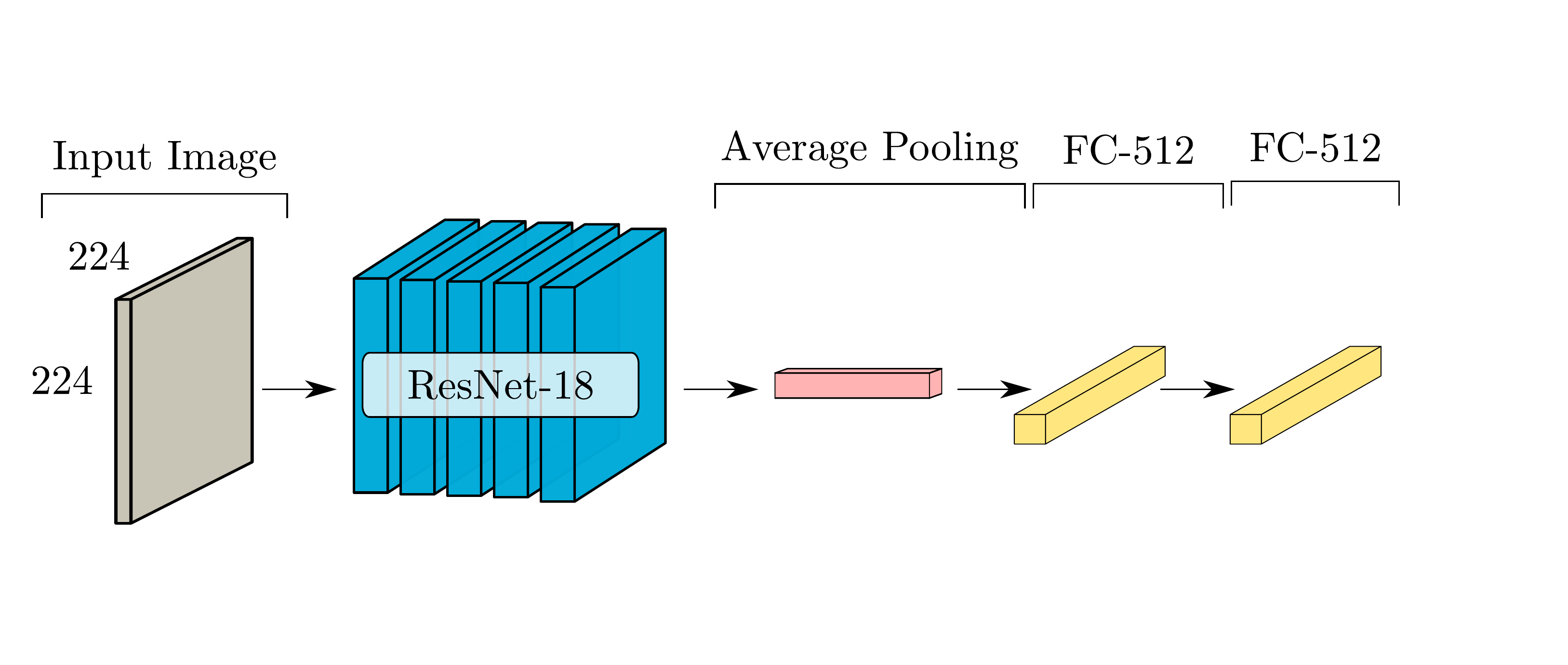}
        \caption{Single-view 3D Reconstruction}
        \label{fig:single_view_encoder}
    \end{subfigure}
    \vskip\baselineskip
    \begin{subfigure}[b]{\linewidth}
        \centering
        \includegraphics[width=\linewidth]{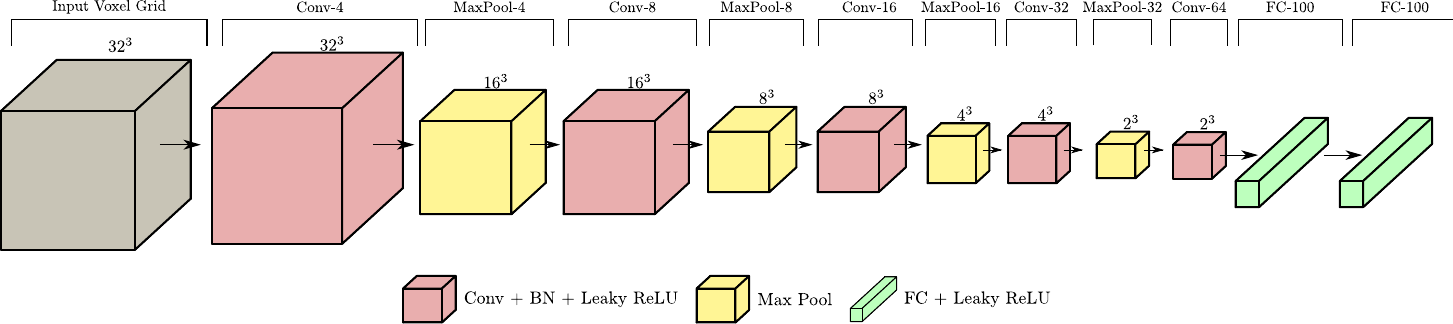}
        \caption{Volumetric 3D Reconstruction}
        \label{fig:volumetric_encoder}
    \end{subfigure}
    \caption{{\bf{Feature Encoder Architectures.}} Depending on the type of the input, we employ two different network architectures. (a) For the single view 3D reconstruction task we use a ResNet-18 architecture \cite{He2016CVPR} (b) For a binary occupancy grid as an input, we leverage the network architecture of \cite{Paschalidou2019CVPR}.}
    \label{fig:encoder_architectures}
\end{figure}

\boldparagraph{Feature Encoder} The feature encoder depends on the type of the input, namely whether it is an image or a binary occupancy grid. For the single view 3D reconstruction task, we use a ResNet-18 architecture \cite{He2016CVPR} (\figref{fig:single_view_encoder}), which was pretrained on the ImageNet dataset \cite{Deng2009CVPR}. From the original design, we ignore the final fully connected layer keeping only the feature vector of length $F=512$ after average pooling. For the volumetric 3D reconstruction task, where the input is a binary occupancy grid, we use the feature encoder proposed in \cite{Paschalidou2019CVPR}(\figref{fig:volumetric_encoder}). Note that the feature encoder is used as a generic feature extractor from the input representation.

\begin{figure}
    \centering
    \begin{subfigure}[b]{0.49\linewidth}
        \centering
        \includegraphics[width=\linewidth]{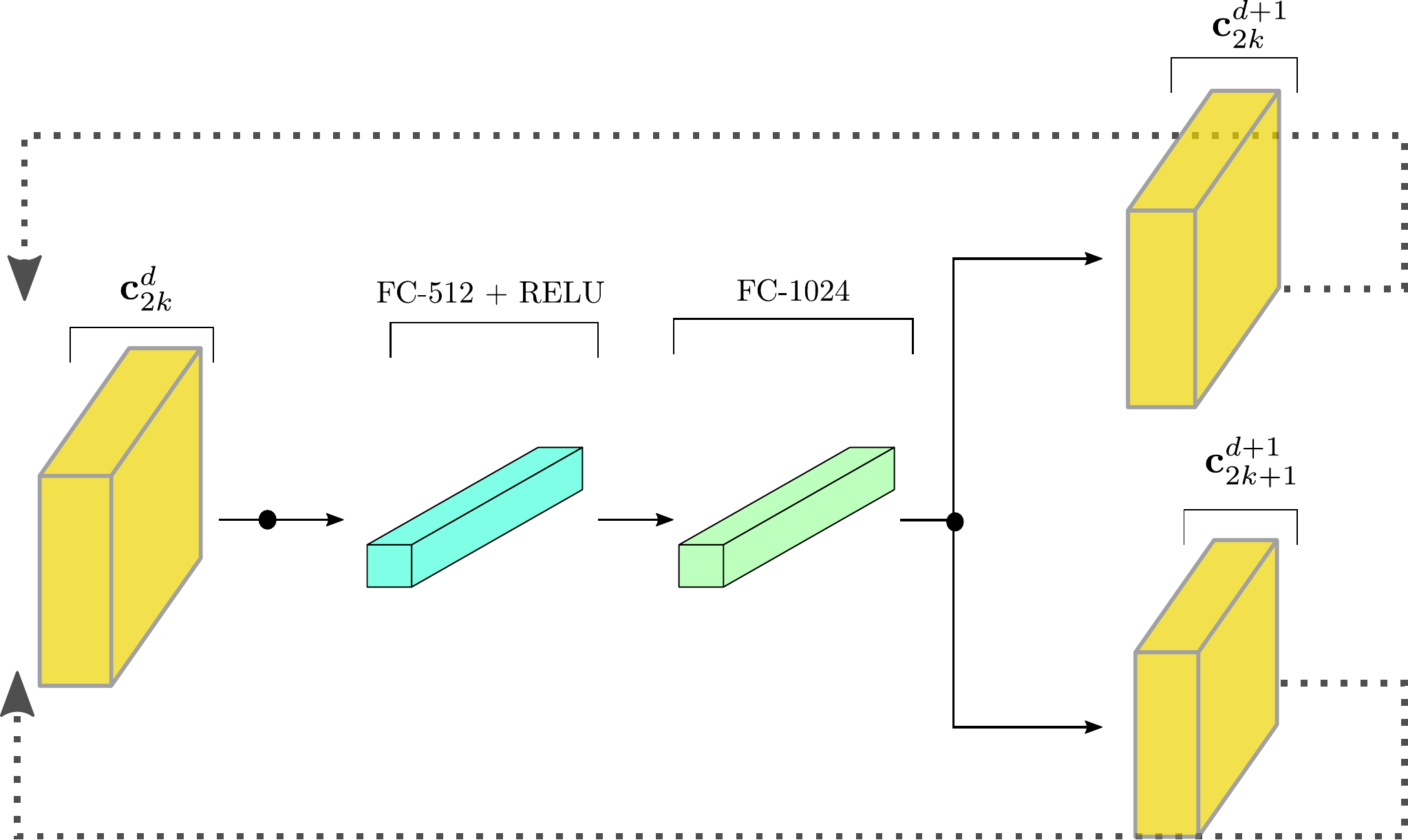}
        \caption{Partition Network}
        \label{fig:partition_network_architecture}
    \end{subfigure}
    \hfill
    \begin{subfigure}[b]{0.5\linewidth}
        \centering
        \includegraphics[width=0.8\linewidth]{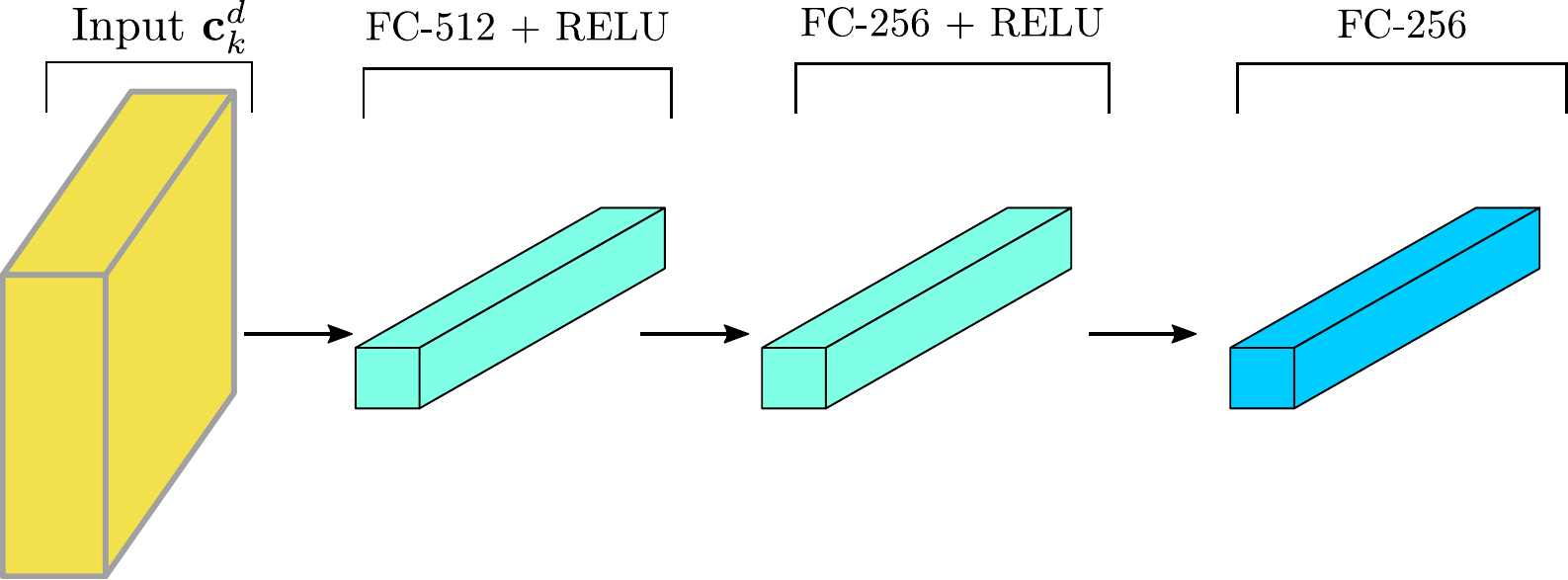}
        \caption{Structure Network}
        \label{fig:structure_network_architecture}
    \end{subfigure}
    \caption{{\bf{Network Architecture Overview.}} The \emph{partition network} (\ref{fig:partition_network_architecture}) is simply one hidden layer fully connected network with RELU non linearity. The {\color{gray}{gray}} dotted lines indicate the recursive partition of the feature representation. Similarly, the \emph{structure network} (\ref{fig:structure_network_architecture}) consists of two fully connected layers followed by RELU non linearity.
    }
\end{figure}

\boldparagraph{Partition Network} The partition network implements a function $p_{\theta}: \mathbb{R}^F \to \mathbb{R}^{2F}$ that recursively partitions the feature representation $\bc_k^d$ of node $p_k^d$ into two feature representations, one for each child $\{p_{2k}^{d+1}, p_{2k+1}^{d+1}\}$. The partition network (\figref{fig:partition_network_architecture}) comprises two fully connected layers followed by RELU non linearity. 

\boldparagraph{Structure Network} The structure network maps each feature representation $\bc_k^d$ to $\bh_k^d$ a spatial location in $\mathbb{R}^3$. The structure network (\figref{fig:structure_network_architecture}) consists of two fully connected layers followed by RELU non linearity.

\begin{figure}
    \centering
    \begin{subfigure}[b]{0.49\linewidth}
        \centering
        \includegraphics[width=0.9\linewidth]{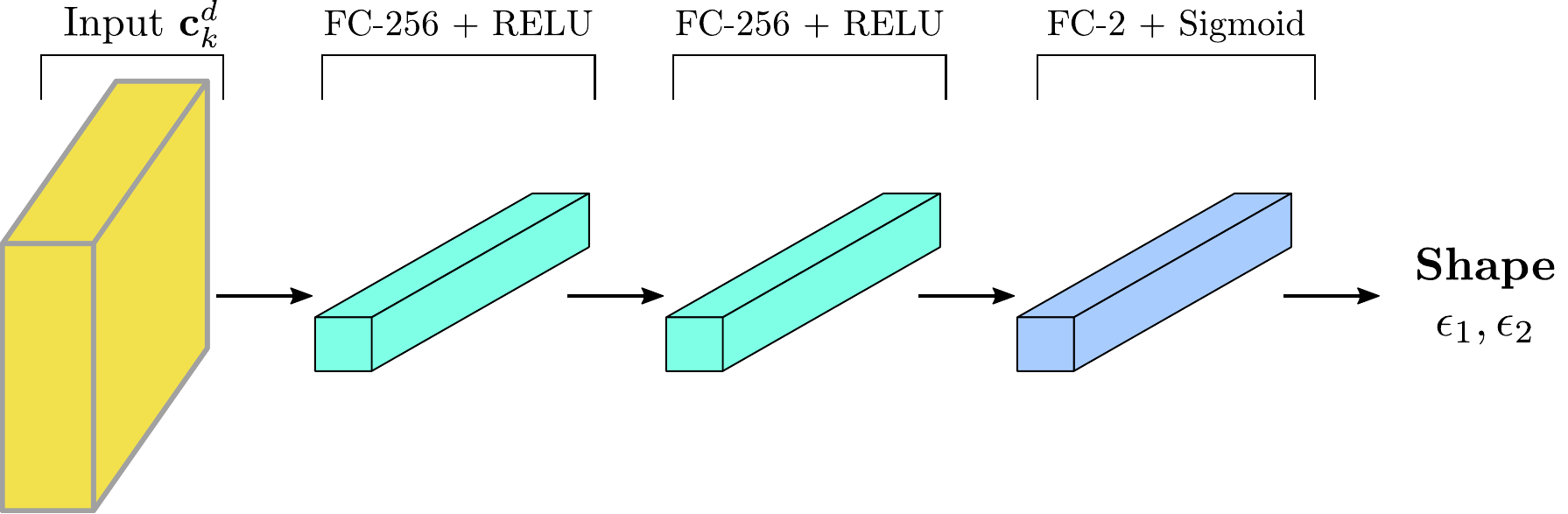}
        \caption{Shape}
        \label{fig:shape}
    \end{subfigure}
    \hfill
    \begin{subfigure}[b]{0.5\linewidth}
        \centering
        \includegraphics[width=0.9\linewidth]{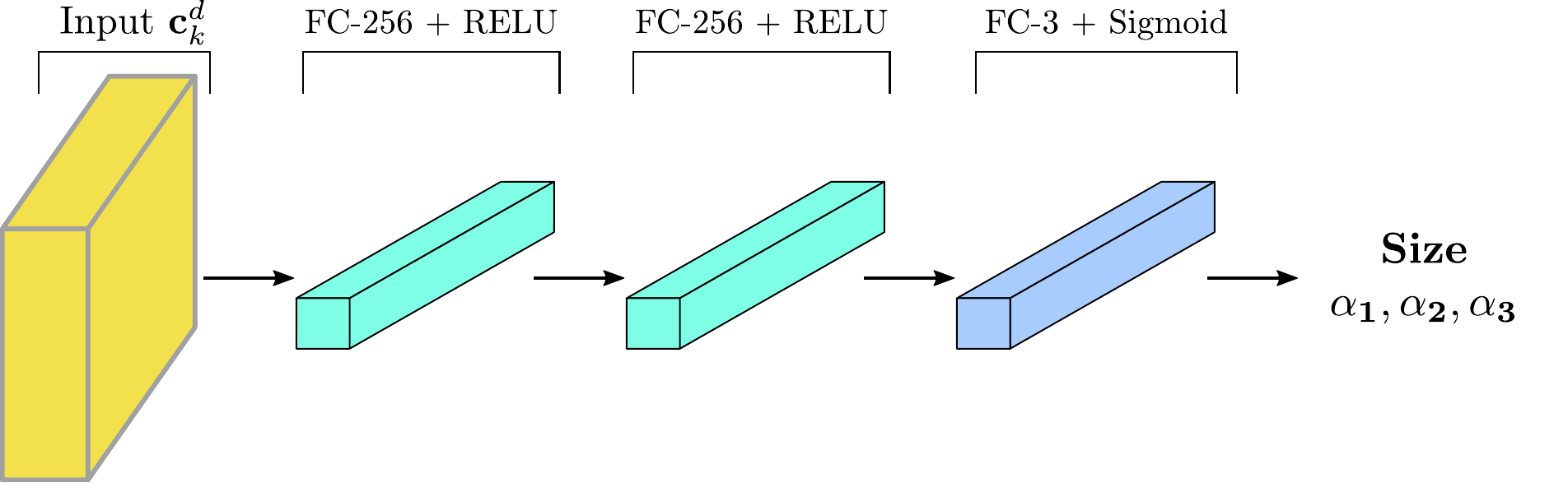}
        \caption{Size}
        \label{fig:size}
    \end{subfigure}
    \vskip\baselineskip
    \begin{subfigure}[b]{0.49\linewidth}
        \centering
        \includegraphics[width=0.9\linewidth]{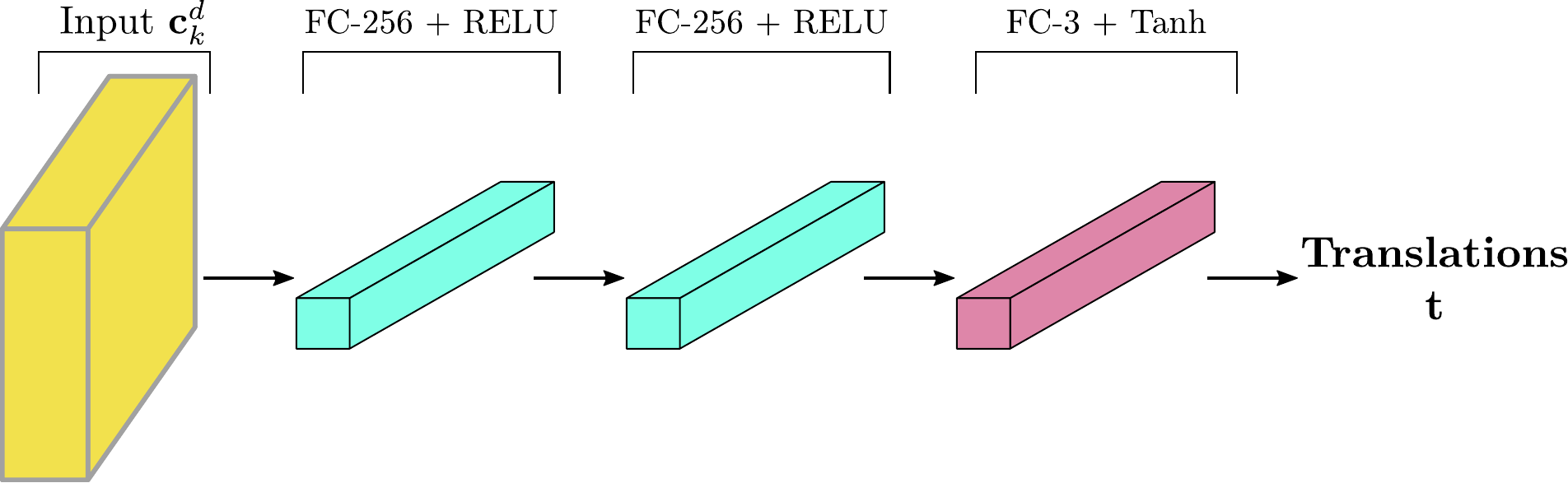}
        \caption{Translation}
        \label{figtranslation}
    \end{subfigure}
    \hfill
    \begin{subfigure}[b]{0.49\linewidth}
        \centering
        \includegraphics[width=0.9\linewidth]{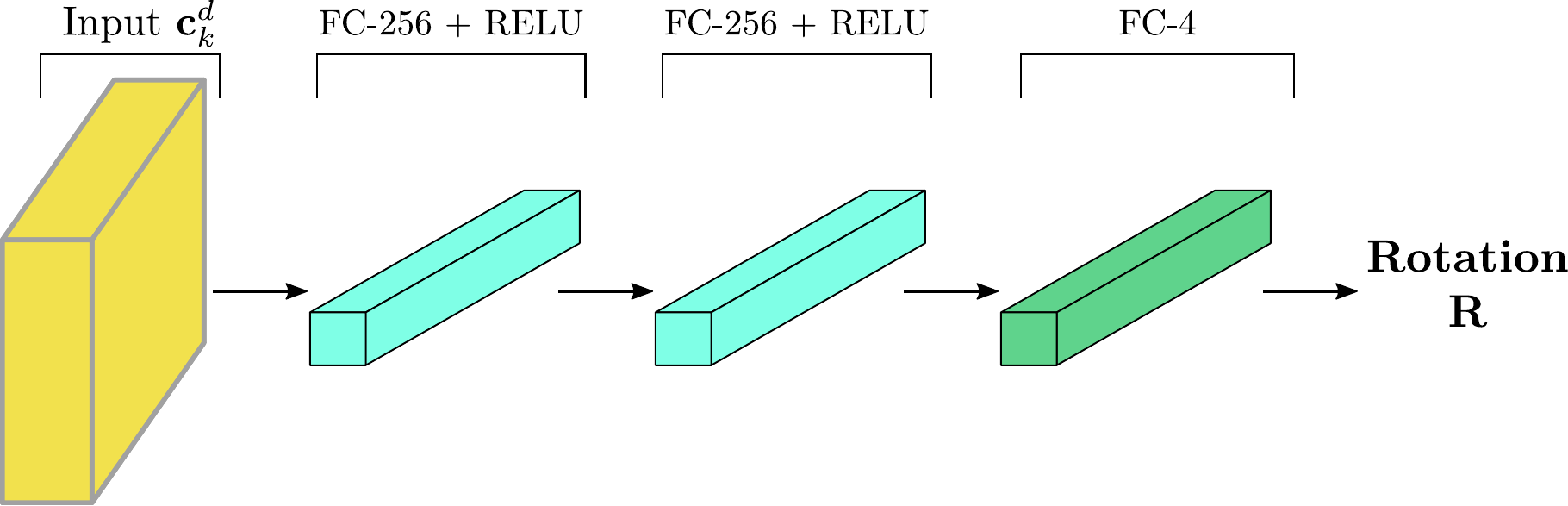}
        \caption{Rotation}
        \label{fig:rotation}
    \end{subfigure}
    \vskip\baselineskip
    \begin{subfigure}[b]{0.49\linewidth}
        \centering
        \includegraphics[width=0.9\linewidth]{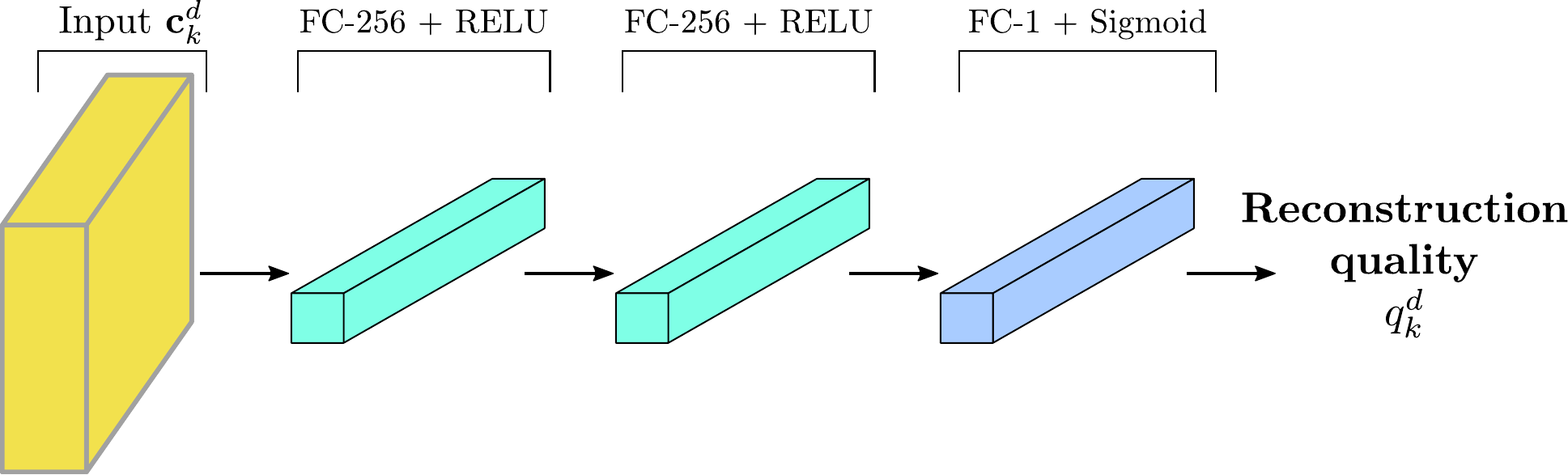}
        \caption{Reconstruction quality}
        \label{fig:qos}
    \end{subfigure}
    \caption{{\bf{Geometry Network.}} We detail the specifics of each regressor for predicting the primitive parameters $\lambda_k^d$ and the reconstruction quality $q_k^d$.}
    \label{fig:geometry_network}
\end{figure}

\boldparagraph{Geometry Network} The geometry network learns a function $r_{\theta}:\mathbb{R}^F \rightarrow \mathbb{R}^K \times [0, 1]$ that maps the feature representation $\bc_k^d$ to its corresponding primitive parametrization $\lambda_k^d$ and the reconstruction quality prediction $q_k^d$. In particular, the geometry network consists of five regressors that predict the parameters of the superquadrics (size $\mathbf{\alpha}$, shape $\mathbf{\epsilon}$ and pose as translation $\mathbf{t}$ and rotation $\mathbf{R}$) in addition to the reconstruction quality $q_k^d$. \figref{fig:geometry_network} presents the details of the implementation of each regressor.

\subsection{Training}
In all our experiments, we use the Adam optimizer \cite{Kingma2015ICLR} with learning rate $0.0001$ and no weight decay. For other hyperparameters of Adam we use the PyTorch defaults. We train all models with a batch size of $32$ for $40$k iterations. We do not perform any additional data augmentation.
We weigh the loss terms of Eq. 9 in our main submission with 0.1, 0.01, 0.01 and 0.1 respectively, in order to enforce that during the first stages of training the network will focus primarily on learning the hierarchical decomposition of the 3D shape ($\mathcal{L}_s + \mathcal{L}_p$). In this way, after the part decomposition is learned, the network also focuses on the part geometries $(\mathcal{L}_r)$. We also experimented with a two-stage optimization scheme, where we first learn the hierarchical part decomposition and then learn the hierarchical representation, but we observed that this made learning harder.

\subsection{Sampling Strategy}
\label{sec:sampling_strategy}
Sampling a point inside the target mesh has a probability proportional to the volume of the mesh. This yields bad reconstructions for thin parts of the object, such as legs of chairs and wings of aeroplanes. In addition, biasing the sampling towards the points inside the target mesh, results in worse reconstructions as also noted in \cite{Mescheder2019CVPR}.
To address the first issue (properly reconstructing thin parts), we use an unbalanced sampling distribution that, in expectation, results in sampling an equal number of points inside and outside the target mesh. To counter the second (biased sampling), we construct an unbiased estimator of the loss by weighing the per-point loss inversely proportionally to its sampling probability.
We refer to our sampling strategy as \textit{unbiased importance sampling}. Note that throughout all our experiments, we sample $10$k points in the bounding box of the target mesh using our sampling strategy.

\subsection{Metrics}
We evaluate our model and our baselines using the volumetric Intersection over Union (IoU) and the Chamfer-$L_1$ distance.
Note that as our method does not predict a single mesh, we sample points from each primitive proportionally to its area, such that the total number of sampled points from all primitives is equal to $100$k. For a fair comparison, we do the same for \cite{Tulsiani2017CVPRa, Paschalidou2019CVPR}. Below, we discuss in detail the computation of the volumetric IoU and the Chamfer-$L_1$.

Volumetric IoU is defined as the quotient of the volume of the intersection of the target $S_{target} $and the predicted $S_{pred}$ mesh and the volume of their union.
We obtain unbiased estimates of the volume of the intersection and the union by randomly sampling $100$k points from the bounding volume and determining if the points lie inside or outside the target / predicted mesh,

\begin{equation}
    \text{IoU}(S_{pred}, S_{target}) = \frac{\mid V(S_{pred} \cap S_{target})\mid}{\mid V(S_{pred} \cup S_{target})\mid}
\end{equation}
where $V(.)$ is a function that computes the volume of a mesh.

We obtain an unbiased estimator of the Chamfer-$L_1$ distance by sampling $100$k points on the surface of the target $S_{target}$ and the predicted $S_{pred}$ mesh. We denote $\cX=\{\bx_i\}_{i=1}^N$ the set of points sampled on the surface of the target mesh and $\cY=\{\by_i\}_{i=1}^N$ the set of points sampled on the surface of the predicted mesh.
We compute the Chamfer-$L_1$ as follows:

\begin{equation}
    D_{\text{chamfer}}(\cX, \cY) = \frac{1}{N} \sum_{\bx_i \in \cX}
    \min_{\by_j \in \cup\cY}\| \bx_i - \by_j\| +
    \frac{1}{N} \sum_{\by_i \in \cup\cY}
    \min_{\bx_j \in \cX}\| \by_i - \bx_j \|
    \label{eq:chamfer}
\end{equation}
The first term of \eqref{eq:chamfer} measures the \textit{completeness} of the predicted shape, namely how far is on average the closest predicted point from a ground-truth point. The second term measures the \textit{accuracy} of the predicted shape, namely how far on average is the closest ground-truth point from a predicted point.

To ensure a fair comparison with our baselines, we use the evaluation code of \cite{Mescheder2019CVPR} for the estimation of both the Volumetric IoU and the Chamfer-$L_1$.

\subsection{Empirical Analysis of Loss Formulation}\label{sec:ablations}
In this section, we investigate the impact of how various components of our model affect the performance on the single-image 3D reconstruction task.

\subsubsection{Impact of Sampling Strategy}
We first discuss how the sampling strategy affects the performance of our model. Towards this goal, we evaluate our model on the single-view 3D reconstruction task using three different sampling strategies: (a) uniform sampling in the bounding box that contains the target object (b) biased sampling (namely sampling an equal number of points inside and outside the target mesh without reweighing) and (c) unbiased importance sampling as described in \secref{sec:sampling_strategy}. 
All models are trained on the "chair" object category of ShapeNet using the same network architecture, the same number of sampled points ($N=$10k) and the same maximum number of primitives ($D=16$). The quantitative results on the test set of the "chair" category are shown in Table \ref{tab:sampling}. We observe that the proposed importance sampling strategy achieves the best results in terms of IoU.

Furthermore, we also examine the impact of the number of sampled points on the performance of our model. In particular, we train our model on the ``chair'' category while varying the number of sampled points inside the bounding box that contains the target mesh. As expected, increasing the number of sampled points results in an improvement in reconstruction quality. We empirically found that sampling $10$k points results in the best compromise between training time and reconstruction performance.
\begin{table}
    \centering
    \def\arraystretch{0.95}
    \begin{subtable}{1\columnwidth}
    \centering
    \begin{tabular}{lcc}
        \toprule
        & IoU & Chamfer-$L_1$ \\
        \midrule
        Uniform & 0.383 & 0.073\\
        Biased & 0.351  & 0.041\\
        Importance & \bf{0.491} & 0.073\\
        \bottomrule
    \end{tabular}
    \caption{Influence of sampling strategy}
    \end{subtable}
    \def\arraystretch{0.95}
    \begin{subtable}{1\columnwidth}
    \centering
    \begin{tabular}{lcc}
        \toprule
        & IoU & Chamfer-$L_1$ \\
        \midrule
        Importance 2k & 0.370  & 0.074\\
        Importance 5k & 0.380  & 0.076\\
        Importance 10k & \bf{0.491} & \bf{0.073}\\
                \bottomrule
    \end{tabular}
    \caption{Influence of number of sampled points.}
    \end{subtable}
    \caption{{\bf Sampling Strategy.} We evaluate the performance of our model while varying the sampling scheme and the number of the sampled points inside the bounding box of the target mesh. We report the volumetric IoU (higher is better) and the Chamfer distance (lower is better) on the test set of the "chair category".}
    \label{tab:sampling}
\end{table}

\subsubsection{Impact of Proximity loss}

In this section, we explain empirically the vanishing gradient problem that emerges from the use of the sigmoid in the occupancy function of \eqref{eq:implicit_sq_actual}.
To this end, we train two variants of our model, one with and without the proximity loss of Eq. 15, in our main submission. For this experiment, we train both variants on D-FAUST for the single image 3D reconstruction task. Both models are trained for a maximum number of $32$ primitives and $s=10$ and for the same number of iterations.
\begin{figure}[h!]
	\centering
	\begin{subfigure}[b]{0.15\linewidth}
    	\centering
        \includegraphics[width=0.5\linewidth]{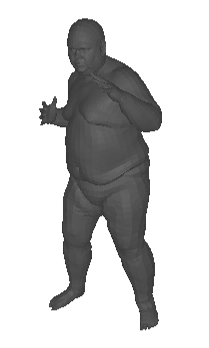}
    \end{subfigure}
    \hfill
    \begin{subfigure}[b]{0.15\linewidth}
    	\centering
        \includegraphics[width=\linewidth]{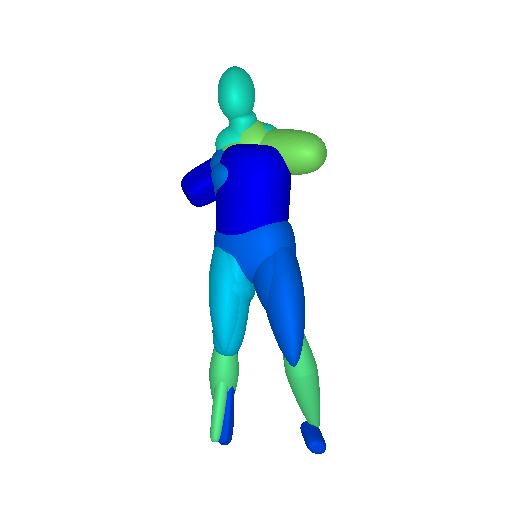}
    \end{subfigure}
    \hfill
    \begin{subfigure}[b]{0.15\linewidth}
    	\centering
        \includegraphics[width=\linewidth]{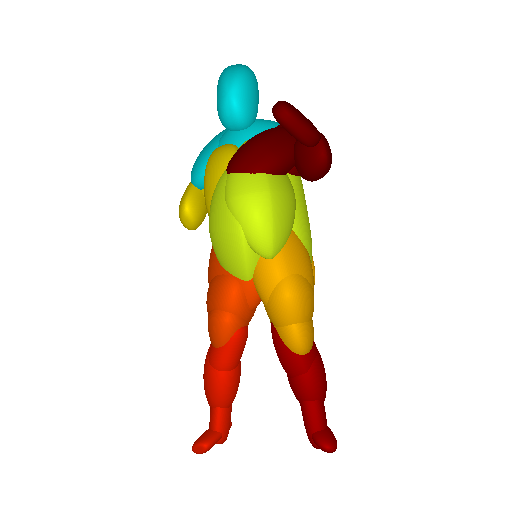}
    \end{subfigure}
    \hfill
    \begin{subfigure}[b]{0.15\linewidth}
    	\centering
        \includegraphics[width=0.76\linewidth]{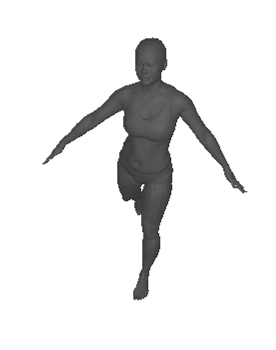}
    \end{subfigure}
    \hfill
    \begin{subfigure}[b]{0.15\linewidth}
    	\centering
        \includegraphics[width=\linewidth]{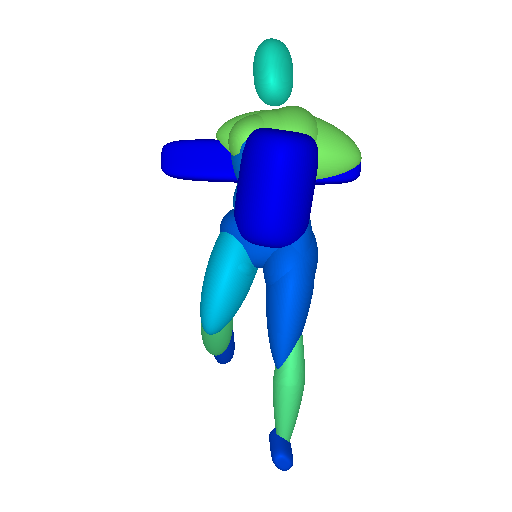}
    \end{subfigure}
    \hfill
    \begin{subfigure}[b]{0.15\linewidth}
    	\centering
        \includegraphics[width=\linewidth]{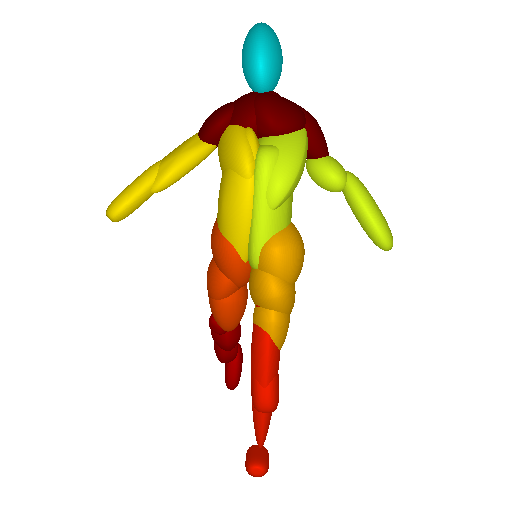}
    \end{subfigure}
    \vskip\baselineskip	\begin{subfigure}[b]{0.15\linewidth}
    	\centering
        \includegraphics[width=0.5\linewidth]{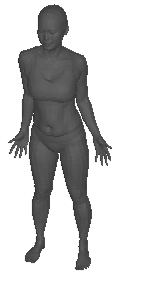}
    \end{subfigure}
    \hfill
    \begin{subfigure}[b]{0.15\linewidth}
    	\centering
        \includegraphics[width=1.0\linewidth]{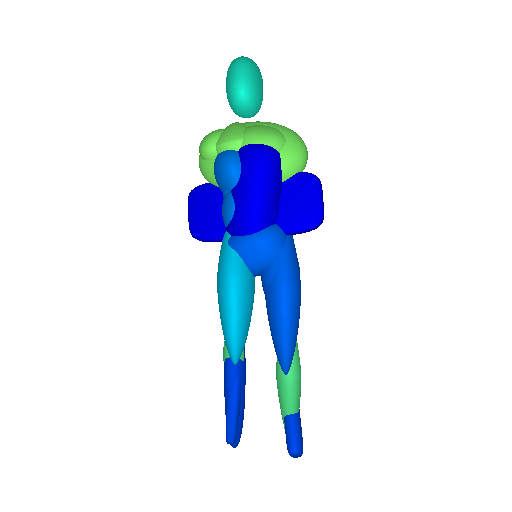}
    \end{subfigure}
    \hfill
    \begin{subfigure}[b]{0.15\linewidth}
    	\centering
        \includegraphics[width=1.0\linewidth]{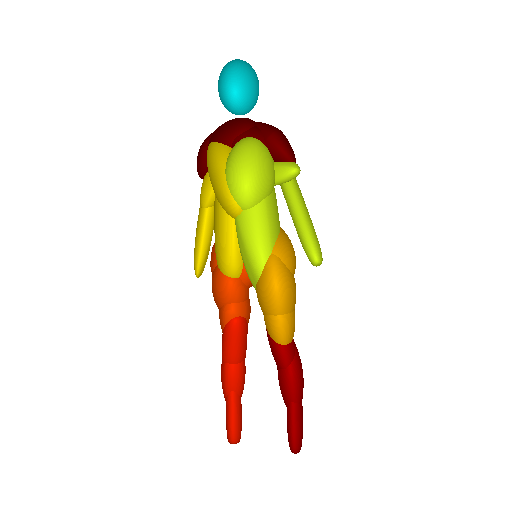}
    \end{subfigure}
    \hfill
    \begin{subfigure}[b]{0.15\linewidth}
    	\centering
        \includegraphics[width=0.7\linewidth]{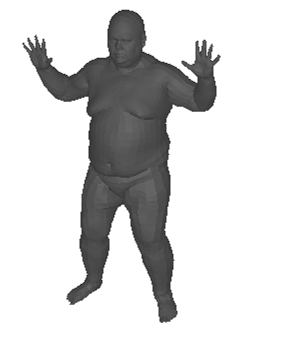}
    \end{subfigure}
    \hfill
    \begin{subfigure}[b]{0.15\linewidth}
    	\centering
        \includegraphics[width=\linewidth]{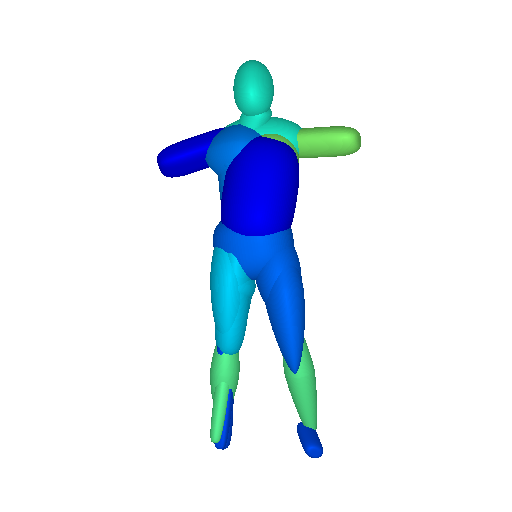}
    \end{subfigure}
    \hfill
    \begin{subfigure}[b]{0.15\linewidth}
    	\centering
        \includegraphics[width=\linewidth]{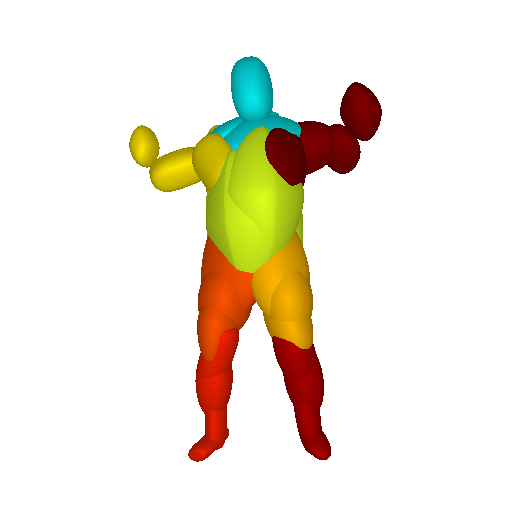}
    \end{subfigure}
    \vskip\baselineskip    \vspace{-1.2em}
    \begin{subfigure}[b]{0.15\linewidth}
		\centering
        \caption{\textbf{Input}}
    \end{subfigure}
    \hfill
    \begin{subfigure}[b]{0.15\linewidth}
		\centering
        \caption{\bf{without}}
    \end{subfigure}
    \hfill
    \begin{subfigure}[b]{0.15\linewidth}
		\centering
        \caption{\textbf{Ours}}
    \end{subfigure}
    \hfill
    \begin{subfigure}[b]{0.15\linewidth}
		\centering
        \caption{\textbf{Input}}
    \end{subfigure}
    \hfill
    \begin{subfigure}[b]{0.15\linewidth}
		\centering
        \caption{\bf{without}}
    \end{subfigure}
    \hfill
    \begin{subfigure}[b]{0.15\linewidth}
		\centering
        \caption{\textbf{Ours}}
    \end{subfigure}
    \caption{\textbf{Vanishing gradients.} We visualize the predictions of two two models, one trained with ({\bf{Ours}}) and one {\bf{without}} the proximity loss term. On the left, we visualize the input RGB image (a, d), in the middle the predictions without the proximity loss (b,c) and on the right the predictions of our model with this additional loss term.}
    \label{fig:vanishing_gradients_problem}
\end{figure}

\figref{fig:vanishing_gradients_problem} illustrates the predictions of both variants. We remark that the predictions of the model that was trained without the proximity loss are less accurate. Note that due to the vanishing gradient problem, the network is not able to properly "move" primitives and as a result, instead of reconstructing the hands of the humans using two or four primitives, the network uses only one. Interestingly, the reconstructions in some cases \eg (e) do not even capture the human shape properly. However, even though the reconstruction quality is bad, the network is not able to fix it because the gradients of the reconstruction loss are small (even though the reconstruction loss itself is high). This is also validated quantitatively, as can be observed from Table \ref{tab:quantitative_proximity}.

\def\arraystretch{0.95}
\begin{table}
    \centering
    \begin{tabular}{lcc}
        \toprule
        & IoU & Chamfer-$L_1$ \\
        \midrule
        Ours w/o proximity loss & 0.605 & 0.171\\
        Ours & \bf{0.699} & \bf{0.098}\\
        \bottomrule
    \end{tabular}
    \caption{{\bf Proximity loss.} We investigate the impact of the proximity loss. We report the volumetric IoU and the Chamfer distance for two variants of our model, one with and without the proximity loss term.}
    \label{tab:quantitative_proximity}
    \vspace{-1em}
\end{table}

\newpage
\section{Additional Results on ShapeNet}
In this section, we provide additional qualitative results on various object types from the ShapeNet dataset \cite{Chang2015ARXIV}. Furthermore, we also demonstrate the ability of our model to predict semantic hierarchies, where the same node is used for representing the same part of the object. We compare our model qualitatively with \cite{Paschalidou2019CVPR}. In particular, we train both models on the single-view 3D reconstruction task, using the same image renderings and train/test splits as \cite{Choy2016ECCV}. Both methods are trained for a maximum number of $64$ primitives. For our method, we empirically observed that a sharpness value $s=10$ led to good reconstructions.
Note that we do not compare qualitatively with \cite{Genova2019ICCV, Deng2019ARXIV} as they do not provide code.
Finally, we also compare our model with \cite{Tulsiani2017CVPRa, Paschalidou2019CVPR} on the volumetric reconstruction task, where the input to all networks is a binary voxel grid. For a fair comparison, all models leverage the same feature encoder architecture proposed in \cite{Paschalidou2019CVPR}.

In \figref{fig:shapenet_supp}$+$\ref{fig:shapenet_supp_2}, we qualitatively compare our predictions with \cite{Paschalidou2019CVPR} for various ShapeNet objects. We observe that our model yields more accurate reconstructions compared to our baseline. Due to the use of the reconstruction quality $q_k^d$, our model dynamically decides whether a node should be split or not. For example, our model represents the phone in \figref{fig:shapenet_supp} (a) using one primitive (root node) and the phone in \figref{fig:shapenet_supp} (b), that consists of two parts, with two primitives. This can be also noted for the case of the displays \figref{fig:shapenet_supp} (g$+$j). For more complicated objects, such as aeroplanes, tables and chairs, our network uses more primitives to accurately capture the geometry of the target object.
Note that for this experiment we set the threshold for $q_k^d$ to $0.8$.

Our network associates the same node with the same part of the object, as it can be seen from the predicted hierarchies in \figref{fig:shapenet_supp}$+$\ref{fig:shapenet_supp_2}. For example, for the displays the second primitive at the first depth level is used for representing the monitor of the display, for the aeroplanes the $4$-th primitive in the second depth level is used for representing the front part of the aeroplanes.

\subsection{Volumetric Reconstruction}

\begin{figure}[h!]
    \centering
    \begin{subfigure}[b]{0.12\linewidth}
		\centering
		\includegraphics[width=\linewidth]{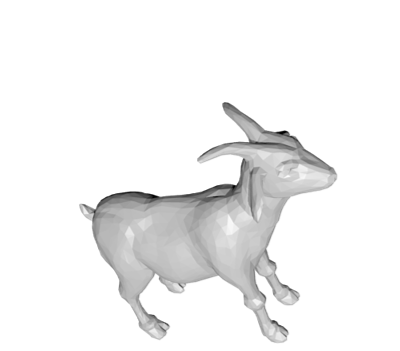}
    \end{subfigure}
    \hfill
    \begin{subfigure}[b]{0.12\linewidth}
		\centering
		\includegraphics[width=\linewidth]{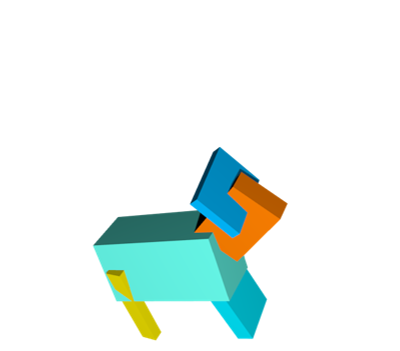}
    \end{subfigure}
    \hfill
    \begin{subfigure}[b]{0.12\linewidth}
		\centering
		\includegraphics[width=\linewidth]{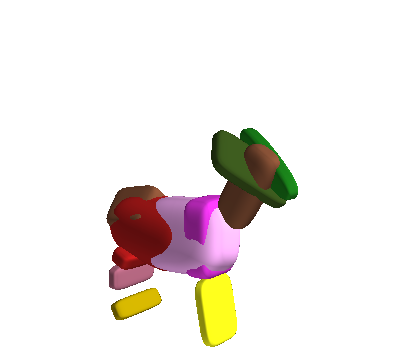}
    \end{subfigure}
    \hfill
    \begin{subfigure}[b]{0.12\linewidth}
		\centering
		\includegraphics[width=0.7\linewidth]{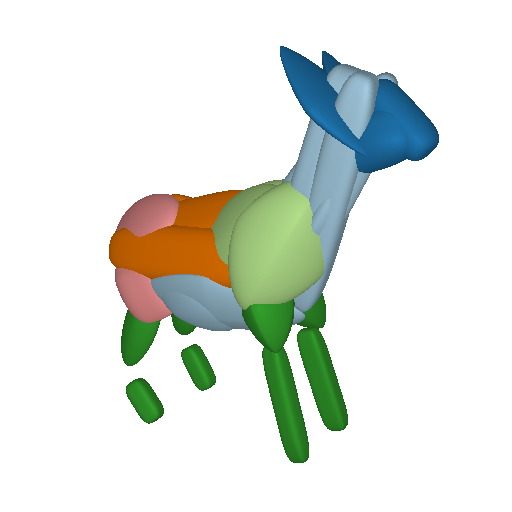}
    \end{subfigure}
    \begin{subfigure}[b]{0.12\linewidth}
		\centering
		\includegraphics[width=\linewidth]{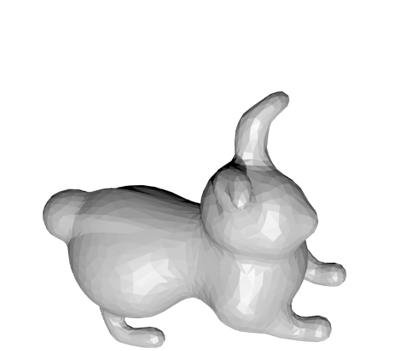}
    \end{subfigure}
    \hfill
    \begin{subfigure}[b]{0.12\linewidth}
		\centering
		\includegraphics[width=0.7\linewidth]{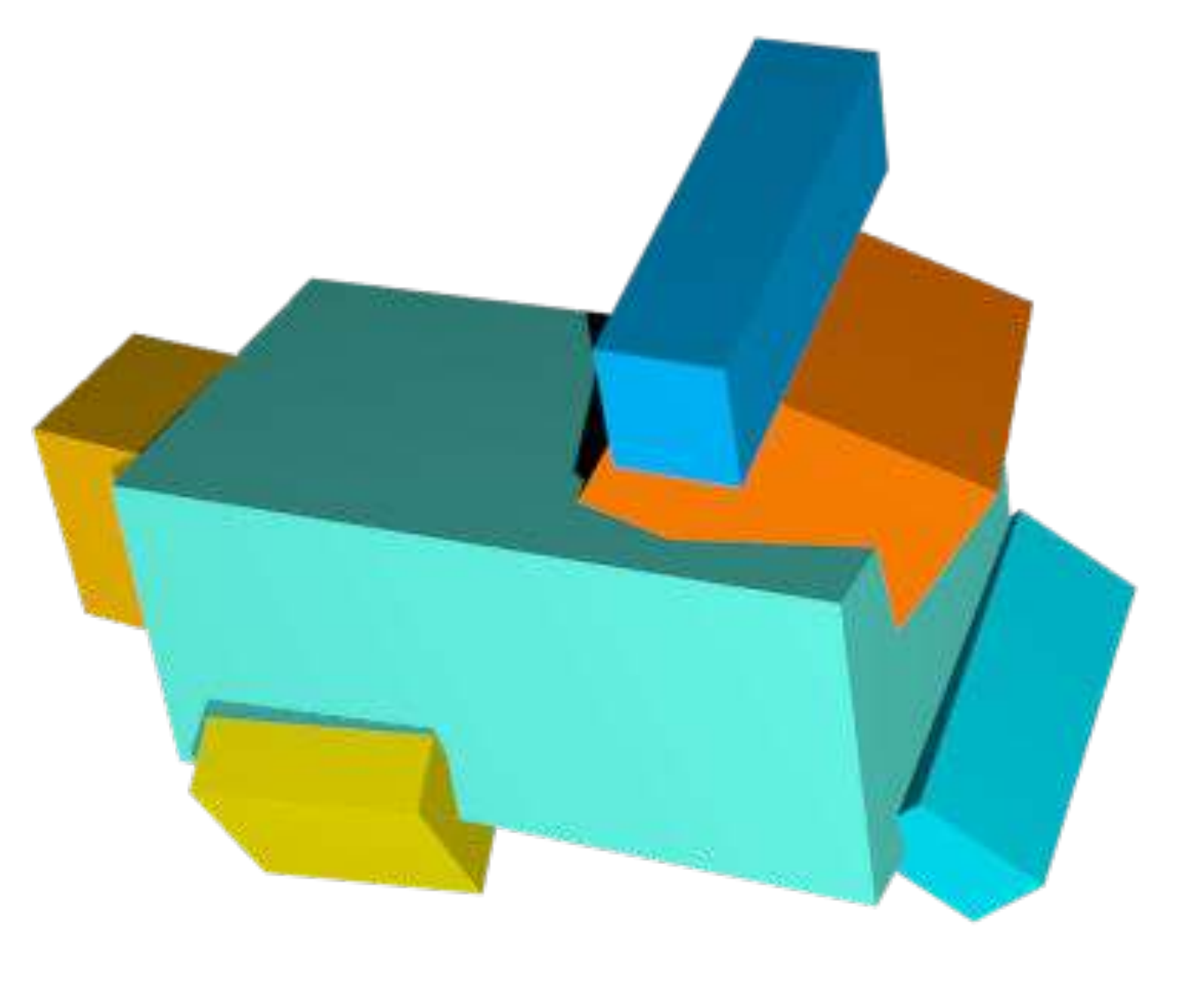}
    \end{subfigure}
    \hfill
    \begin{subfigure}[b]{0.12\linewidth}
		\centering
		\includegraphics[width=0.7\linewidth]{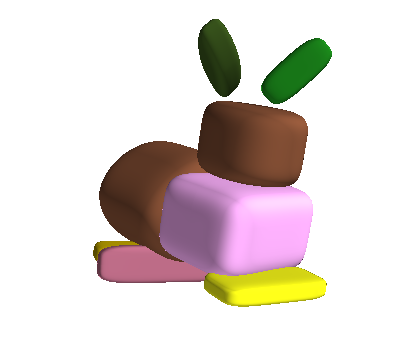}
    \end{subfigure}
    \hfill
    \begin{subfigure}[b]{0.12\linewidth}
		\centering
		\includegraphics[width=0.8\linewidth]{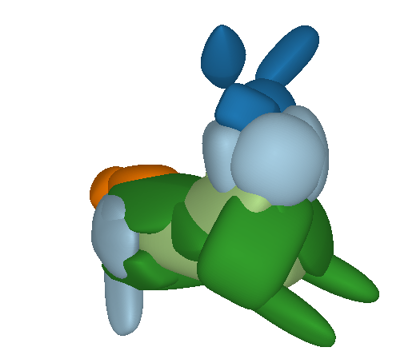}
    \end{subfigure}
    \vskip\baselineskip    \vspace{-1.2em}
    \begin{subfigure}[b]{0.12\linewidth}
		\centering
        \caption{\textbf{Input}}
    \end{subfigure}
    \hfill
    \begin{subfigure}[b]{0.12\linewidth}
		\centering
        \caption{\cite{Tulsiani2017CVPRa}}
    \end{subfigure}
    \hfill
    \begin{subfigure}[b]{0.12\linewidth}
		\centering
        \caption{\cite{Paschalidou2019CVPR}}
    \end{subfigure}
    \hfill
    \begin{subfigure}[b]{0.12\linewidth}
		\centering
        \caption{\textbf{Ours}}
    \end{subfigure}
    \hfill
    \begin{subfigure}[b]{0.12\linewidth}
		\centering
        \caption{\textbf{Input}}
    \end{subfigure}
    \hfill
    \begin{subfigure}[b]{0.12\linewidth}
		\centering
        \caption{\cite{Tulsiani2017CVPRa}}
    \end{subfigure}
    \hfill
    \begin{subfigure}[b]{0.12\linewidth}
		\centering
        \caption{\cite{Paschalidou2019CVPR}}
    \end{subfigure}
    \hfill
    \begin{subfigure}[b]{0.12\linewidth}
		\centering
        \caption{\textbf{Ours}}
    \end{subfigure}
    \begin{subfigure}[b]{0.49\linewidth}
    	\centering
        \includegraphics[width=\linewidth]{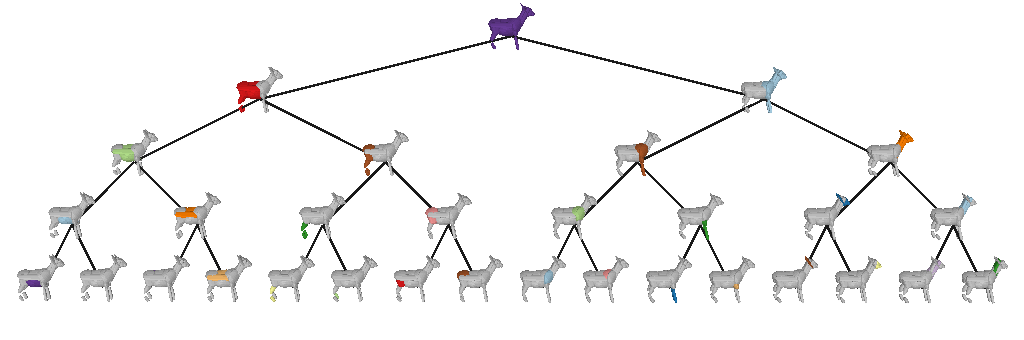}
    \end{subfigure}
    \hfill
    \begin{subfigure}[b]{0.49\linewidth}
    	\centering
        \includegraphics[width=\linewidth]{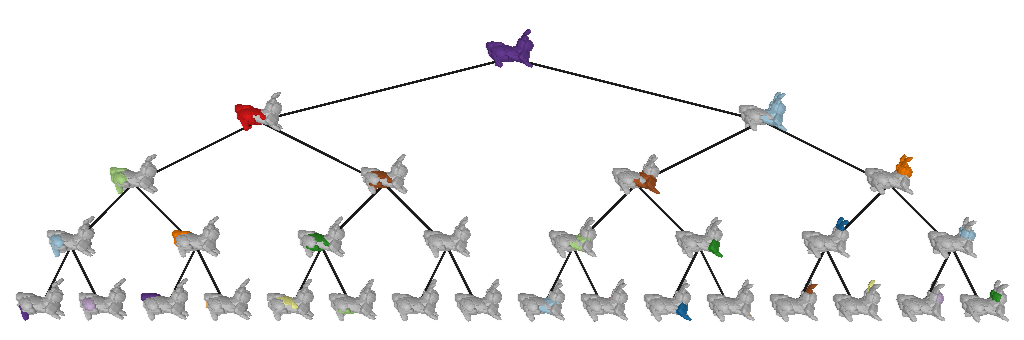}
    \end{subfigure}
    \vskip\baselineskip    \vspace{-1.2em}
    \begin{subfigure}[b]{0.49\linewidth}
		\centering
        \caption{\textbf{Predicted Hierarchy}}
    \end{subfigure}
    \hfill
    \begin{subfigure}[b]{0.49\linewidth}
		\centering
        \caption{\textbf{Predicted Hierarchy}}
    \end{subfigure}
    \vskip\baselineskip
    \begin{subfigure}[b]{0.12\linewidth}
		\centering
		\includegraphics[width=\linewidth]{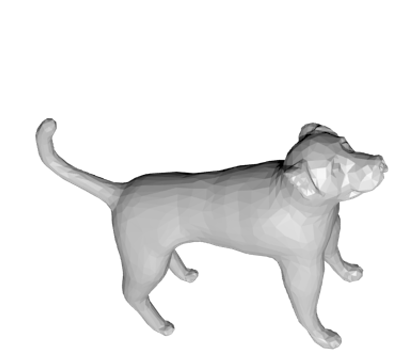}
    \end{subfigure}
    \hfill
    \begin{subfigure}[b]{0.12\linewidth}
		\centering
		\includegraphics[width=0.8\linewidth]{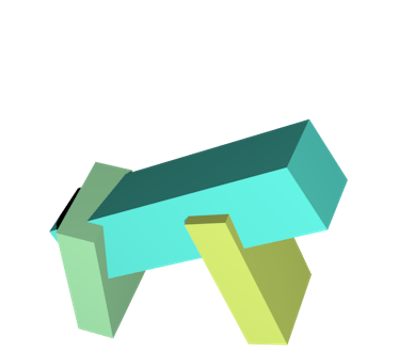}
    \end{subfigure}
    \hfill
    \begin{subfigure}[b]{0.12\linewidth}
		\centering
		\includegraphics[width=0.7\linewidth]{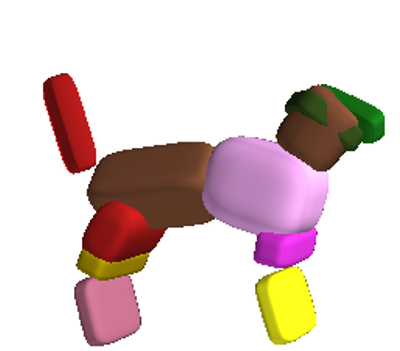}
    \end{subfigure}
    \hfill
    \begin{subfigure}[b]{0.12\linewidth}
		\centering
		\includegraphics[width=0.7\linewidth]{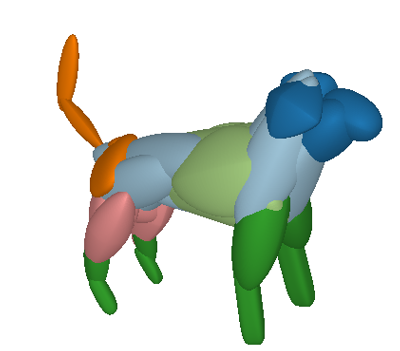}
    \end{subfigure}
    \begin{subfigure}[b]{0.12\linewidth}
		\centering
		\includegraphics[width=\linewidth]{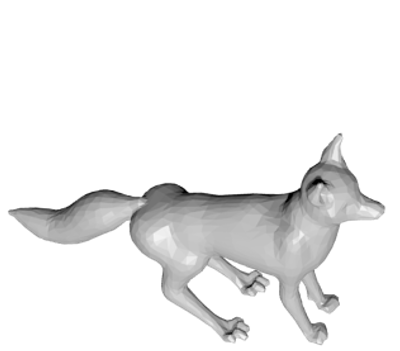}
    \end{subfigure}
    \hfill
    \begin{subfigure}[b]{0.12\linewidth}
		\centering
		\includegraphics[width=0.8\linewidth]{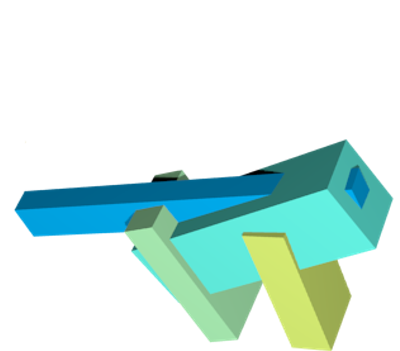}
    \end{subfigure}
    \hfill
    \begin{subfigure}[b]{0.12\linewidth}
		\centering
		\includegraphics[width=0.8\linewidth]{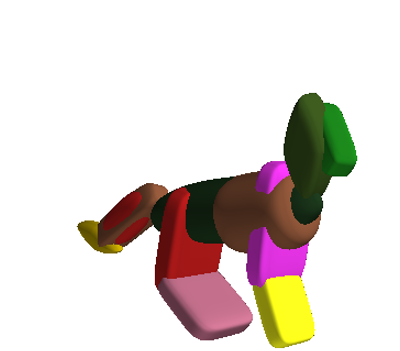}
    \end{subfigure}
    \hfill
    \begin{subfigure}[b]{0.12\linewidth}
		\centering
		\includegraphics[width=0.8\linewidth]{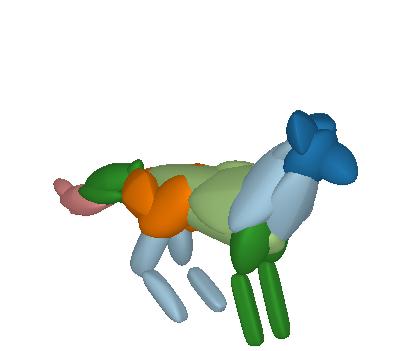}
    \end{subfigure}
    \vskip\baselineskip    \vspace{-1.2em}
    \begin{subfigure}[b]{0.12\linewidth}
		\centering
        \caption{\textbf{Input}}
    \end{subfigure}
    \hfill
    \begin{subfigure}[b]{0.12\linewidth}
		\centering
        \caption{\cite{Tulsiani2017CVPRa}}
    \end{subfigure}
    \hfill
    \begin{subfigure}[b]{0.12\linewidth}
		\centering
        \caption{\cite{Paschalidou2019CVPR}}
    \end{subfigure}
    \hfill
    \begin{subfigure}[b]{0.12\linewidth}
		\centering
        \caption{\textbf{Ours}}
    \end{subfigure}
    \hfill
    \begin{subfigure}[b]{0.12\linewidth}
		\centering
        \caption{\textbf{Input}}
    \end{subfigure}
    \hfill
    \begin{subfigure}[b]{0.12\linewidth}
		\centering
        \caption{\cite{Tulsiani2017CVPRa}}
    \end{subfigure}
    \hfill
    \begin{subfigure}[b]{0.12\linewidth}
		\centering
        \caption{\cite{Paschalidou2019CVPR}}
    \end{subfigure}
    \hfill
    \begin{subfigure}[b]{0.12\linewidth}
		\centering
        \caption{\textbf{Ours}}
    \end{subfigure}
    \begin{subfigure}[b]{0.49\linewidth}
    	\centering
        \includegraphics[width=\linewidth]{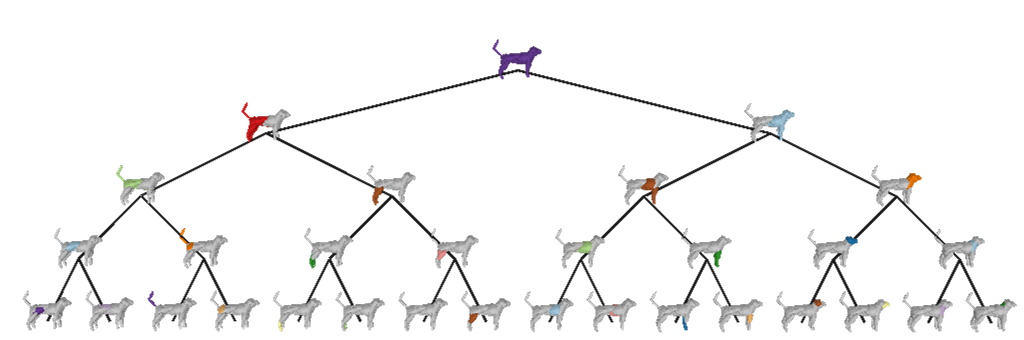}
    \end{subfigure}
    \hfill
    \begin{subfigure}[b]{0.49\linewidth}
    	\centering
        \includegraphics[width=\linewidth]{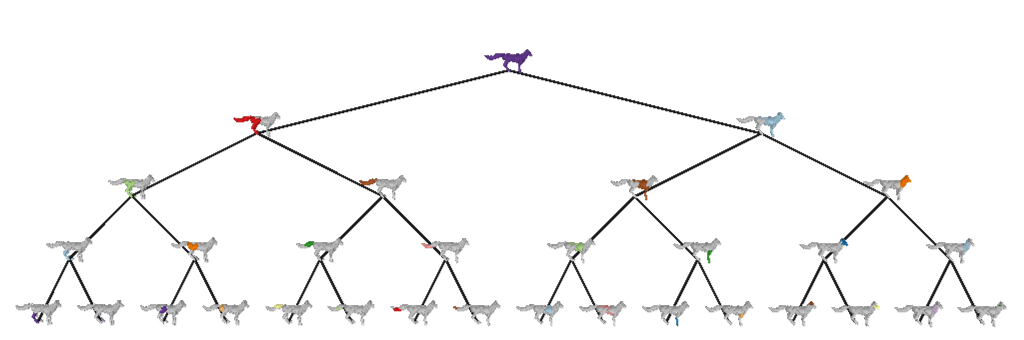}
    \end{subfigure}
    \vskip\baselineskip    \vspace{-1.2em}
    \begin{subfigure}[b]{0.49\linewidth}
		\centering
        \caption{\textbf{Predicted Hierarchy}}
    \end{subfigure}
    \hfill
    \begin{subfigure}[b]{0.49\linewidth}
		\centering
        \caption{\textbf{Predicted Hierarchy}}
    \end{subfigure}
    \vskip\baselineskip    \caption{\textbf{Volumetric Reconstruction.} We note that our reconstructions are geometrically more accurate. In contrast to \cite{Paschalidou2019CVPR}, our model yields reconstructions where the legs of the animals are not connected. Furthermore, our model accurately captures the ears and tails of the different animals.
    }
    \label{fig:volumetric}
    \vspace{-0.2em}
\end{figure}

\begin{figure}[h!]
	\centering
	\begin{subfigure}[b]{0.15\linewidth}
    	\centering
        \includegraphics[width=\linewidth]{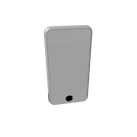}
    \end{subfigure}
    \hfill
    \begin{subfigure}[b]{0.15\linewidth}
    	\centering
        \includegraphics[width=\linewidth]{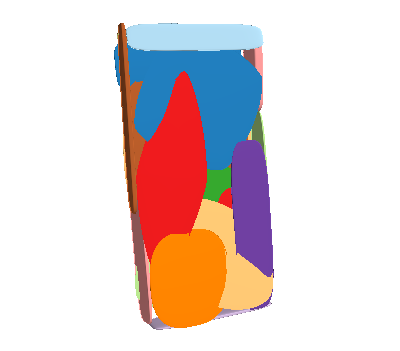}
    \end{subfigure}
    \hfill
    \begin{subfigure}[b]{0.15\linewidth}
    	\centering
        \includegraphics[width=\linewidth]{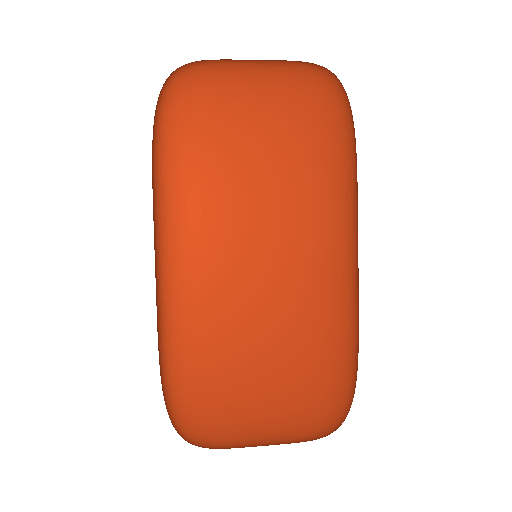}
    \end{subfigure}
    \hfill
    \begin{subfigure}[b]{0.15\linewidth}
    	\centering
        \includegraphics[width=\linewidth]{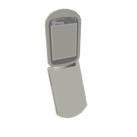}
    \end{subfigure}
    \hfill
    \begin{subfigure}[b]{0.15\linewidth}
    	\centering
        \includegraphics[width=\linewidth]{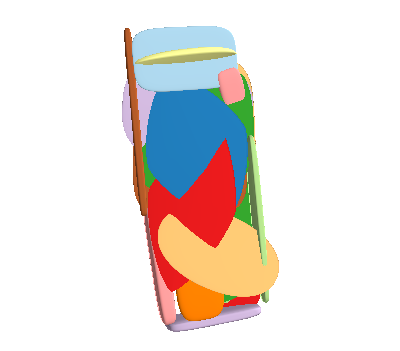}
    \end{subfigure}
    \hfill
    \begin{subfigure}[b]{0.15\linewidth}
    	\centering
        \includegraphics[width=\linewidth]{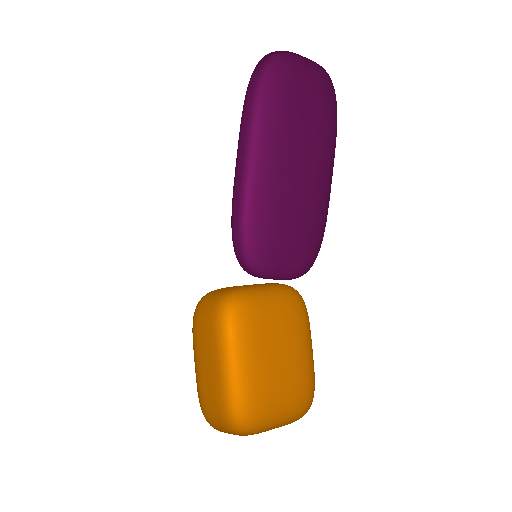}
    \end{subfigure}

    \vskip\baselineskip    \vspace{-1.2em}
    \begin{subfigure}[b]{0.15\linewidth}
		\centering
        \caption{\textbf{Input}}
    \end{subfigure}
    \hfill
    \begin{subfigure}[b]{0.15\linewidth}
		\centering
        \caption{\bf{SQs}\cite{Paschalidou2019CVPR}}
    \end{subfigure}
    \hfill
    \begin{subfigure}[b]{0.15\linewidth}
		\centering
        \caption{\textbf{Ours}}
    \end{subfigure}
    \hfill
    \begin{subfigure}[b]{0.15\linewidth}
		\centering
        \caption{\textbf{Input}}
    \end{subfigure}
    \hfill
    \begin{subfigure}[b]{0.15\linewidth}
		\centering
        \caption{\bf{SQs}\cite{Paschalidou2019CVPR}}
    \end{subfigure}
    \hfill
    \begin{subfigure}[b]{0.15\linewidth}
		\centering
        \caption{\textbf{Ours}}
    \end{subfigure}
    \vskip\baselineskip
	\begin{subfigure}[b]{0.15\linewidth}
    	\centering
        \includegraphics[width=\linewidth]{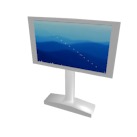}
    \end{subfigure}
    \hfill
    \begin{subfigure}[b]{0.15\linewidth}
    	\centering
        \includegraphics[width=\linewidth]{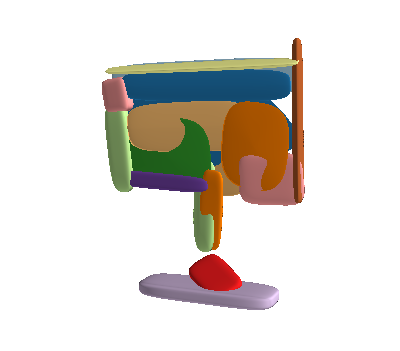}
    \end{subfigure}
    \hfill
    \begin{subfigure}[b]{0.15\linewidth}
    	\centering
        \includegraphics[width=\linewidth]{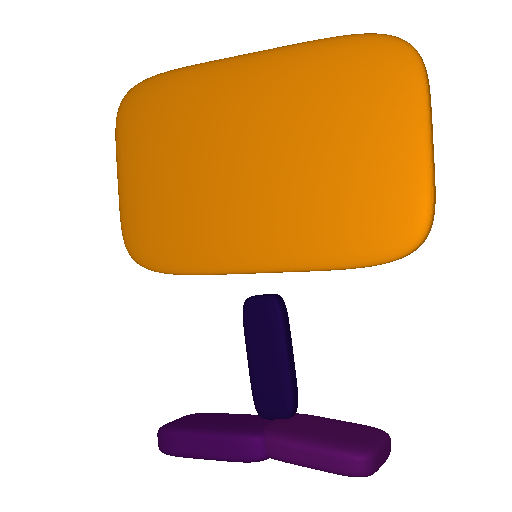}
    \end{subfigure}
    \hfill
    \begin{subfigure}[b]{0.15\linewidth}
    	\centering
        \includegraphics[width=\linewidth]{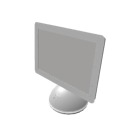}
    \end{subfigure}
    \hfill
    \begin{subfigure}[b]{0.15\linewidth}
    	\centering
        \includegraphics[width=\linewidth]{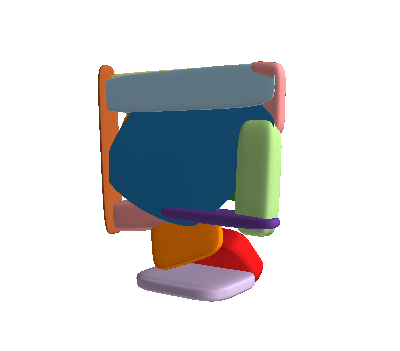}
    \end{subfigure}
    \hfill
    \begin{subfigure}[b]{0.15\linewidth}
    	\centering
        \includegraphics[width=\linewidth]{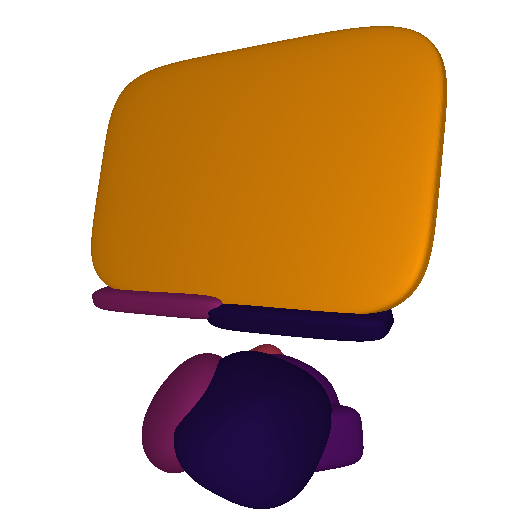}
    \end{subfigure}

    \vskip\baselineskip    \vspace{-1.2em}
    \begin{subfigure}[b]{0.15\linewidth}
		\centering
        \caption{\textbf{Input}}
    \end{subfigure}
    \hfill
    \begin{subfigure}[b]{0.15\linewidth}
		\centering
        \caption{\bf{SQs}\cite{Paschalidou2019CVPR}}
    \end{subfigure}
    \hfill
    \begin{subfigure}[b]{0.15\linewidth}
		\centering
        \caption{\textbf{Ours}}
    \end{subfigure}
    \hfill
    \begin{subfigure}[b]{0.15\linewidth}
		\centering
        \caption{\textbf{Input}}
    \end{subfigure}
    \hfill
    \begin{subfigure}[b]{0.15\linewidth}
		\centering
        \caption{\bf{SQs}\cite{Paschalidou2019CVPR}}
    \end{subfigure}
    \hfill
    \begin{subfigure}[b]{0.15\linewidth}
		\centering
        \caption{\textbf{Ours}}
    \end{subfigure}
    \vskip\baselineskip
    \begin{subfigure}[b]{0.49\linewidth}
    	\centering
        \includegraphics[width=1.0\linewidth]{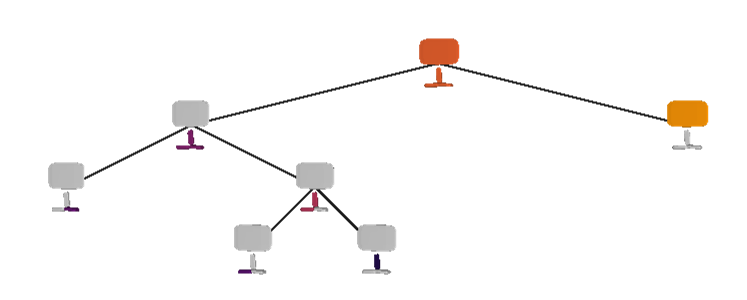}
    \end{subfigure}
    \hfill
    \begin{subfigure}[b]{0.49\linewidth}
    	\centering
        \includegraphics[width=\linewidth]{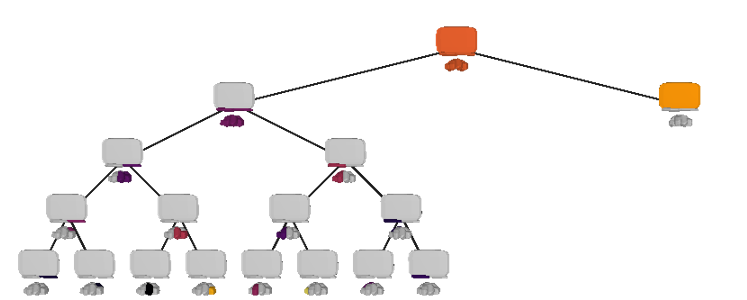}
    \end{subfigure}
    \vskip\baselineskip    \vspace{-1.2em}
    \begin{subfigure}[b]{0.49\linewidth}
		\centering
        \caption{\textbf{Predicted Hierarchy}}
    \end{subfigure}
    \hfill
    \begin{subfigure}[b]{0.49\linewidth}
		\centering
        \caption{\textbf{Predicted Hierarchy}}
    \end{subfigure}

	\begin{subfigure}[b]{0.15\linewidth}
    	\centering
        \includegraphics[width=\linewidth]{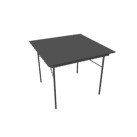}
    \end{subfigure}
    \hfill
    \begin{subfigure}[b]{0.15\linewidth}
    	\centering
        \includegraphics[width=\linewidth]{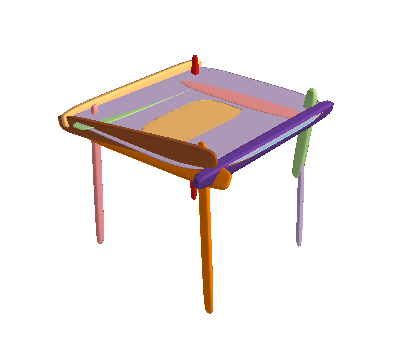}
    \end{subfigure}
    \hfill
    \begin{subfigure}[b]{0.15\linewidth}
    	\centering
        \includegraphics[width=\linewidth]{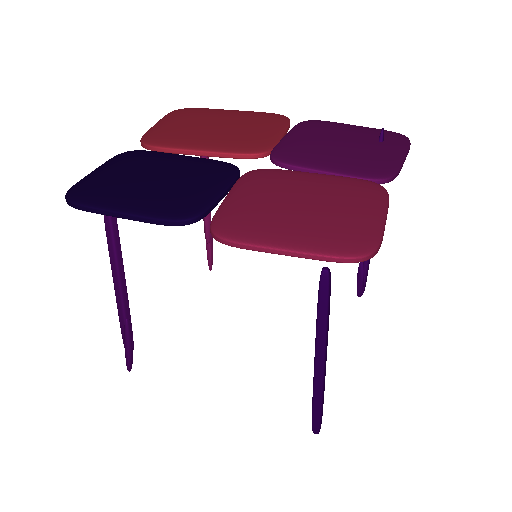}
    \end{subfigure}
    \hfill
    \begin{subfigure}[b]{0.15\linewidth}
    	\centering
        \includegraphics[width=\linewidth]{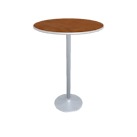}
    \end{subfigure}
    \hfill
    \begin{subfigure}[b]{0.15\linewidth}
    	\centering
        \includegraphics[width=\linewidth]{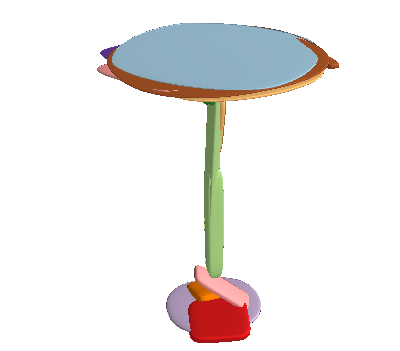}
    \end{subfigure}
    \hfill
    \begin{subfigure}[b]{0.15\linewidth}
    	\centering
        \includegraphics[width=\linewidth]{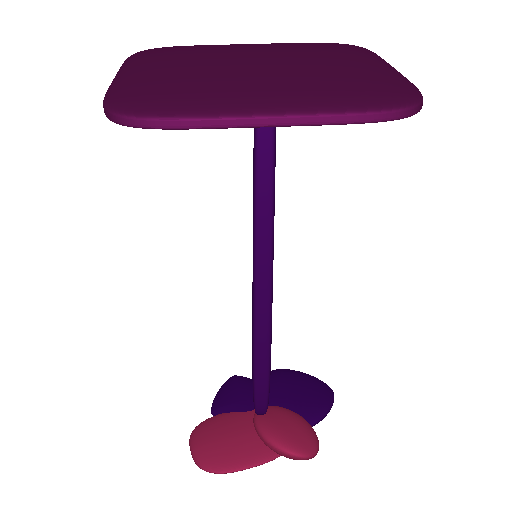}
    \end{subfigure}

    \vskip\baselineskip    \vspace{-1.2em}
    \begin{subfigure}[b]{0.15\linewidth}
		\centering
        \caption{\textbf{Input}}
    \end{subfigure}
    \hfill
    \begin{subfigure}[b]{0.15\linewidth}
		\centering
        \caption{\bf{SQs}\cite{Paschalidou2019CVPR}}
    \end{subfigure}
    \hfill
    \begin{subfigure}[b]{0.15\linewidth}
		\centering
        \caption{\textbf{Ours}}
    \end{subfigure}
    \hfill
    \begin{subfigure}[b]{0.15\linewidth}
		\centering
        \caption{\textbf{Input}}
    \end{subfigure}
    \hfill
    \begin{subfigure}[b]{0.15\linewidth}
		\centering
        \caption{\bf{SQs}\cite{Paschalidou2019CVPR}}
    \end{subfigure}
    \hfill
    \begin{subfigure}[b]{0.15\linewidth}
		\centering
        \caption{\textbf{Ours}}
    \end{subfigure}
    \vskip\baselineskip
    \begin{subfigure}[b]{0.49\linewidth}
    	\centering
        \includegraphics[width=\linewidth]{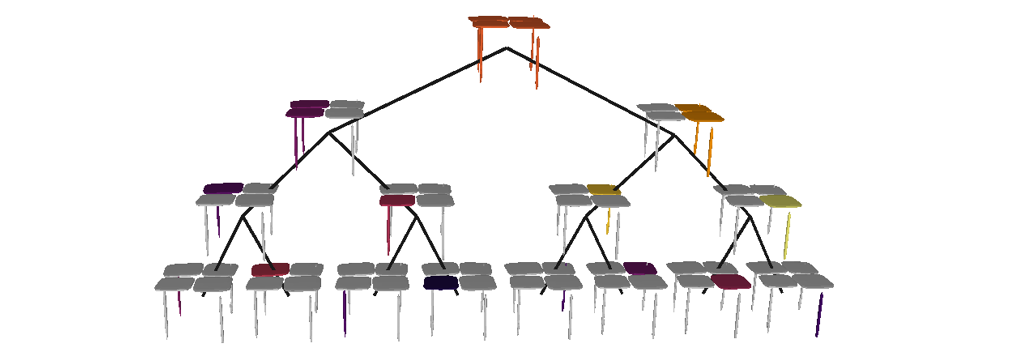}
    \end{subfigure}
    \hfill
    \begin{subfigure}[b]{0.49\linewidth}
    	\centering
        \includegraphics[width=\linewidth]{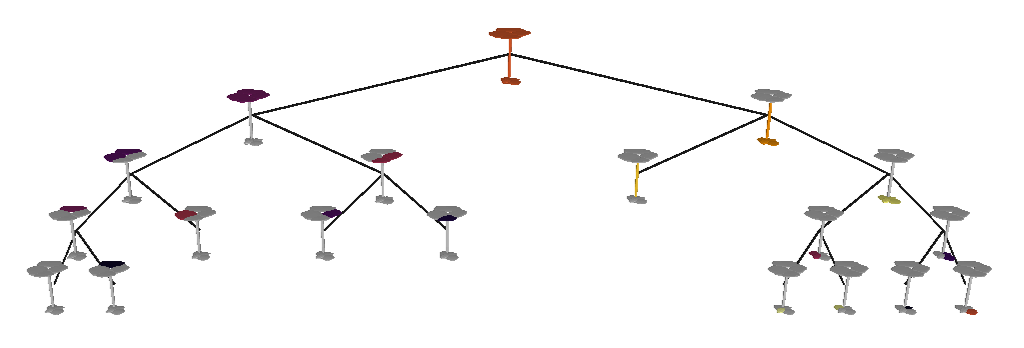}
    \end{subfigure}
    \vskip\baselineskip    \vspace{-1.2em}
    \begin{subfigure}[b]{0.49\linewidth}
		\centering
        \caption{\textbf{Predicted Hierarchy}}
    \end{subfigure}
    \hfill
    \begin{subfigure}[b]{0.49\linewidth}
		\centering
        \caption{\textbf{Predicted Hierarchy}}
    \end{subfigure}
    \caption{\textbf{Single Image 3D Reconstruction on ShapeNet.} We visualize the predictions of our model on various ShapeNet objects and compare to \cite{Paschalidou2019CVPR}. For objects that are represented with more than two primitives, we also visualize the predicted hierarchy.}
    \label{fig:shapenet_supp}
\end{figure}

\begin{figure}[h!]
	\centering
	\begin{subfigure}[b]{0.15\linewidth}
    	\centering
        \includegraphics[width=\linewidth]{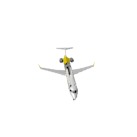}
    \end{subfigure}
    \hfill
    \begin{subfigure}[b]{0.15\linewidth}
    	\centering
        \includegraphics[width=\linewidth]{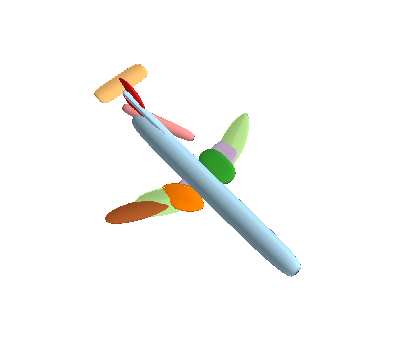}
    \end{subfigure}
    \hfill
    \begin{subfigure}[b]{0.15\linewidth}
    	\centering
        \includegraphics[width=\linewidth]{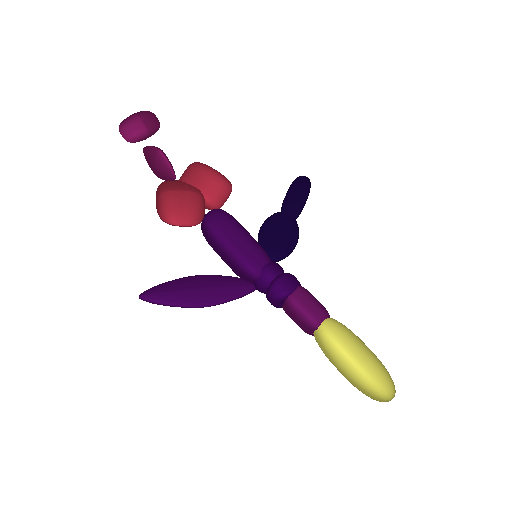}
    \end{subfigure}
    \hfill
    \begin{subfigure}[b]{0.15\linewidth}
    	\centering
        \includegraphics[width=\linewidth]{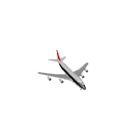}
    \end{subfigure}
    \hfill
    \begin{subfigure}[b]{0.15\linewidth}
    	\centering
        \includegraphics[width=\linewidth]{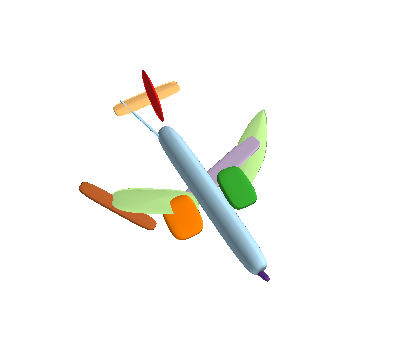}
    \end{subfigure}
    \hfill
    \begin{subfigure}[b]{0.15\linewidth}
    	\centering
        \includegraphics[width=\linewidth]{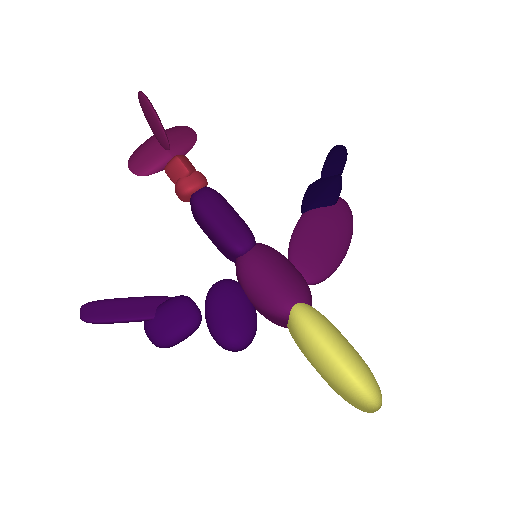}
    \end{subfigure}

    \vskip\baselineskip    \vspace{-1.2em}
    \begin{subfigure}[b]{0.15\linewidth}
		\centering
        \caption{\textbf{Input}}
    \end{subfigure}
    \hfill
    \begin{subfigure}[b]{0.15\linewidth}
		\centering
        \caption{\bf{SQs}\cite{Paschalidou2019CVPR}}
    \end{subfigure}
    \hfill
    \begin{subfigure}[b]{0.15\linewidth}
		\centering
        \caption{\textbf{Ours}}
    \end{subfigure}
    \hfill
    \begin{subfigure}[b]{0.15\linewidth}
		\centering
        \caption{\textbf{Input}}
    \end{subfigure}
    \hfill
    \begin{subfigure}[b]{0.15\linewidth}
		\centering
        \caption{\bf{SQs}\cite{Paschalidou2019CVPR}}
    \end{subfigure}
    \hfill
    \begin{subfigure}[b]{0.15\linewidth}
		\centering
        \caption{\textbf{Ours}}
    \end{subfigure}
    \vskip\baselineskip
    \begin{subfigure}[b]{0.49\linewidth}
    	\centering
        \includegraphics[width=\linewidth]{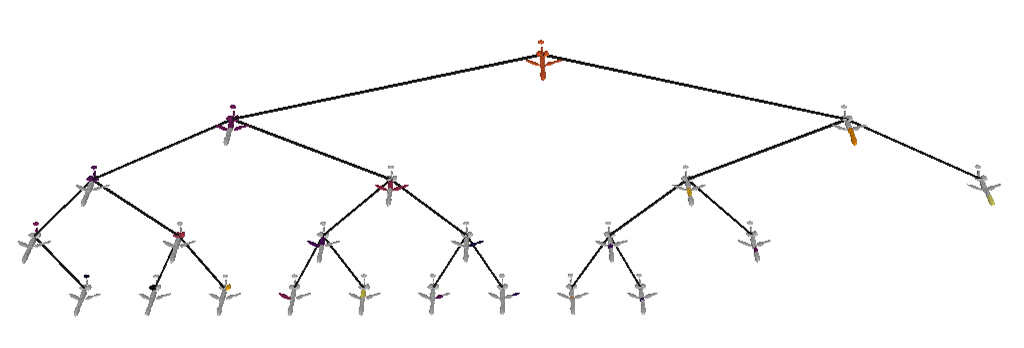}
    \end{subfigure}
    \hfill
    \begin{subfigure}[b]{0.49\linewidth}
    	\centering
        \includegraphics[width=\linewidth]{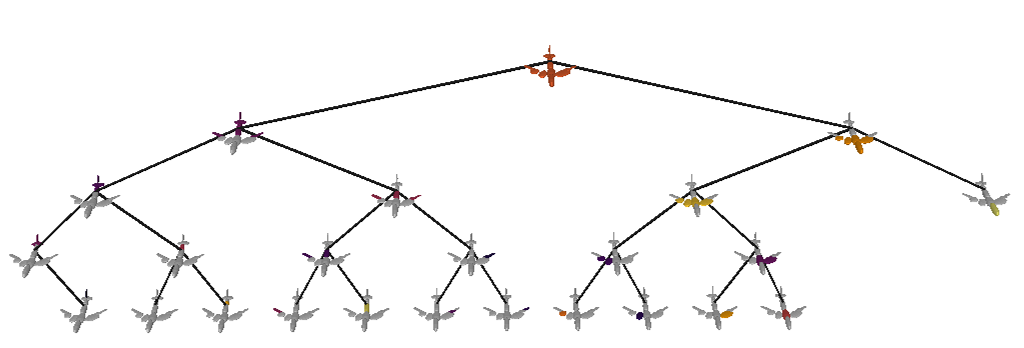}
    \end{subfigure}
    \vskip\baselineskip    \vspace{-1.2em}
    \begin{subfigure}[b]{0.49\linewidth}
		\centering
        \caption{\textbf{Predicted Hierarchy}}
    \end{subfigure}
    \hfill
    \begin{subfigure}[b]{0.49\linewidth}
		\centering
        \caption{\textbf{Predicted Hierarchy}}
    \end{subfigure}

	\begin{subfigure}[b]{0.15\linewidth}
    	\centering
        \includegraphics[width=\linewidth]{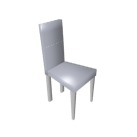}
    \end{subfigure}
    \hfill
    \begin{subfigure}[b]{0.15\linewidth}
    	\centering
        \includegraphics[width=\linewidth]{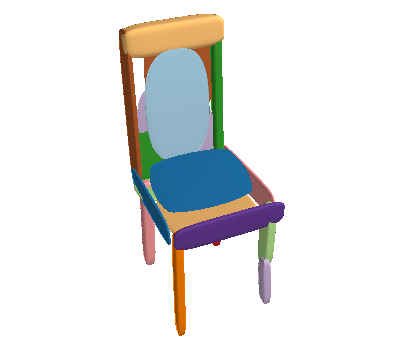}
    \end{subfigure}
    \hfill
    \begin{subfigure}[b]{0.15\linewidth}
    	\centering
        \includegraphics[width=\linewidth]{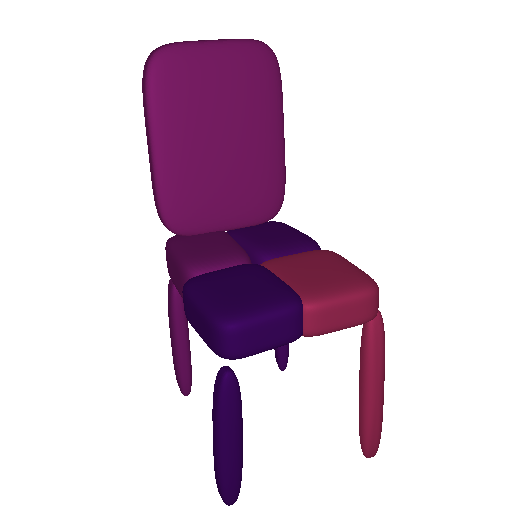}
    \end{subfigure}
    \hfill
    \begin{subfigure}[b]{0.15\linewidth}
    	\centering
        \includegraphics[width=\linewidth]{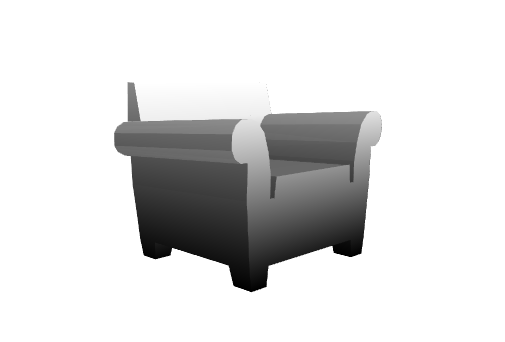}
    \end{subfigure}
    \hfill
    \begin{subfigure}[b]{0.15\linewidth}
    	\centering
        \includegraphics[width=\linewidth]{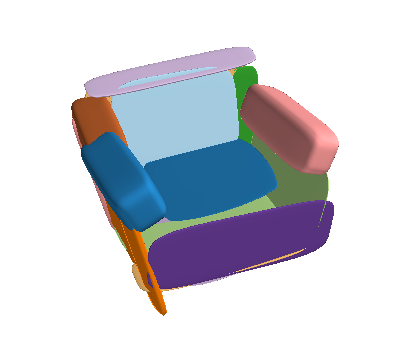}
    \end{subfigure}
    \hfill
    \begin{subfigure}[b]{0.15\linewidth}
    	\centering
        \includegraphics[width=\linewidth]{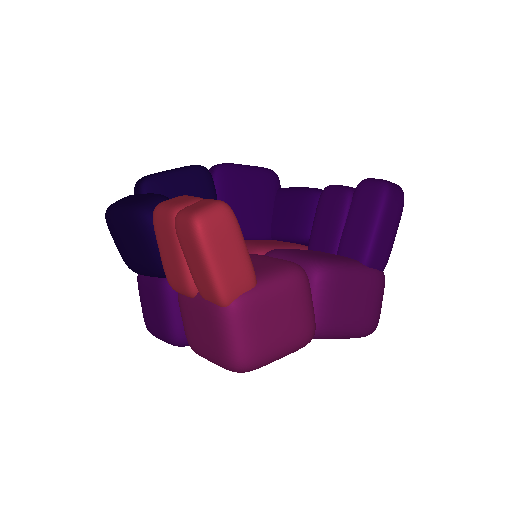}
    \end{subfigure}

    \vskip\baselineskip    \vspace{-1.2em}
    \begin{subfigure}[b]{0.15\linewidth}
		\centering
        \caption{\textbf{Input}}
    \end{subfigure}
    \hfill
    \begin{subfigure}[b]{0.15\linewidth}
		\centering
        \caption{\bf{SQs}\cite{Paschalidou2019CVPR}}
    \end{subfigure}
    \hfill
    \begin{subfigure}[b]{0.15\linewidth}
		\centering
        \caption{\textbf{Ours}}
    \end{subfigure}
    \hfill
    \begin{subfigure}[b]{0.15\linewidth}
		\centering
        \caption{\textbf{Input}}
    \end{subfigure}
    \hfill
    \begin{subfigure}[b]{0.15\linewidth}
		\centering
        \caption{\bf{SQs}\cite{Paschalidou2019CVPR}}
    \end{subfigure}
    \hfill
    \begin{subfigure}[b]{0.15\linewidth}
		\centering
        \caption{\textbf{Ours}}
    \end{subfigure}
    \vskip\baselineskip
    \begin{subfigure}[b]{0.49\linewidth}
    	\centering
        \includegraphics[width=\linewidth]{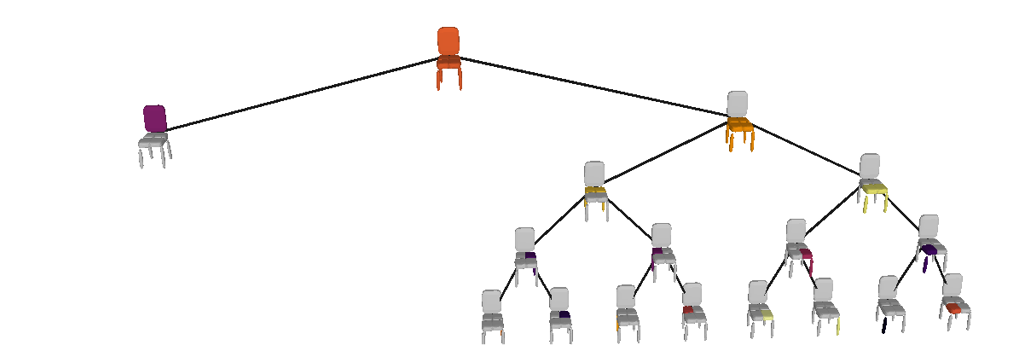}
    \end{subfigure}
    \hfill
    \begin{subfigure}[b]{0.49\linewidth}
    	\centering
        \includegraphics[width=\linewidth]{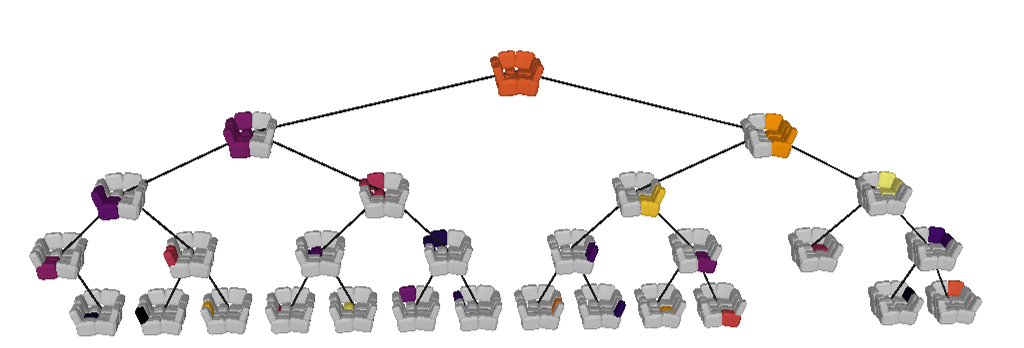}
    \end{subfigure}
    \vskip\baselineskip    \vspace{-1.2em}
    \begin{subfigure}[b]{0.49\linewidth}
		\centering
        \caption{\textbf{Predicted Hierarchy}}
    \end{subfigure}
    \hfill
    \begin{subfigure}[b]{0.49\linewidth}
		\centering
        \caption{\textbf{Predicted Hierarchy}}
    \end{subfigure}
    \caption{\textbf{Single Image 3D Reconstruction on ShapeNet.} We visualize the predictions of our model on various ShapeNet objects and compare to \cite{Paschalidou2019CVPR}. For objects that are represented with more than two primitives, we also visualize the predicted hierarchy.}
    \label{fig:shapenet_supp_2}
\end{figure}

Our model is closely related to the works of Tulsiani \etal \cite{Tulsiani2017CVPRa} and Paschalidou \etal \cite{Paschalidou2019CVPR}. Both \cite{Tulsiani2017CVPRa, Paschalidou2019CVPR} were originally introduced using a binary occupancy grid as an input to their model, thus we also compare our model with \cite{Tulsiani2017CVPRa, Paschalidou2019CVPR} using a voxelized input of size $32 \times 32 \times 32$. We evaluate the modelling accuracy of these three methods on the \emph{animal} class of the ShapeNet dataset. To ensure a fair comparison, we use the feature encoder proposed in \cite{Tulsiani2017CVPRa} for all three. A qualitative evaluation is provided in \figref{fig:volumetric}.

Our model yields more detailed reconstructions compared to \cite{Tulsiani2017CVPRa, Paschalidou2019CVPR}. For example, in our reconstructions the legs of the animals are not connected and the tails better capture the geometry of the target shape. Again, we observe that our network predicts semantic hierarchies, where the same node is used for representing the same part of the animal.

\newpage
\section{Additional Results on D-FAUST}

In this section, we provide additional qualitative results on the D-FAUST dataset \cite{Bogo2017CVPR}. Furthermore, we also demonstrate that the learned hierarchies are indeed semantic as the same node is used to represent the same part of the human body. Similar to the experiment of Section 4.2 in our main submission, we evaluate our model on the single-view 3D reconstruction task, namely given a single \emph{RGB image as an input}, our network predicts its geometry as a \emph{tree of primitives as an output}. We compare our model with \cite{Paschalidou2019CVPR}. Both methods were trained for a maximum number of $32$ primitives until convergence. For our method, we set the sharpness value $s=10$.
\begin{figure}[h!]
	\centering
	\begin{subfigure}[b]{0.15\linewidth}
    	\centering
        \includegraphics[width=1.2\linewidth]{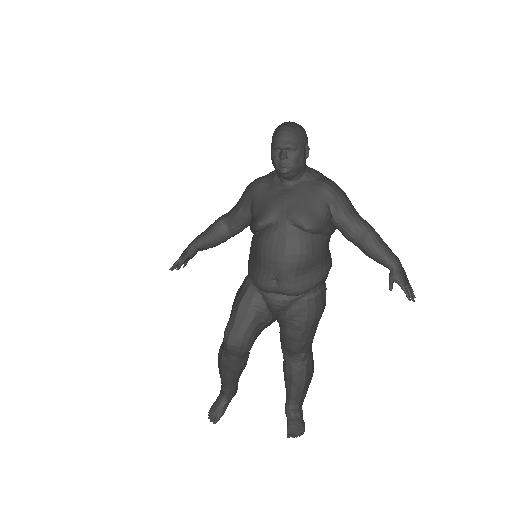}
    \end{subfigure}
    \hfill
    \begin{subfigure}[b]{0.15\linewidth}
    	\centering
        \includegraphics[width=1.2\linewidth]{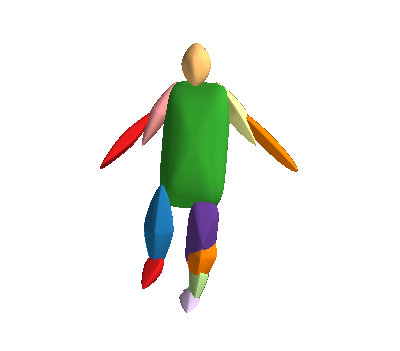}
    \end{subfigure}
    \hfill
    \begin{subfigure}[b]{0.15\linewidth}
    	\centering
        \includegraphics[width=\linewidth]{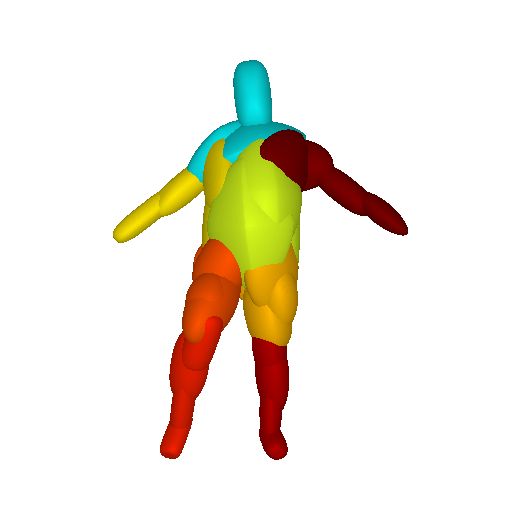}
    \end{subfigure}
    \hfill
    \begin{subfigure}[b]{0.15\linewidth}
    	\centering
        \includegraphics[width=1.2\linewidth]{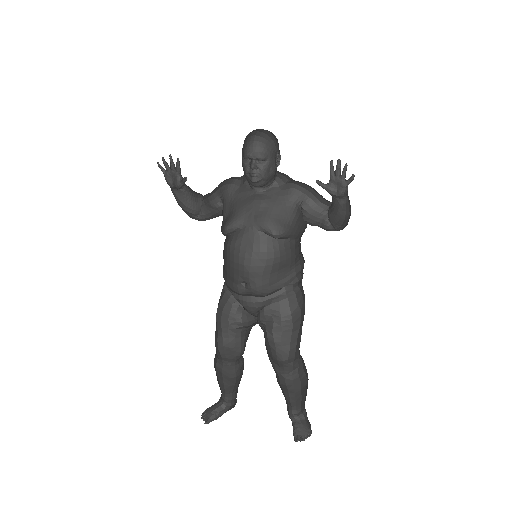}
    \end{subfigure}
    \hfill
    \begin{subfigure}[b]{0.15\linewidth}
    	\centering
        \includegraphics[width=1.2\linewidth]{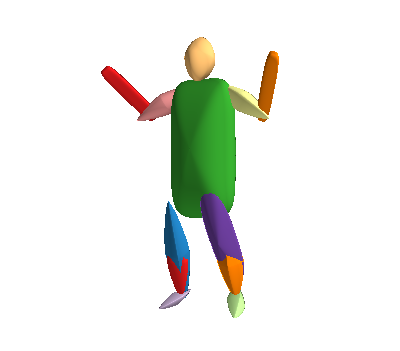}
    \end{subfigure}
    \hfill
    \begin{subfigure}[b]{0.15\linewidth}
    	\centering
        \includegraphics[width=\linewidth]{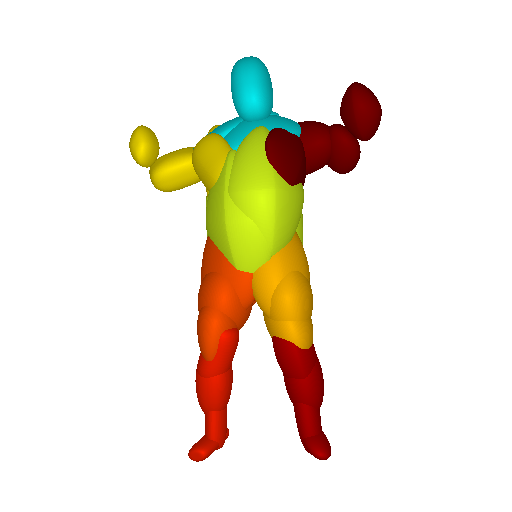}
    \end{subfigure}

    \vskip\baselineskip    \vspace{-1.2em}
    \begin{subfigure}[b]{0.15\linewidth}
		\centering
        \caption{\textbf{Input}}
    \end{subfigure}
    \hfill
    \begin{subfigure}[b]{0.15\linewidth}
		\centering
        \caption{\bf{SQs}\cite{Paschalidou2019CVPR}}
    \end{subfigure}
    \hfill
    \begin{subfigure}[b]{0.15\linewidth}
		\centering
        \caption{\textbf{Ours}}
    \end{subfigure}
    \hfill
    \begin{subfigure}[b]{0.15\linewidth}
		\centering
        \caption{\textbf{Input}}
    \end{subfigure}
    \hfill
    \begin{subfigure}[b]{0.15\linewidth}
		\centering
        \caption{\bf{SQs}\cite{Paschalidou2019CVPR}}
    \end{subfigure}
    \hfill
    \begin{subfigure}[b]{0.15\linewidth}
		\centering
        \caption{\textbf{Ours}}
    \end{subfigure}
    \vskip\baselineskip
    \begin{subfigure}[b]{0.49\linewidth}
    	\centering
        \includegraphics[width=\linewidth]{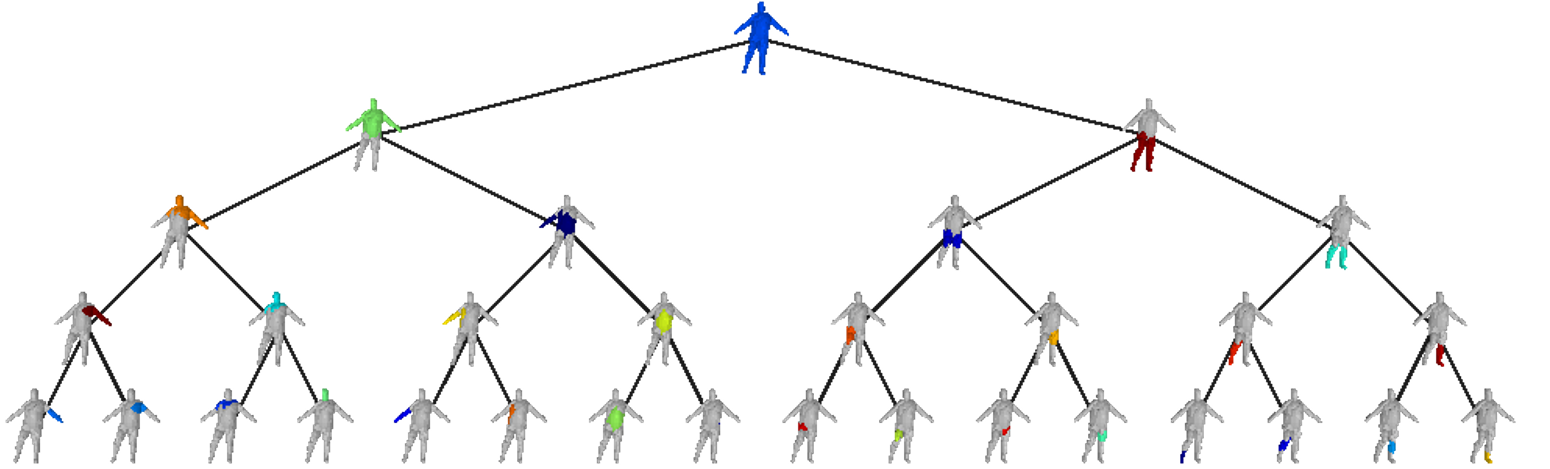}
    \end{subfigure}
    \hfill
    \begin{subfigure}[b]{0.49\linewidth}
    	\centering
        \includegraphics[width=\linewidth]{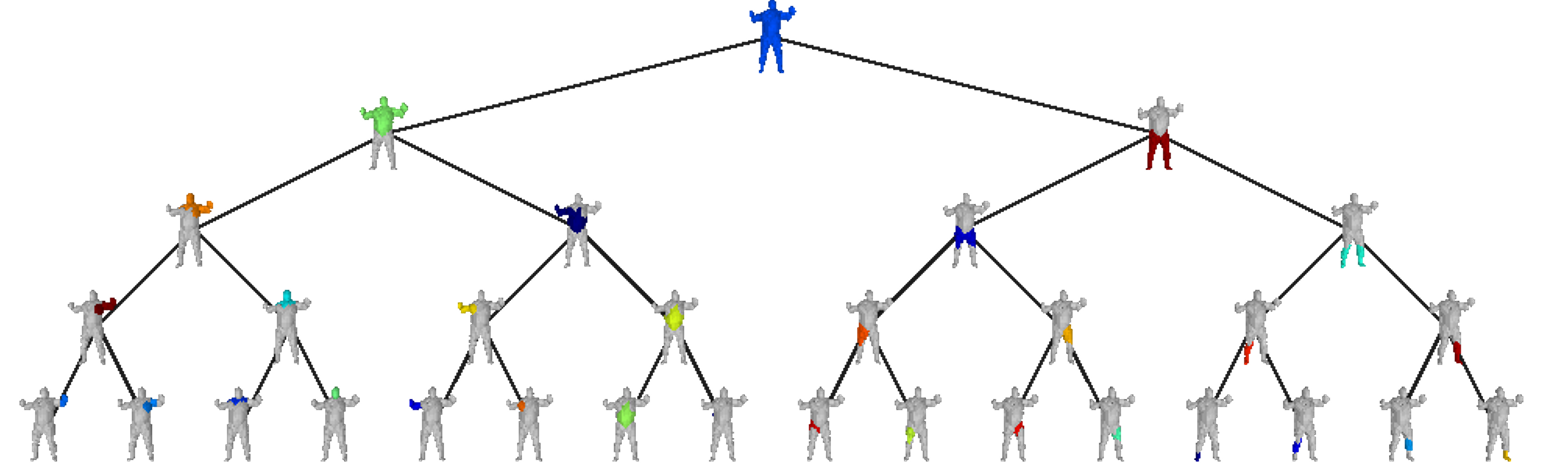}
    \end{subfigure}
    \vskip\baselineskip    \vspace{-1.2em}
    \begin{subfigure}[b]{0.49\linewidth}
		\centering
        \caption{\textbf{Predicted Hierarchy}}
    \end{subfigure}
    \hfill
    \begin{subfigure}[b]{0.49\linewidth}
		\centering
        \caption{\textbf{Predicted Hierarchy}}
    \end{subfigure}
    \vskip\baselineskip
	\begin{subfigure}[b]{0.15\linewidth}
    	\centering
        \includegraphics[width=1.2\linewidth]{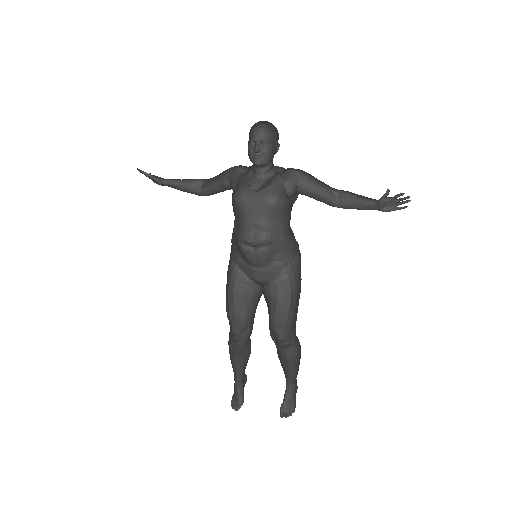}
    \end{subfigure}
    \hfill
    \begin{subfigure}[b]{0.15\linewidth}
    	\centering
        \includegraphics[width=1.2\linewidth]{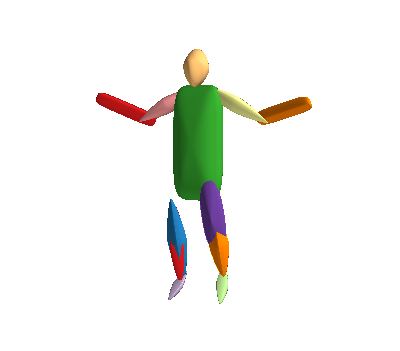}
    \end{subfigure}
    \hfill
    \begin{subfigure}[b]{0.15\linewidth}
    	\centering
        \includegraphics[width=1.0\linewidth]{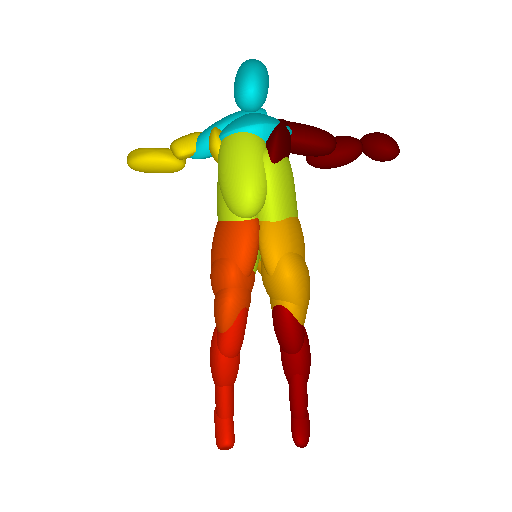}
    \end{subfigure}
    \hfill
    \begin{subfigure}[b]{0.15\linewidth}
    	\centering
        \includegraphics[width=1.4\linewidth]{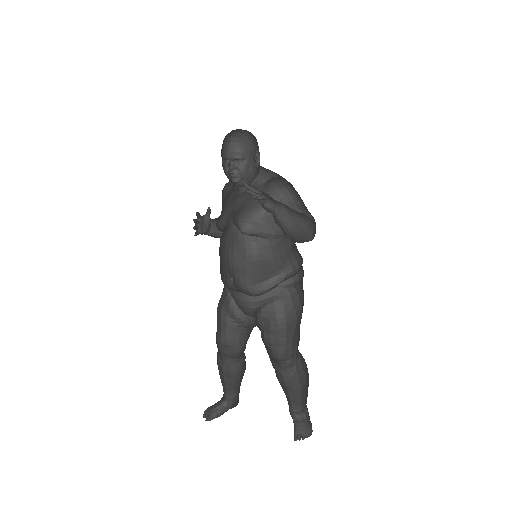}
    \end{subfigure}
    \hfill
    \begin{subfigure}[b]{0.15\linewidth}
    	\centering
        \includegraphics[width=1.2\linewidth]{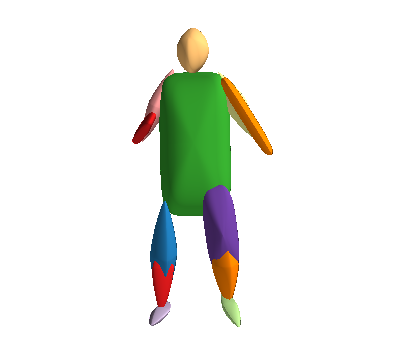}
    \end{subfigure}
    \hfill
    \begin{subfigure}[b]{0.15\linewidth}
    	\centering
        \includegraphics[width=1.2\linewidth]{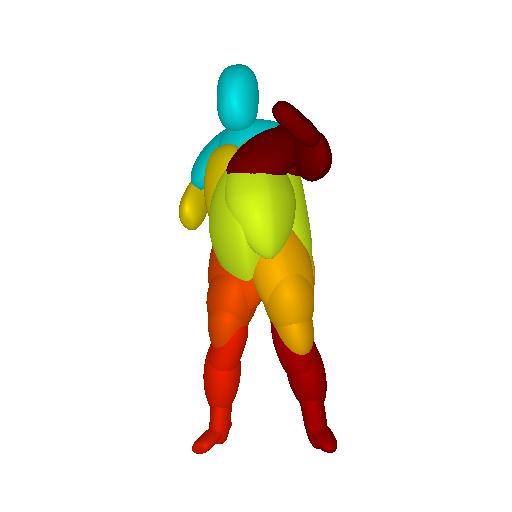}
    \end{subfigure}

    \vskip\baselineskip    \vspace{-1.2em}
    \begin{subfigure}[b]{0.15\linewidth}
		\centering
        \caption{\textbf{Input}}
    \end{subfigure}
    \hfill
    \begin{subfigure}[b]{0.15\linewidth}
		\centering
        \caption{\bf{SQs}\cite{Paschalidou2019CVPR}}
    \end{subfigure}
    \hfill
    \begin{subfigure}[b]{0.15\linewidth}
		\centering
        \caption{\textbf{Ours}}
    \end{subfigure}
    \hfill
    \begin{subfigure}[b]{0.15\linewidth}
		\centering
        \caption{\textbf{Input}}
    \end{subfigure}
    \hfill
    \begin{subfigure}[b]{0.15\linewidth}
		\centering
        \caption{\bf{SQs}\cite{Paschalidou2019CVPR}}
    \end{subfigure}
    \hfill
    \begin{subfigure}[b]{0.15\linewidth}
		\centering
        \caption{\textbf{Ours}}
    \end{subfigure}
    \vskip\baselineskip
    \begin{subfigure}[b]{0.49\linewidth}
    	\centering
        \includegraphics[width=1.0\linewidth]{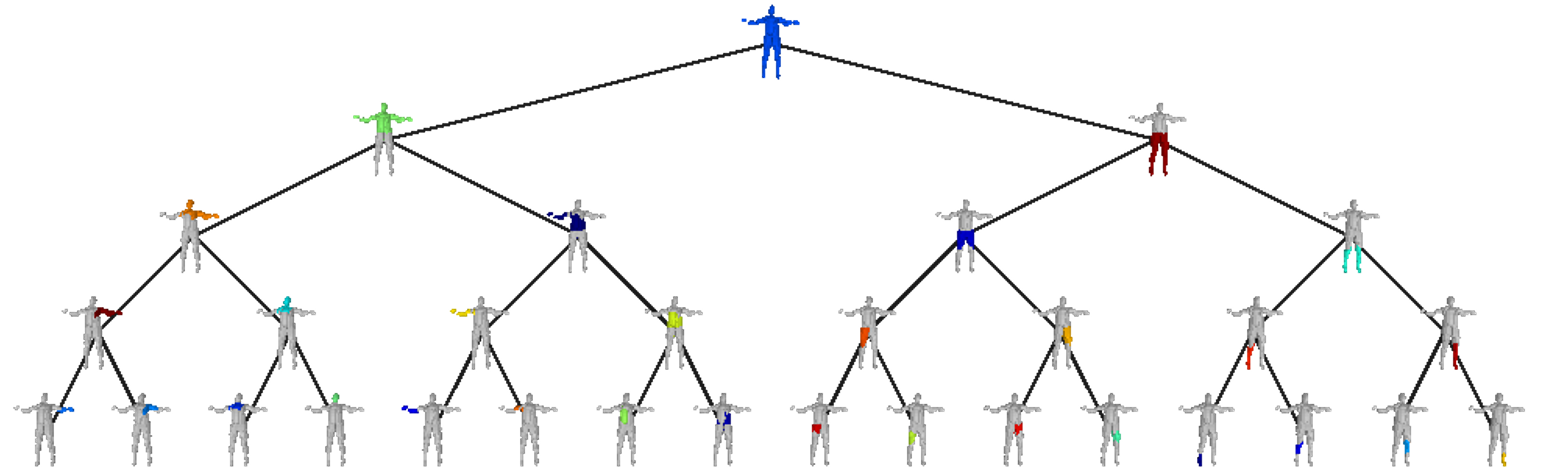}
    \end{subfigure}
    \hfill
    \begin{subfigure}[b]{0.49\linewidth}
    	\centering
        \includegraphics[width=\linewidth]{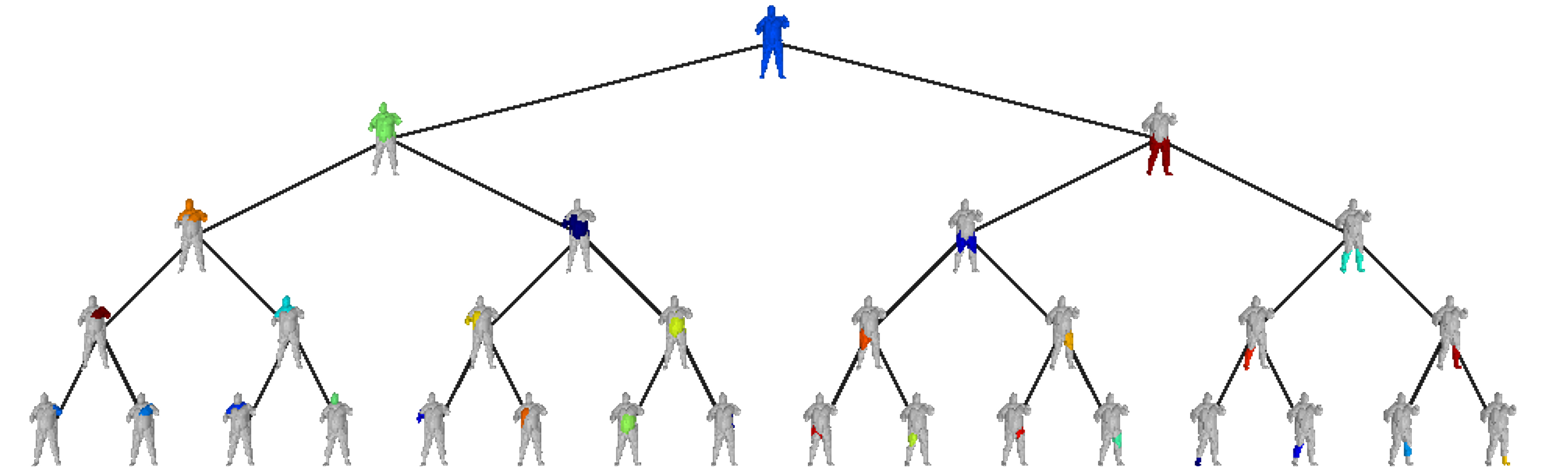}
    \end{subfigure}
    \vskip\baselineskip    \vspace{-1.2em}
    \begin{subfigure}[b]{0.49\linewidth}
		\centering
        \caption{\textbf{Predicted Hierarchy}}
    \end{subfigure}
    \hfill
    \begin{subfigure}[b]{0.49\linewidth}
		\centering
        \caption{\textbf{Predicted Hierarchy}}
    \end{subfigure}
    \caption{\textbf{Qualitative Results on D-FAUST.} Our network learns semantic mappings of body parts across different body shapes and articulations while being geometrical more accurate compared to \cite{Paschalidou2019CVPR}.}
    \label{fig:dfaust_supp}
\end{figure}

\begin{figure}
	\centering
    \begin{subfigure}[b]{\linewidth}
	\begin{subfigure}[b]{0.19\linewidth}
    	\centering
        \includegraphics[width=1.2\linewidth]{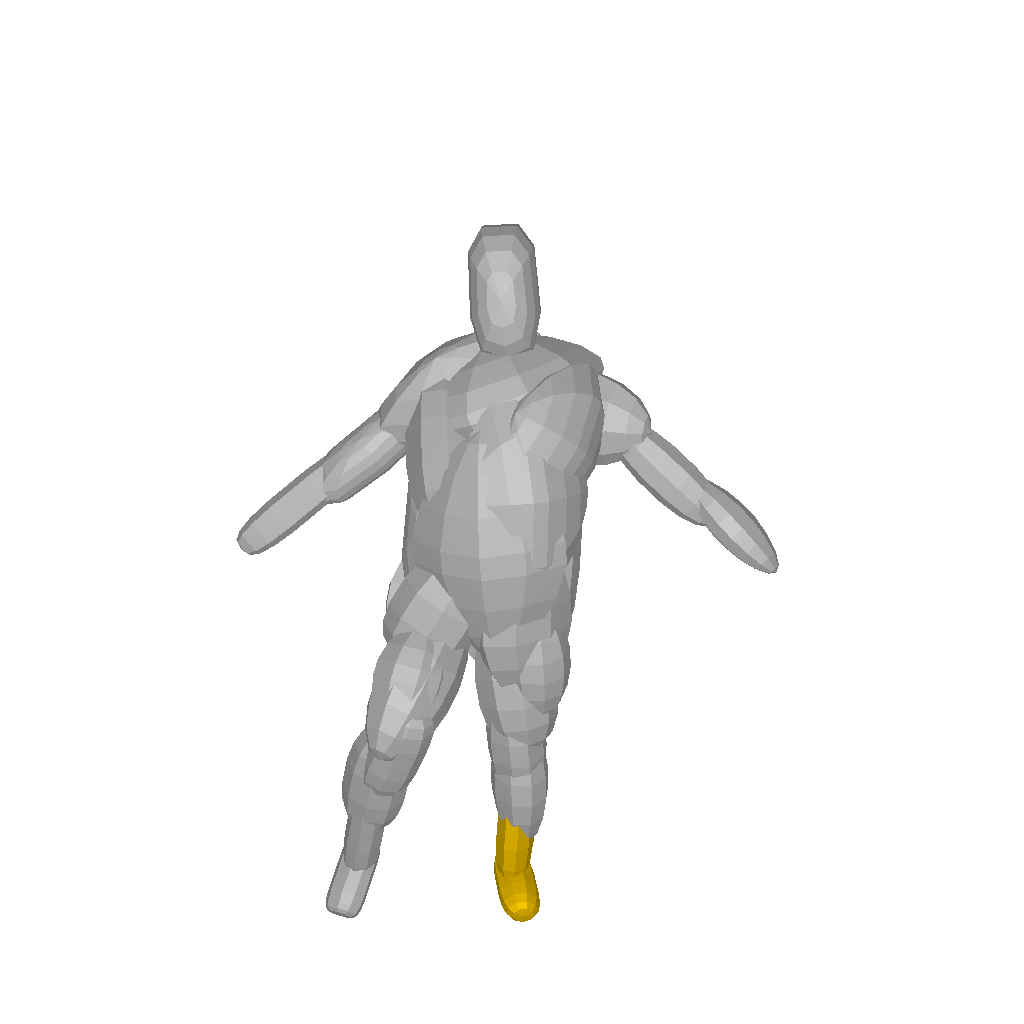}
    \end{subfigure}
    \hfill
    \begin{subfigure}[b]{0.19\linewidth}
    	\centering
        \includegraphics[width=1.2\linewidth]{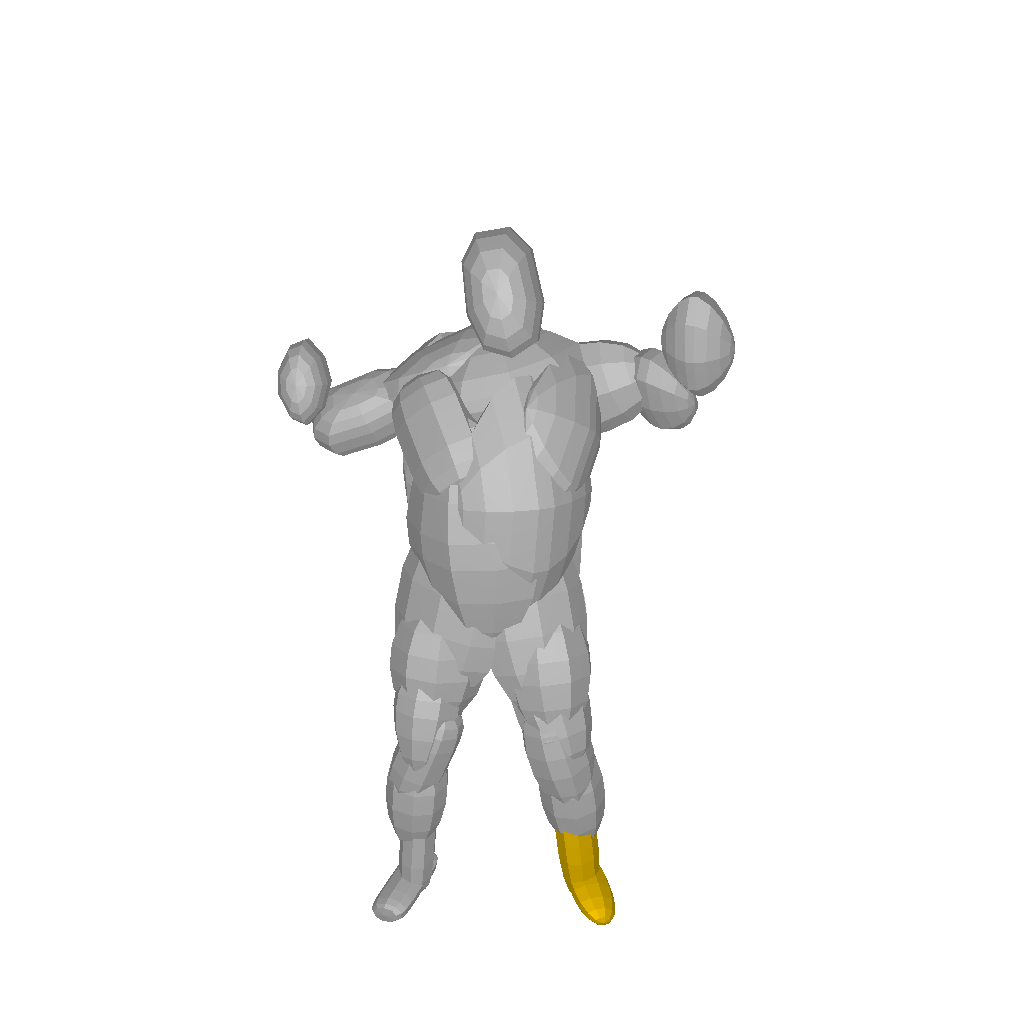}
    \end{subfigure}
    \hfill
    \begin{subfigure}[b]{0.19\linewidth}
    	\centering
        \includegraphics[width=1.2\linewidth]{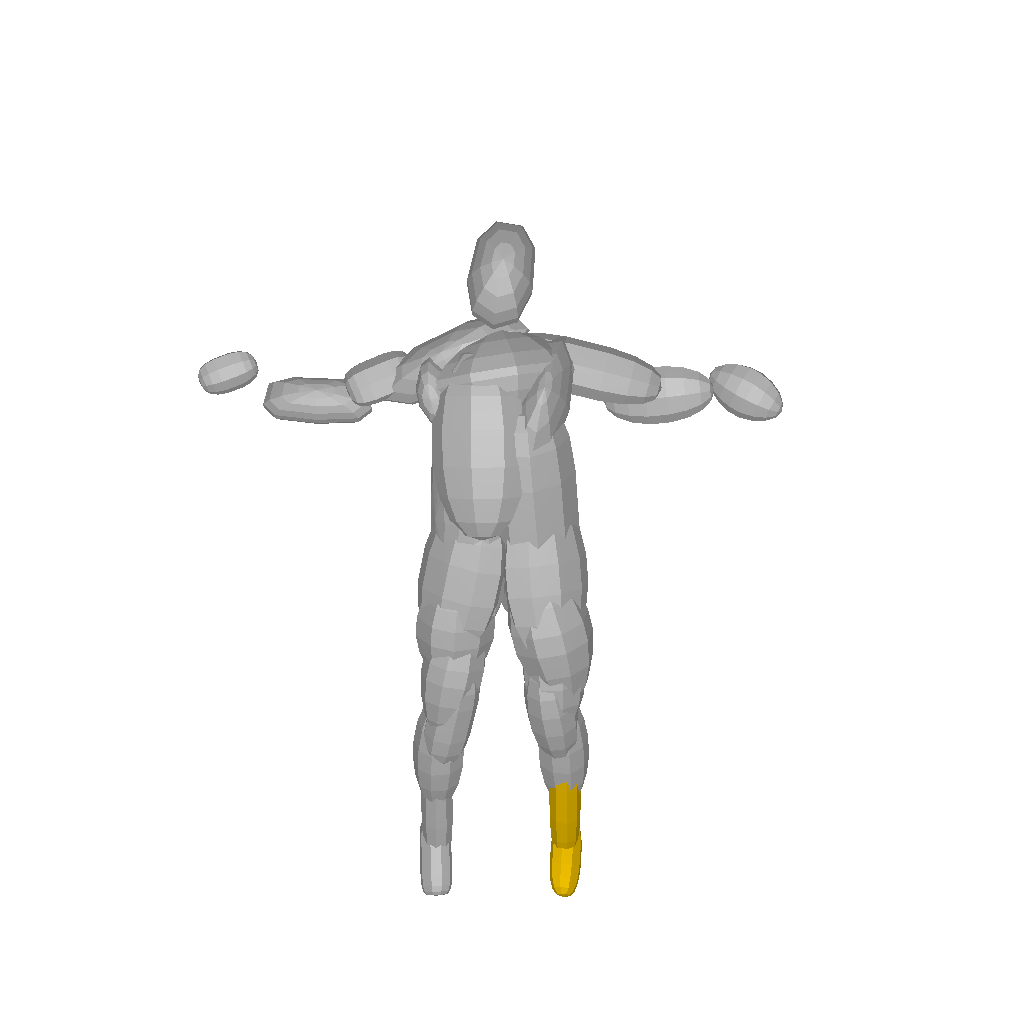}
    \end{subfigure}
    \hfill
    \begin{subfigure}[b]{0.19\linewidth}
    	\centering
        \includegraphics[width=1.2\linewidth]{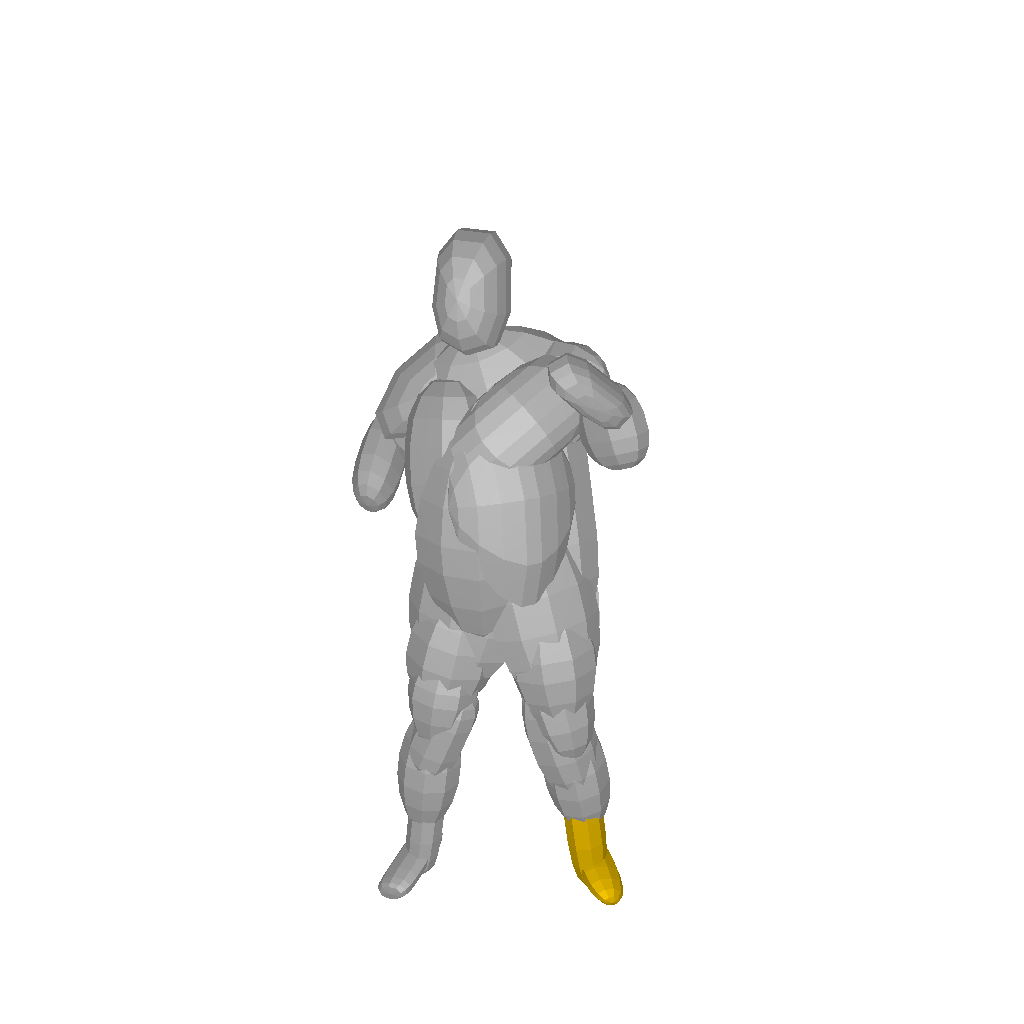}
    \end{subfigure}
    \hfill
    \begin{subfigure}[b]{0.19\linewidth}
    	\centering
        \includegraphics[width=1.2\linewidth]{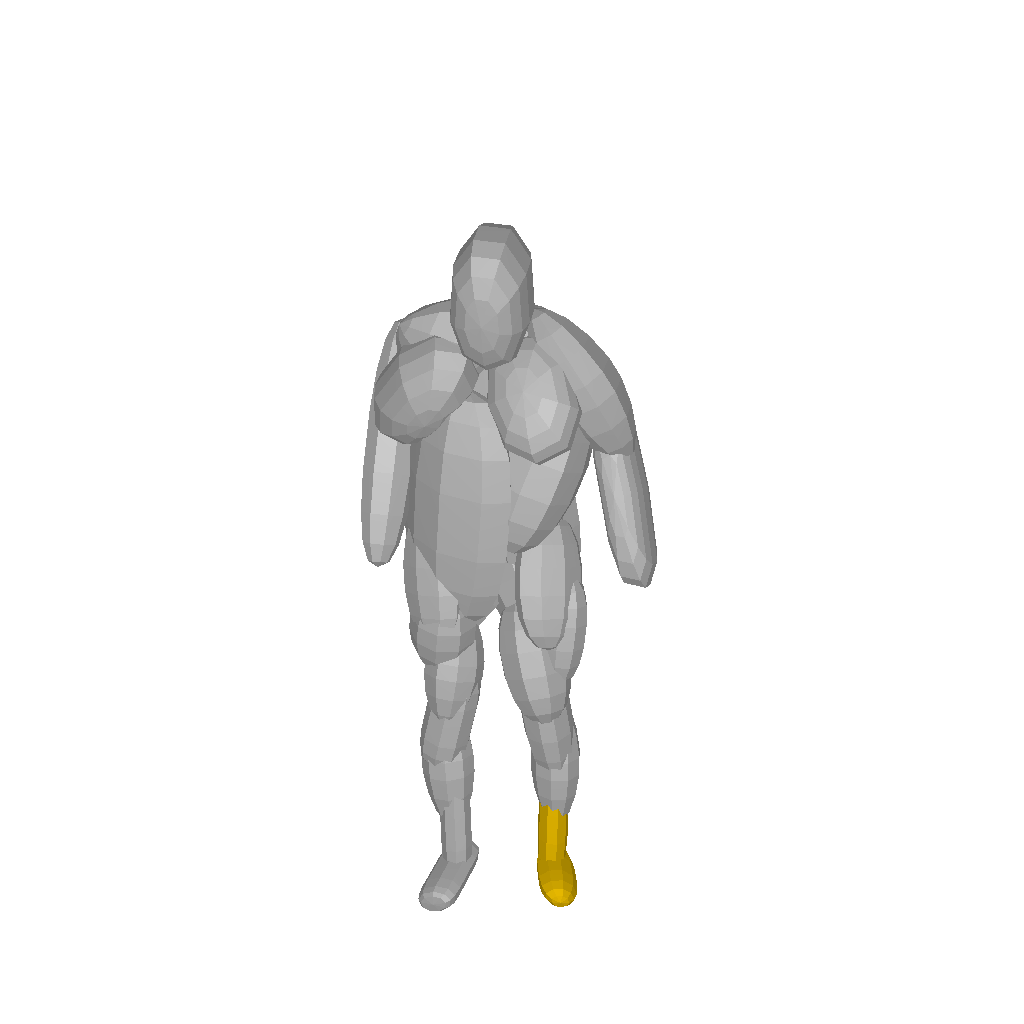}
    \end{subfigure}
    \vspace{-1.3em}
    \caption{Node $(4,0)$}
    \end{subfigure}

    \begin{subfigure}[b]{\linewidth}
	\begin{subfigure}[b]{0.19\linewidth}
    	\centering
        \includegraphics[width=1.2\linewidth]{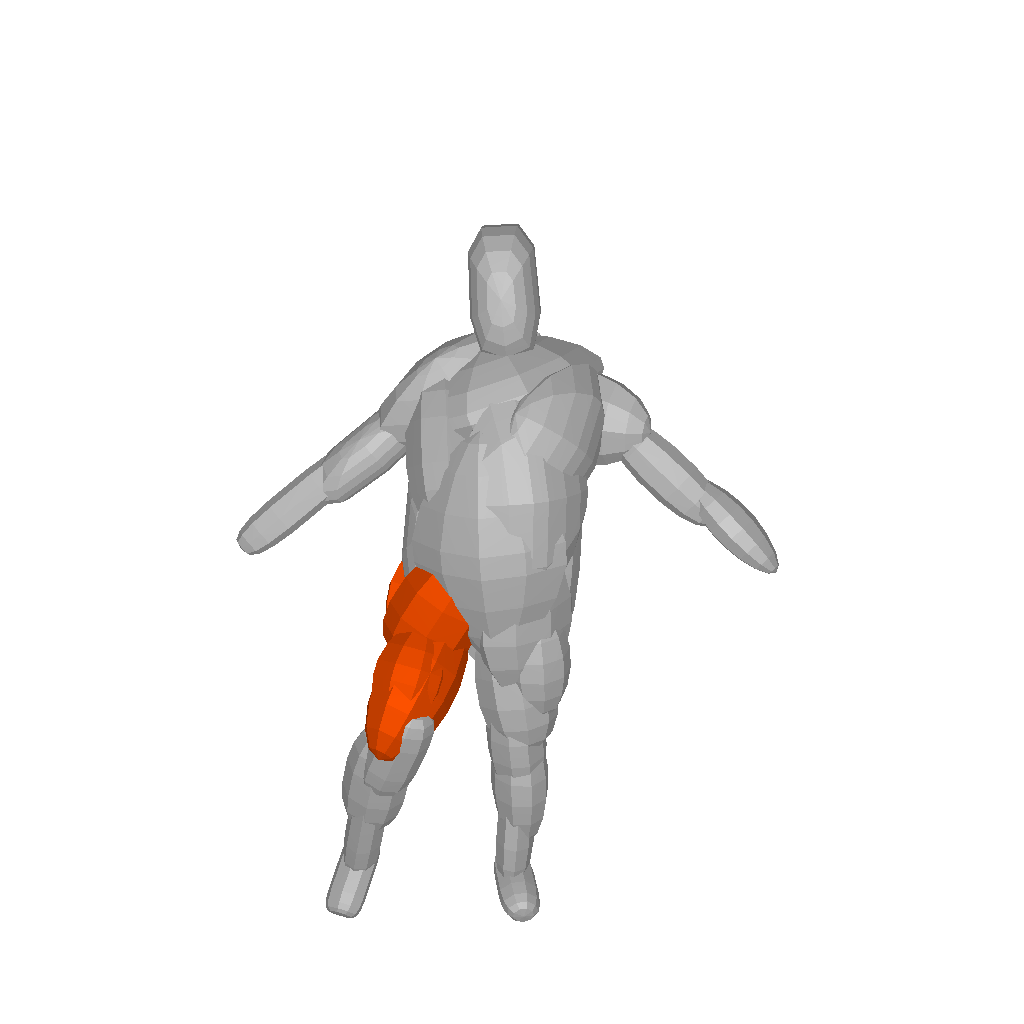}
    \end{subfigure}
    \hfill
    \begin{subfigure}[b]{0.19\linewidth}
    	\centering
        \includegraphics[width=1.2\linewidth]{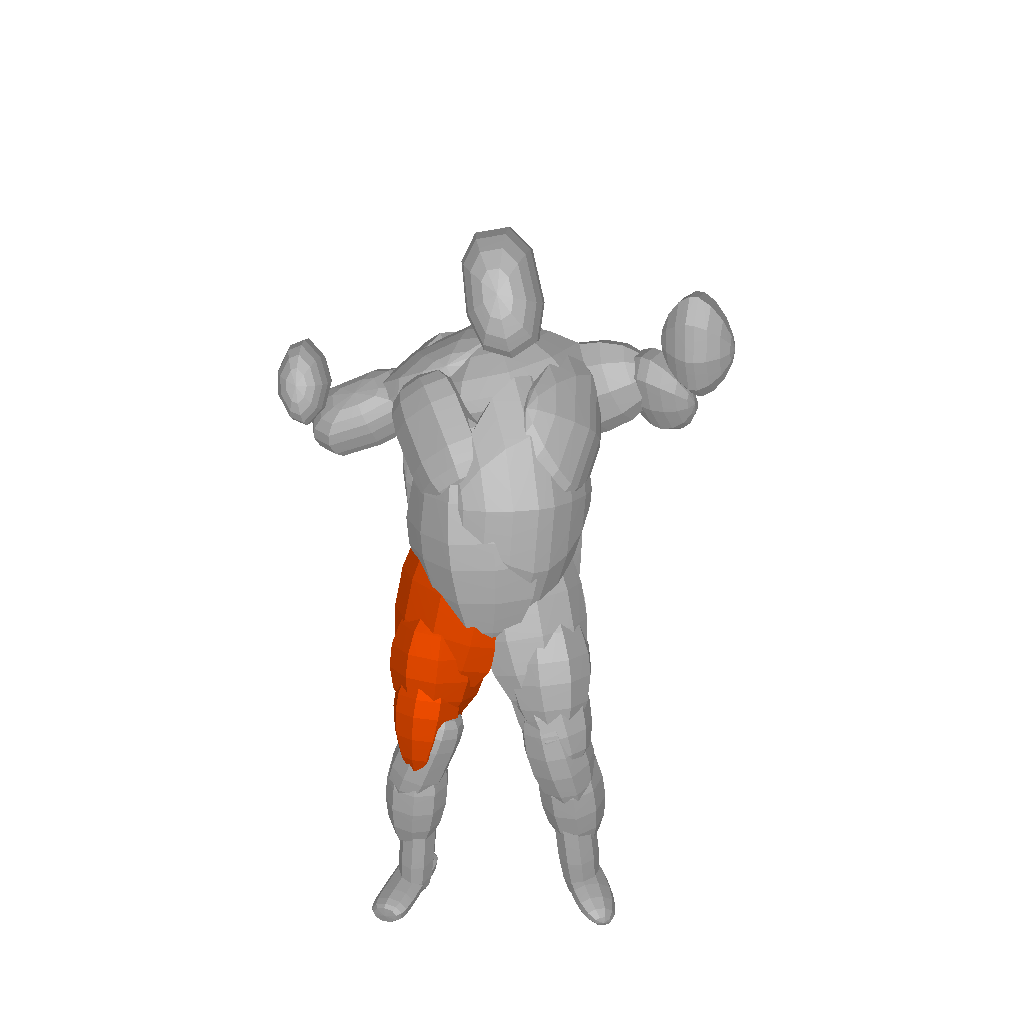}
    \end{subfigure}
    \hfill
    \begin{subfigure}[b]{0.19\linewidth}
    	\centering
        \includegraphics[width=1.2\linewidth]{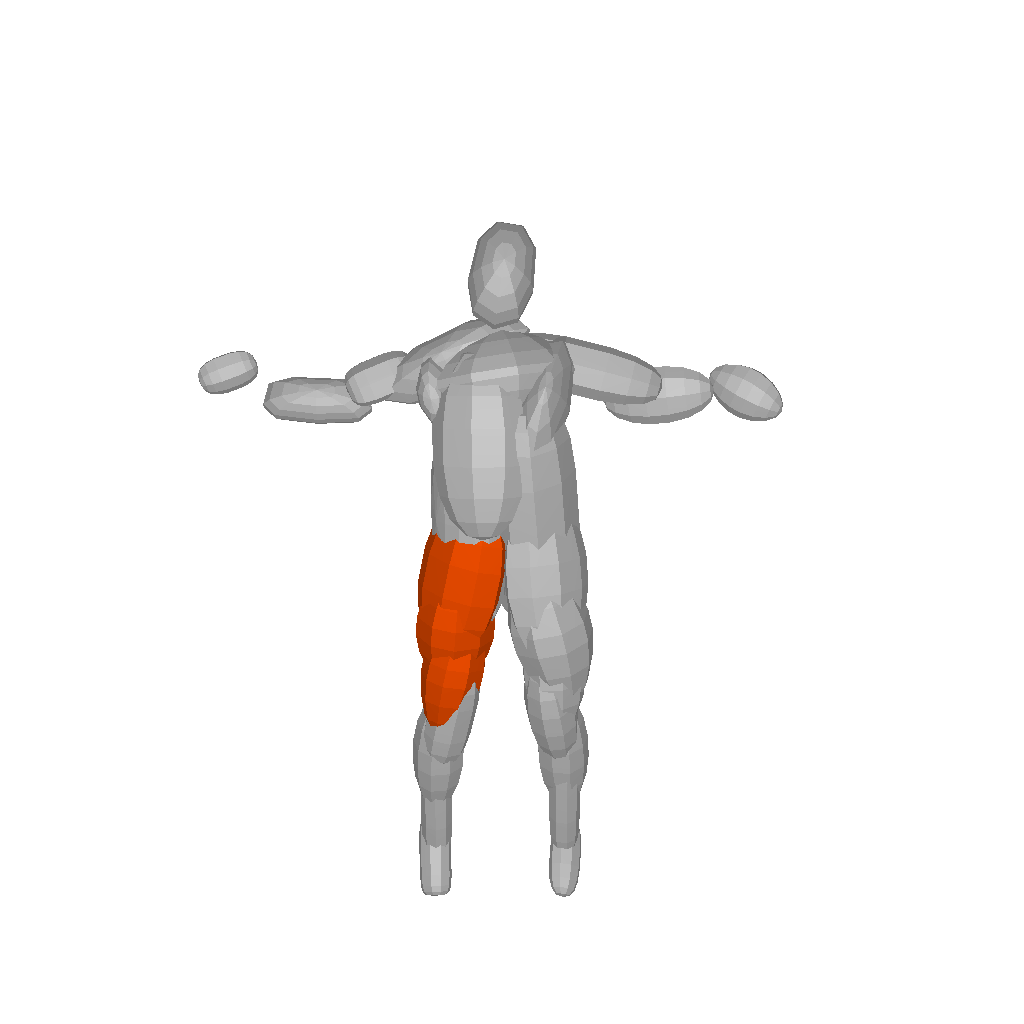}
    \end{subfigure}
    \hfill
    \begin{subfigure}[b]{0.19\linewidth}
    	\centering
        \includegraphics[width=1.2\linewidth]{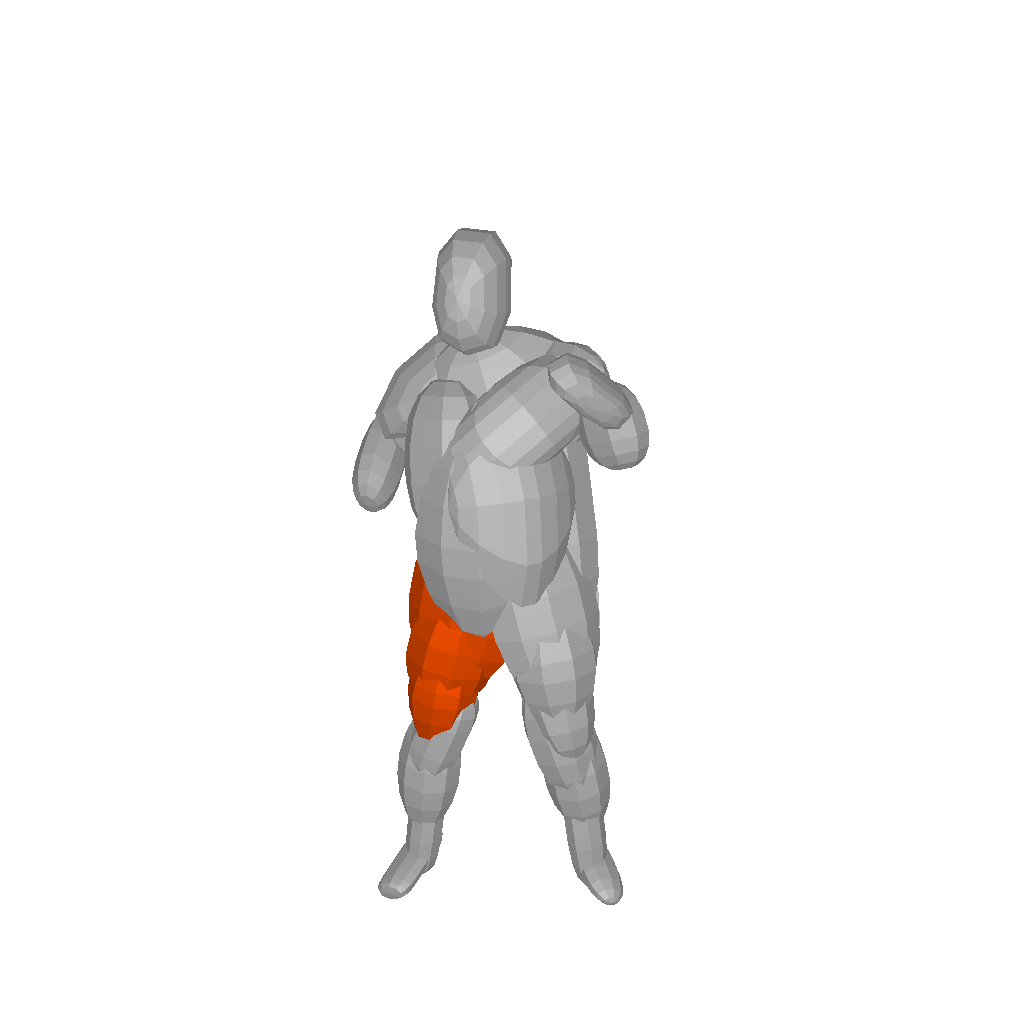}
    \end{subfigure}
    \hfill
    \begin{subfigure}[b]{0.19\linewidth}
    	\centering
        \includegraphics[width=1.2\linewidth]{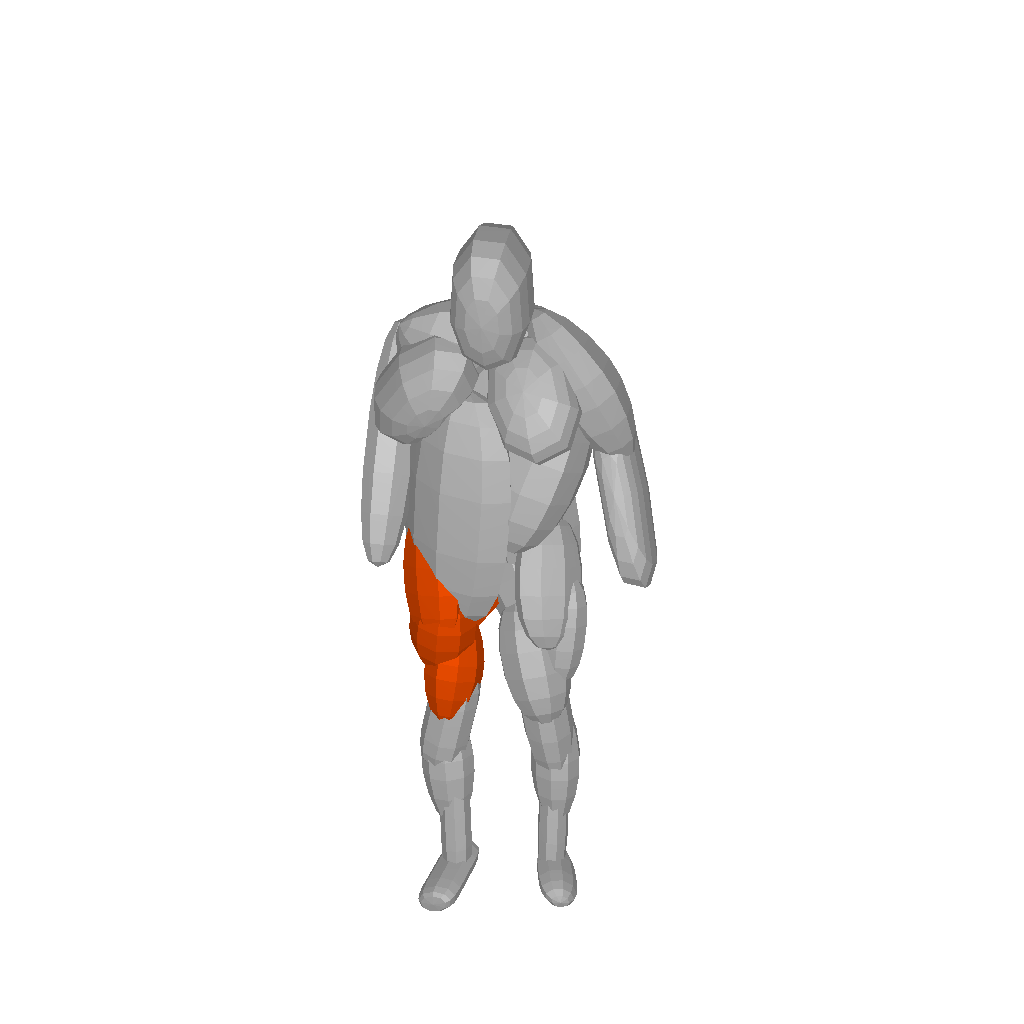}
    \end{subfigure}
    \vspace{-1.3em}
    \caption{Node $(3,3)$}
    \end{subfigure}

    \begin{subfigure}[b]{\linewidth}
	\begin{subfigure}[b]{0.19\linewidth}
    	\centering
        \includegraphics[width=1.2\linewidth]{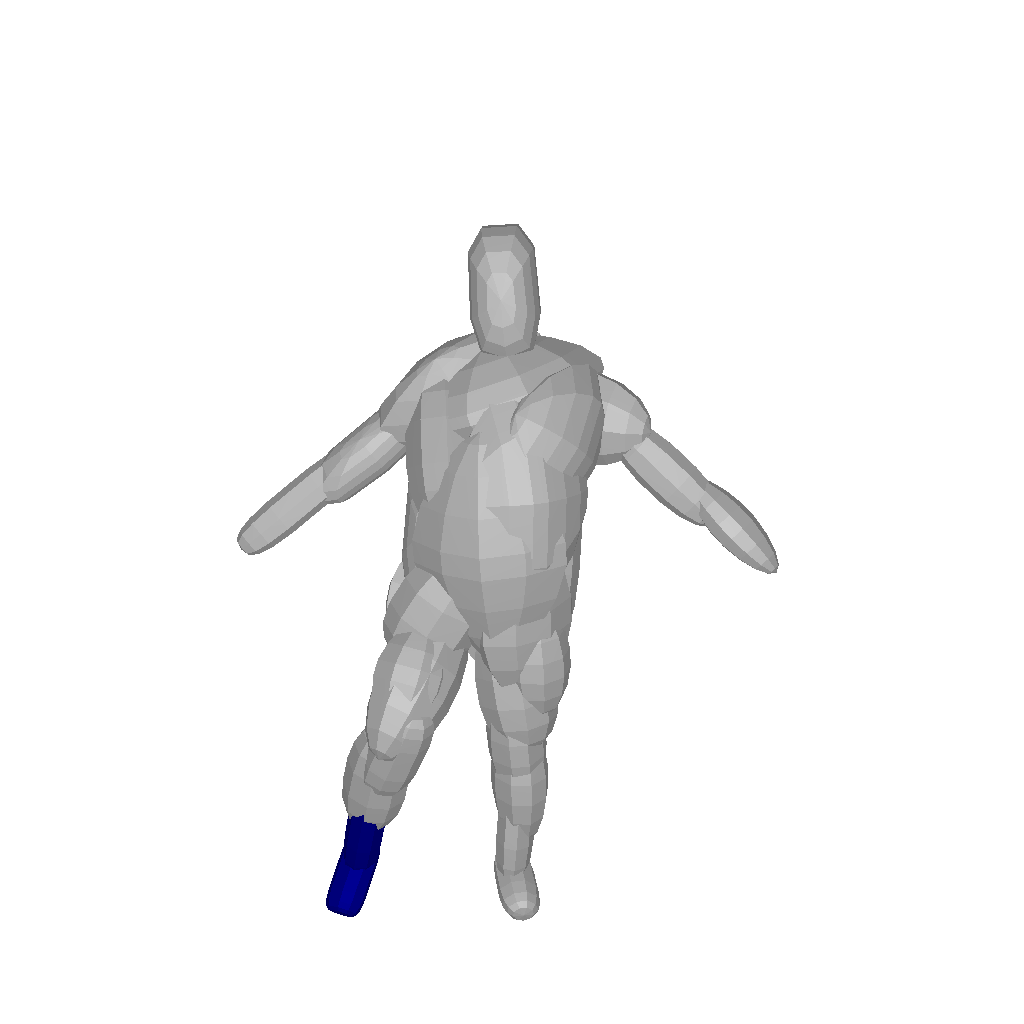}
    \end{subfigure}
    \hfill
    \begin{subfigure}[b]{0.19\linewidth}
    	\centering
        \includegraphics[width=1.2\linewidth]{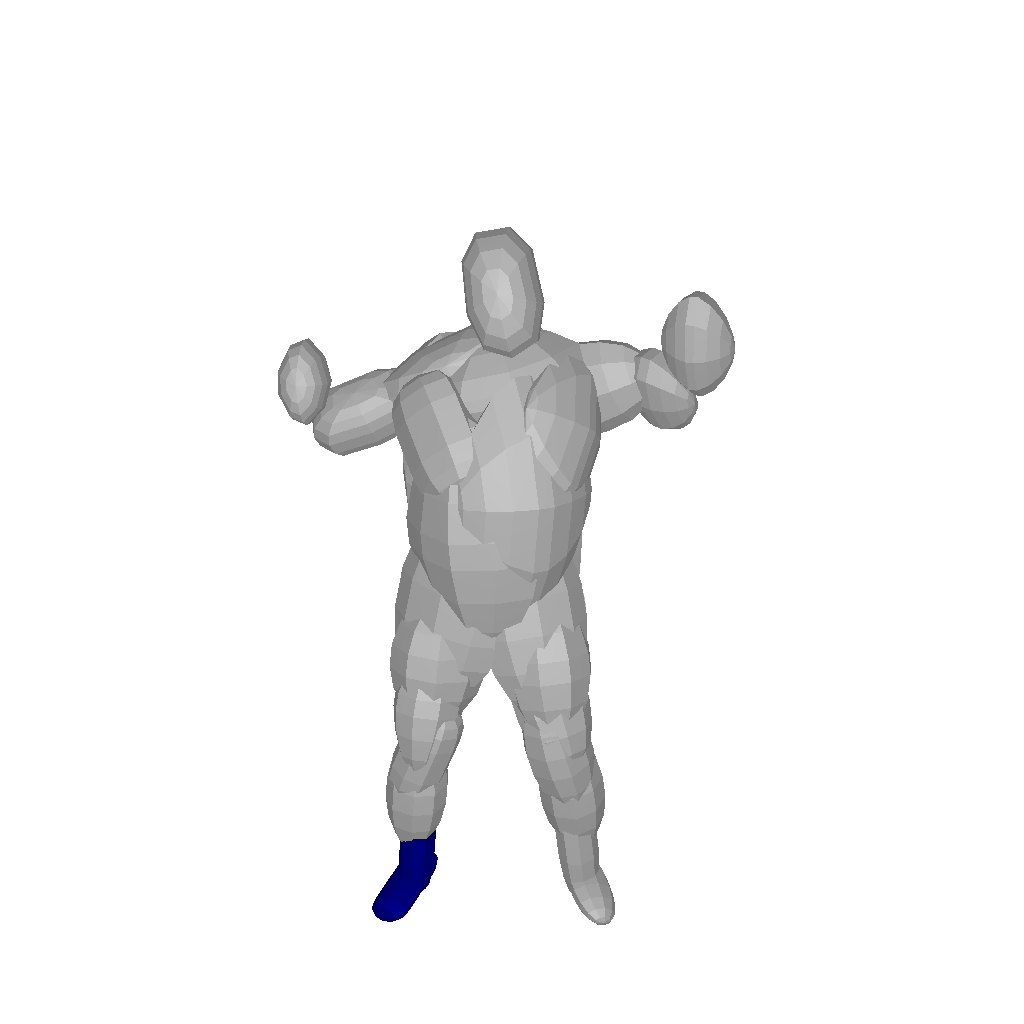}
    \end{subfigure}
    \hfill
    \begin{subfigure}[b]{0.19\linewidth}
    	\centering
        \includegraphics[width=1.2\linewidth]{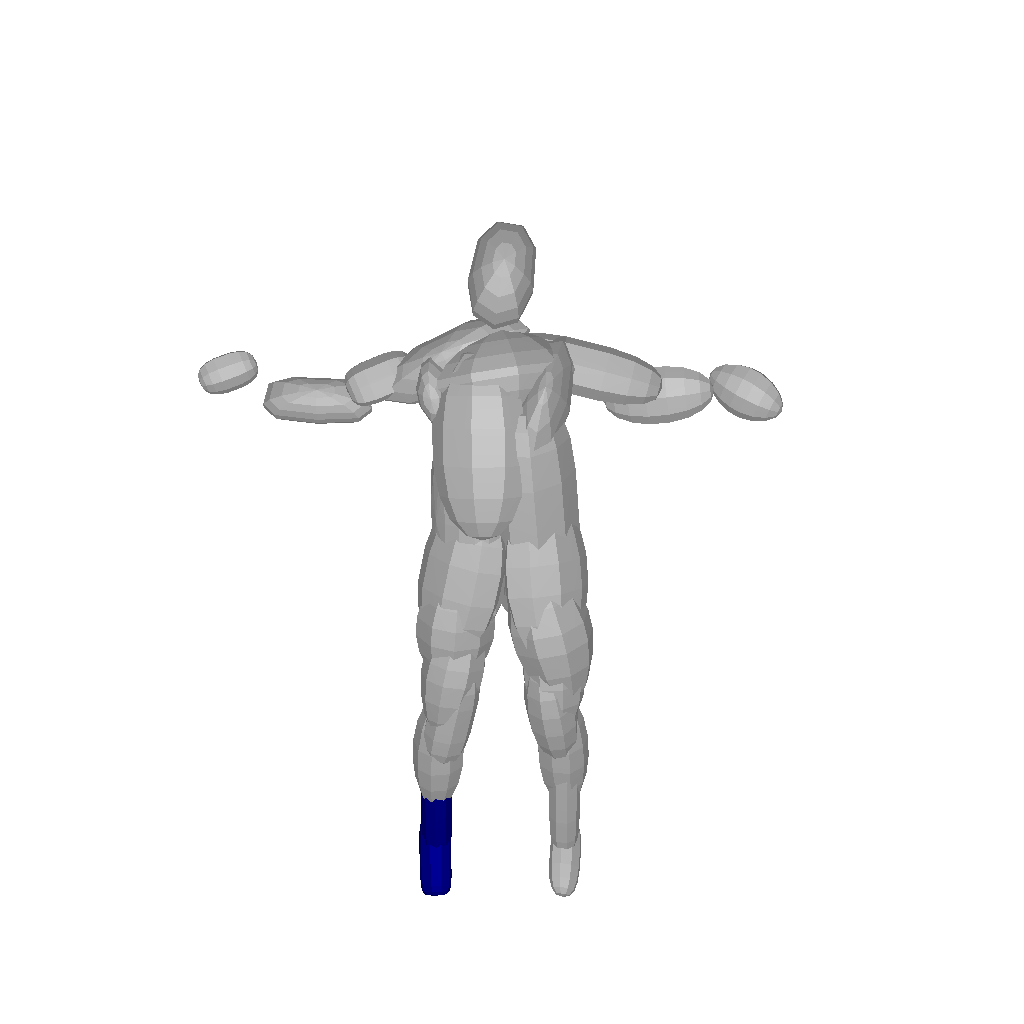}
    \end{subfigure}
    \hfill
    \begin{subfigure}[b]{0.19\linewidth}
    	\centering
        \includegraphics[width=1.2\linewidth]{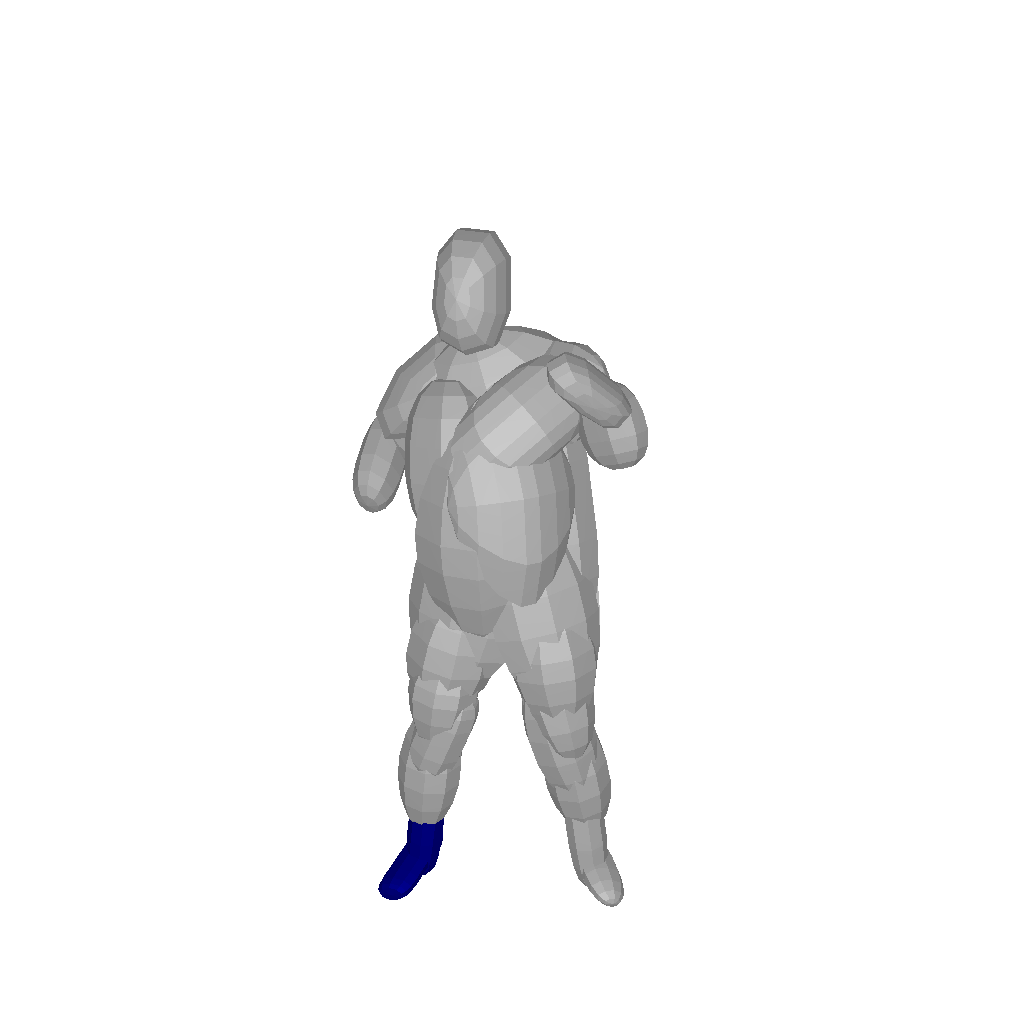}
    \end{subfigure}
    \hfill
    \begin{subfigure}[b]{0.19\linewidth}
    	\centering
        \includegraphics[width=1.2\linewidth]{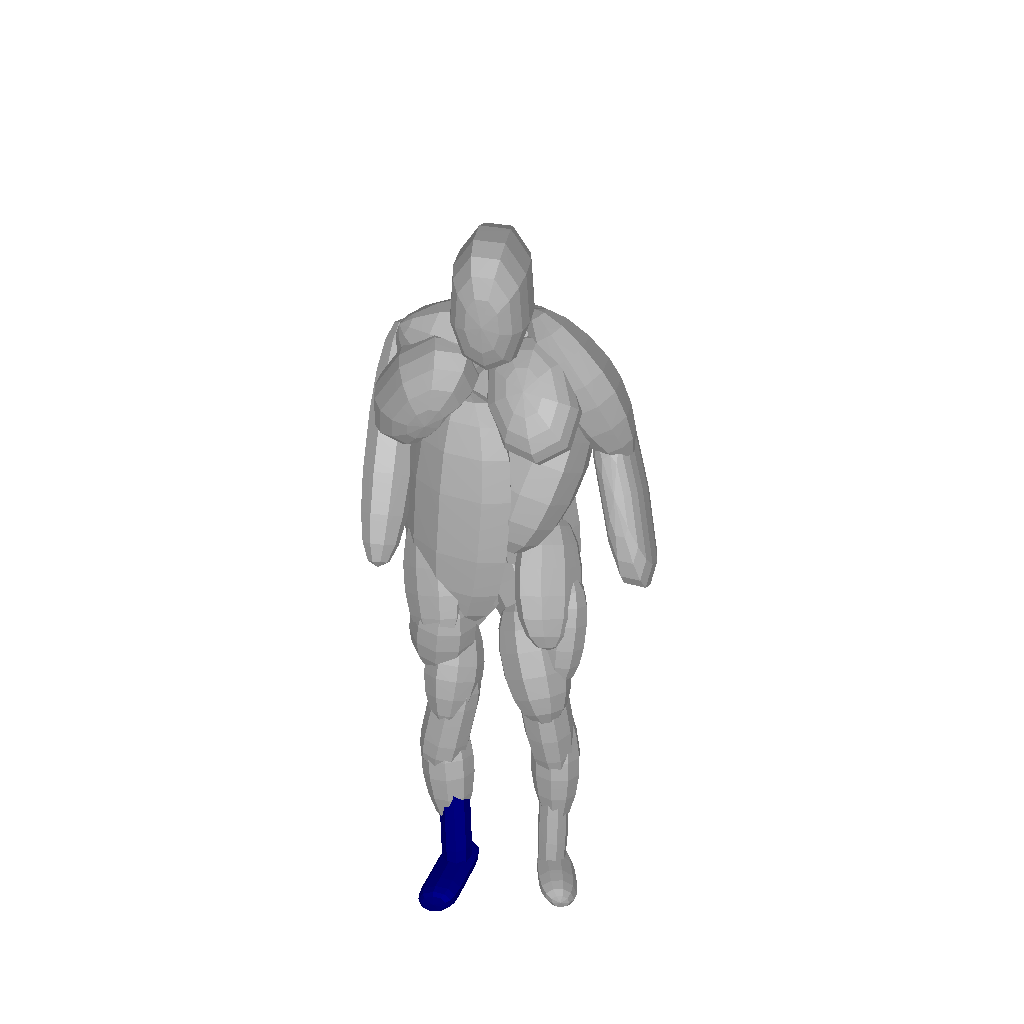}
    \end{subfigure}
    \caption{Node $(4,3)$}
    \end{subfigure}

    \begin{subfigure}[b]{\linewidth}
	\begin{subfigure}[b]{0.19\linewidth}
    	\centering
        \includegraphics[width=1.2\linewidth]{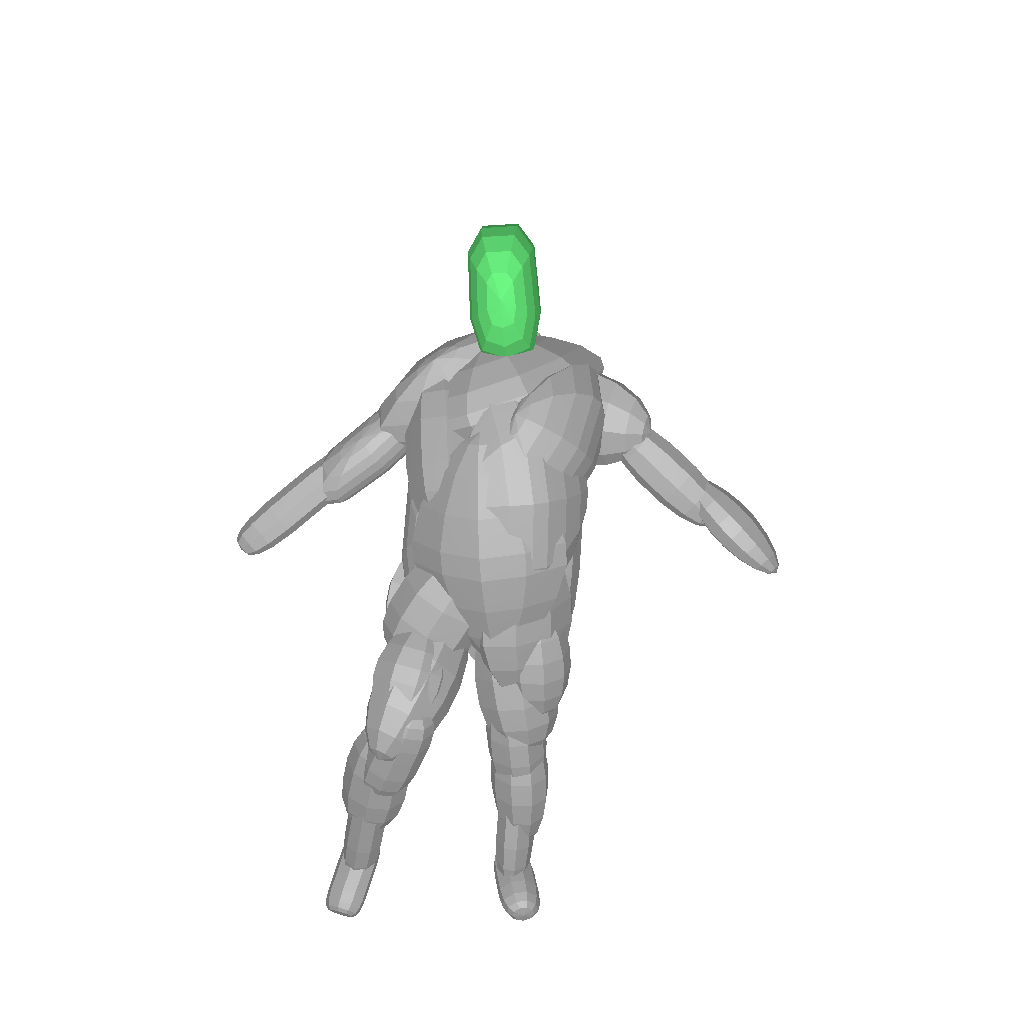}
    \end{subfigure}
    \hfill
    \begin{subfigure}[b]{0.19\linewidth}
    	\centering
        \includegraphics[width=1.2\linewidth]{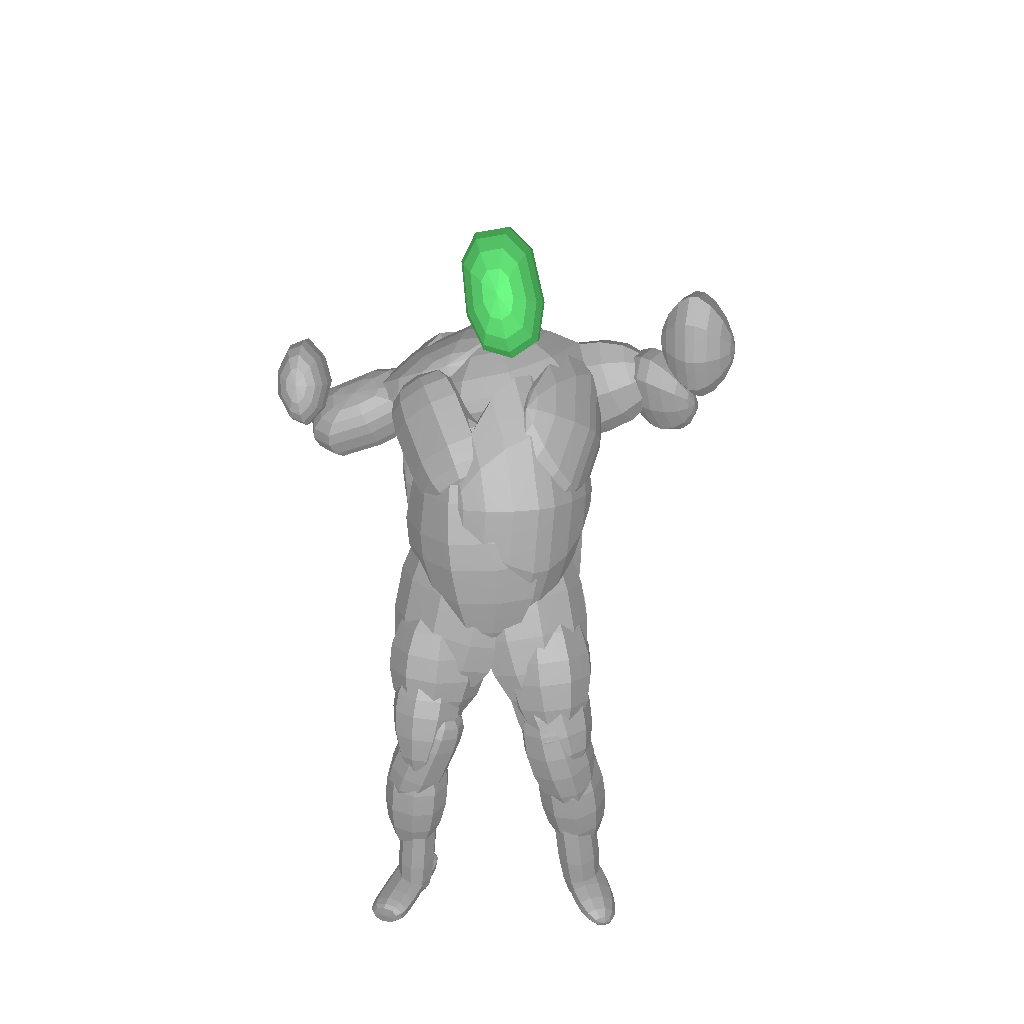}
    \end{subfigure}
    \hfill
    \begin{subfigure}[b]{0.19\linewidth}
    	\centering
        \includegraphics[width=1.2\linewidth]{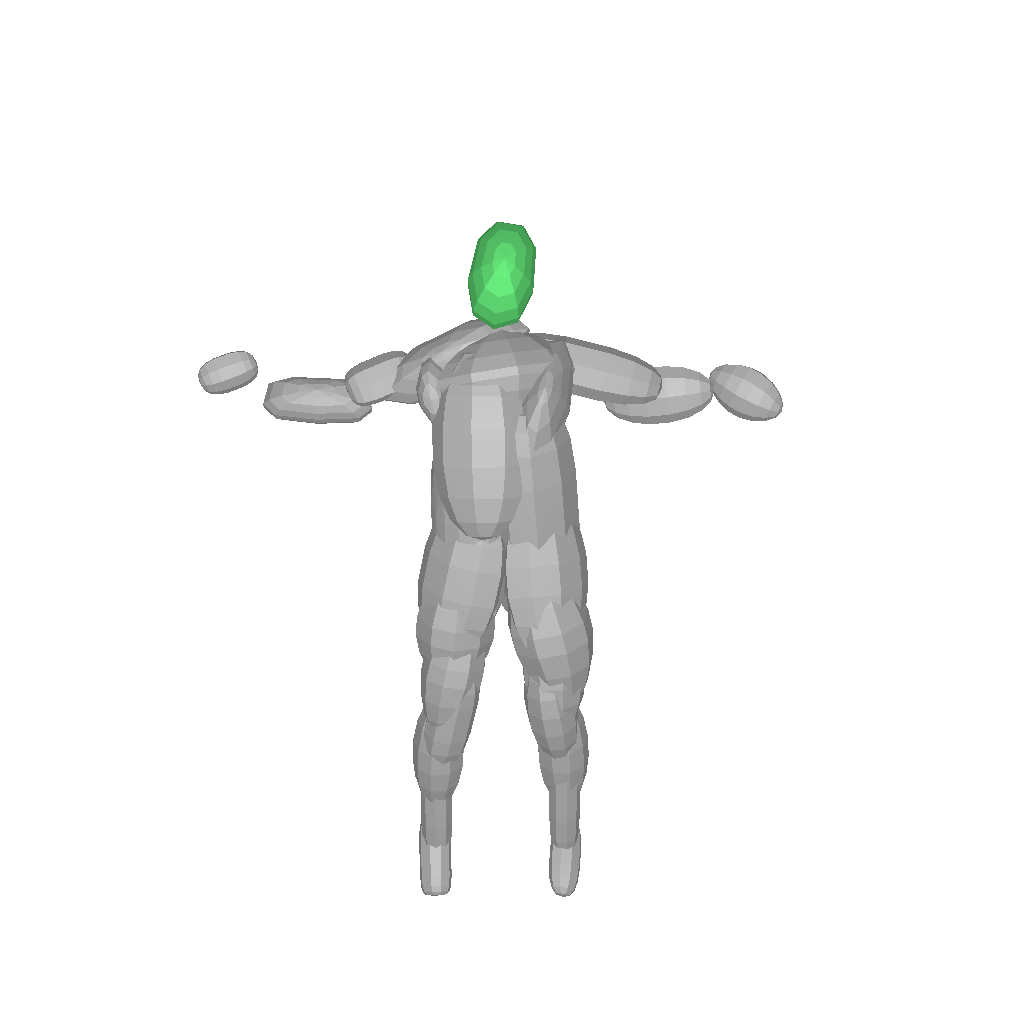}
    \end{subfigure}
    \hfill
    \begin{subfigure}[b]{0.19\linewidth}
    	\centering
        \includegraphics[width=1.2\linewidth]{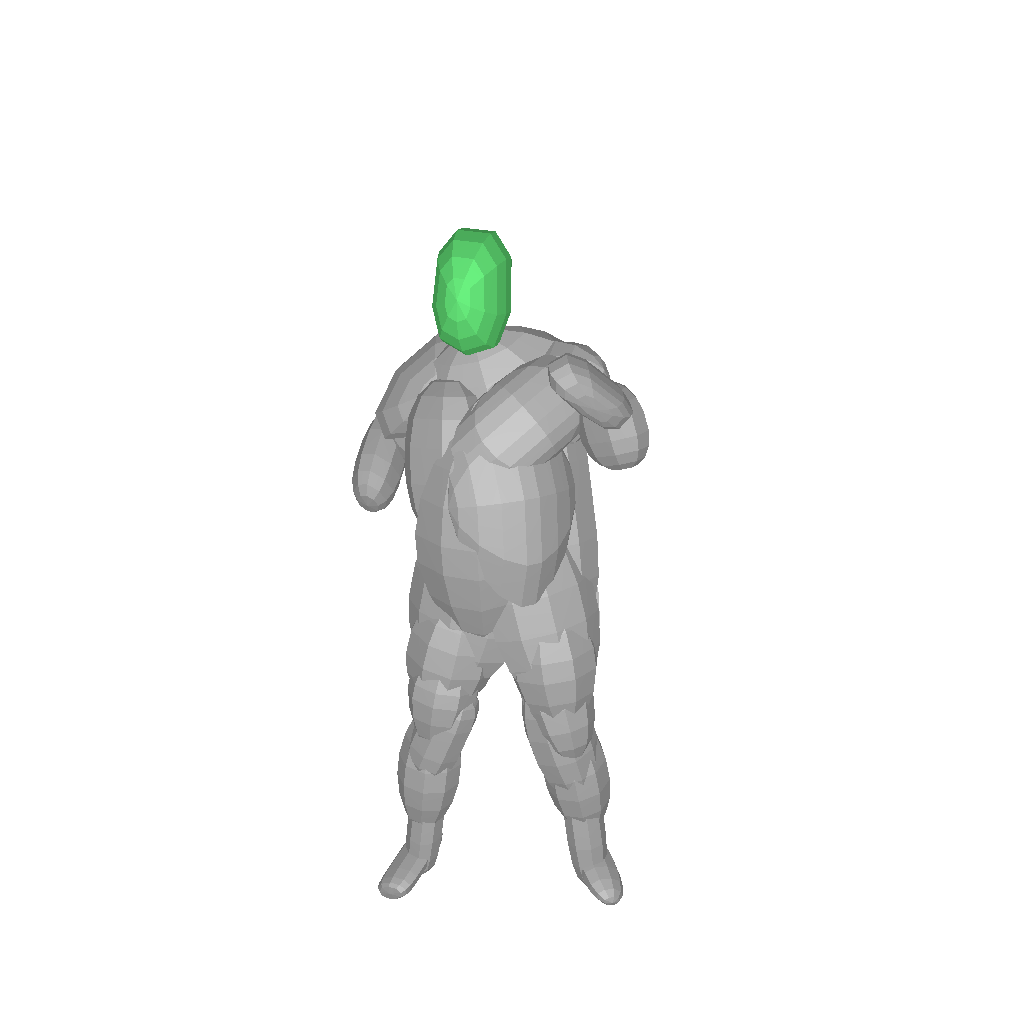}
    \end{subfigure}
    \hfill
    \begin{subfigure}[b]{0.19\linewidth}
    	\centering
        \includegraphics[width=1.2\linewidth]{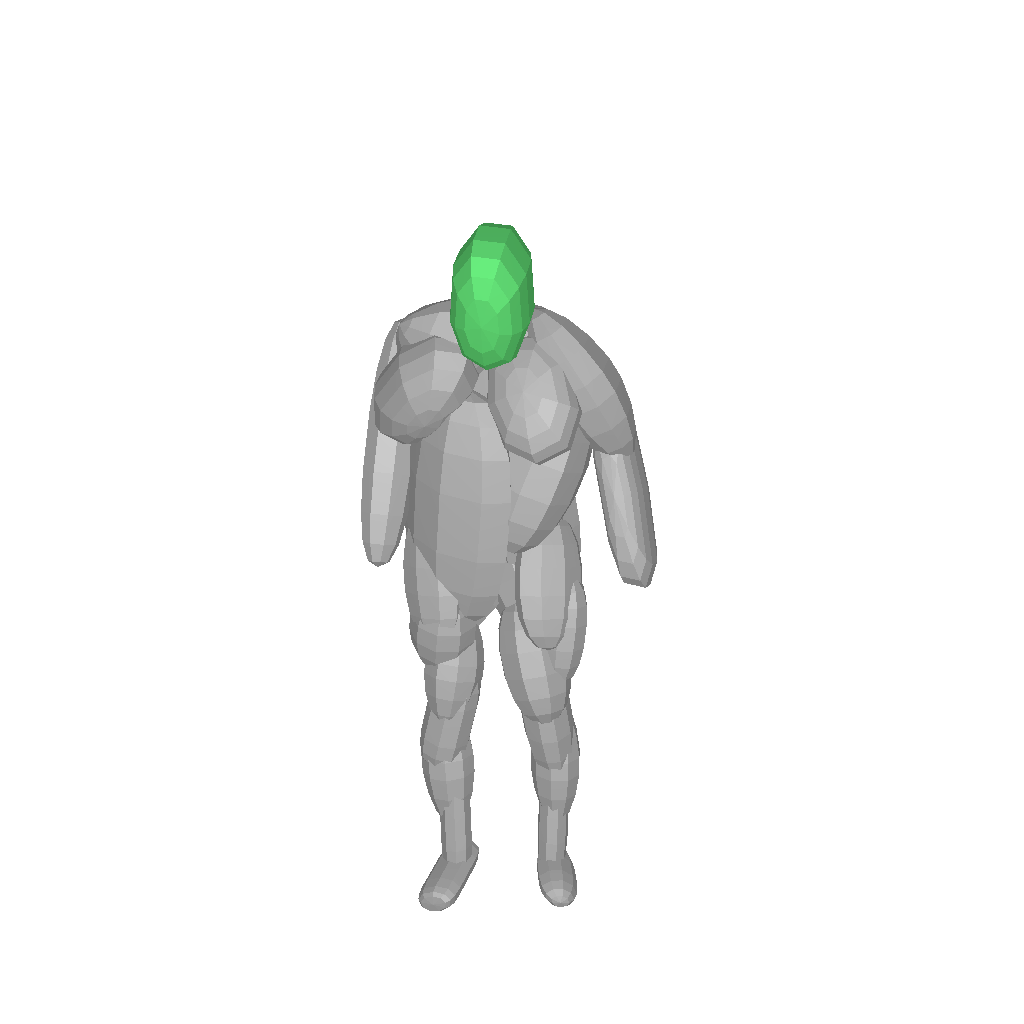}
    \end{subfigure}
    \vspace{-1.3em}
    \caption{Node $(4,12)$}
    \end{subfigure}
    \caption{\textbf{Semantic Predictions on D-FAUST.} To illustrate that our model indeed learns semantic hierarchical layouts of parts, here we color a specific node of the tree for various humans and we observe that it consistently corresponds to the same body part.}
    \label{fig:dfaust_supp_semantic_parts}

\end{figure}

\begin{figure}[h!]
	\centering
	\begin{subfigure}[b]{0.15\linewidth}
    	\centering
        \includegraphics[width=1.2\linewidth]{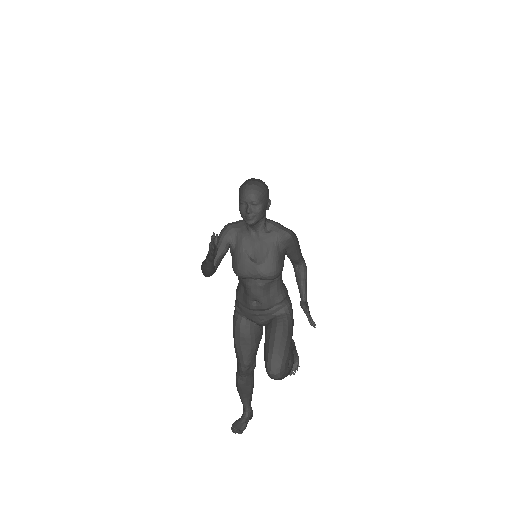}
    \end{subfigure}
    \hfill
    \begin{subfigure}[b]{0.15\linewidth}
    	\centering
        \includegraphics[width=\linewidth]{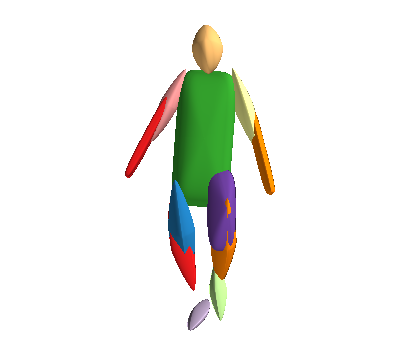}
    \end{subfigure}
    \hfill
    \begin{subfigure}[b]{0.15\linewidth}
    	\centering
        \includegraphics[width=\linewidth]{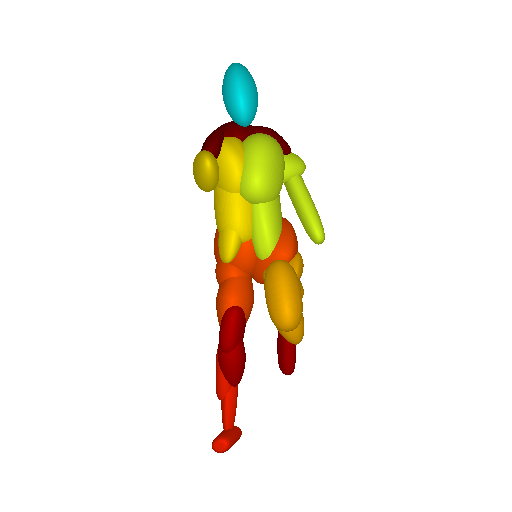}
    \end{subfigure}
    \hfill
    \begin{subfigure}[b]{0.15\linewidth}
    	\centering
        \includegraphics[width=1.2\linewidth]{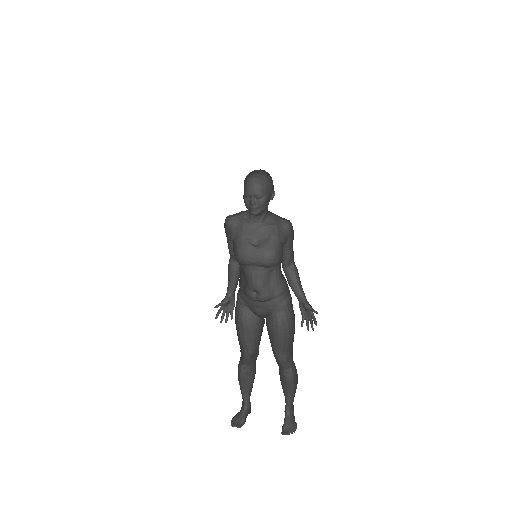}
    \end{subfigure}
    \hfill
    \begin{subfigure}[b]{0.15\linewidth}
    	\centering
        \includegraphics[width=1.0\linewidth]{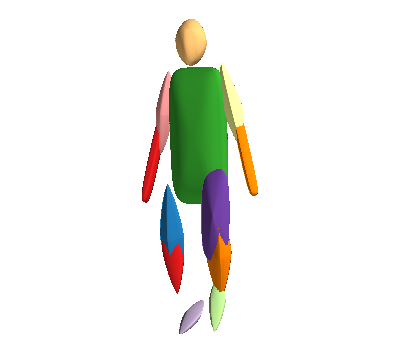}
    \end{subfigure}
    \hfill
    \begin{subfigure}[b]{0.15\linewidth}
    	\centering
        \includegraphics[width=\linewidth]{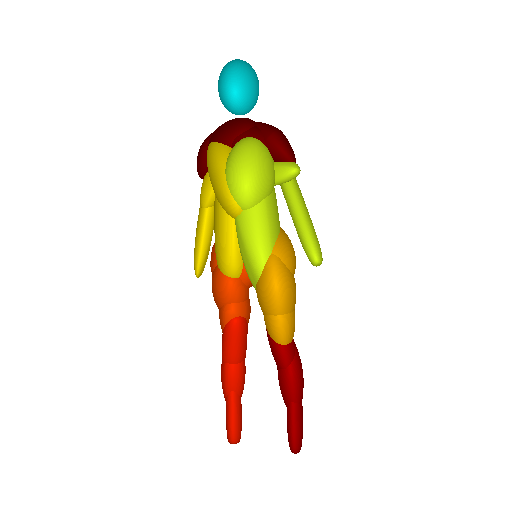}
    \end{subfigure}

    \vskip\baselineskip    \vspace{-1.2em}
    \begin{subfigure}[b]{0.15\linewidth}
		\centering
        \caption{\textbf{Input}}
    \end{subfigure}
    \hfill
    \begin{subfigure}[b]{0.15\linewidth}
		\centering
        \caption{\bf{SQs}\cite{Paschalidou2019CVPR}}
    \end{subfigure}
    \hfill
    \begin{subfigure}[b]{0.15\linewidth}
		\centering
        \caption{\textbf{Ours}}
    \end{subfigure}
    \hfill
    \begin{subfigure}[b]{0.15\linewidth}
		\centering
        \caption{\textbf{Input}}
    \end{subfigure}
    \hfill
    \begin{subfigure}[b]{0.15\linewidth}
		\centering
        \caption{\bf{SQs}\cite{Paschalidou2019CVPR}}
    \end{subfigure}
    \hfill
    \begin{subfigure}[b]{0.15\linewidth}
		\centering
        \caption{\textbf{Ours}}
    \end{subfigure}

    \vskip\baselineskip
    \begin{subfigure}[b]{0.49\linewidth}
    	\centering
        \includegraphics[width=\linewidth]{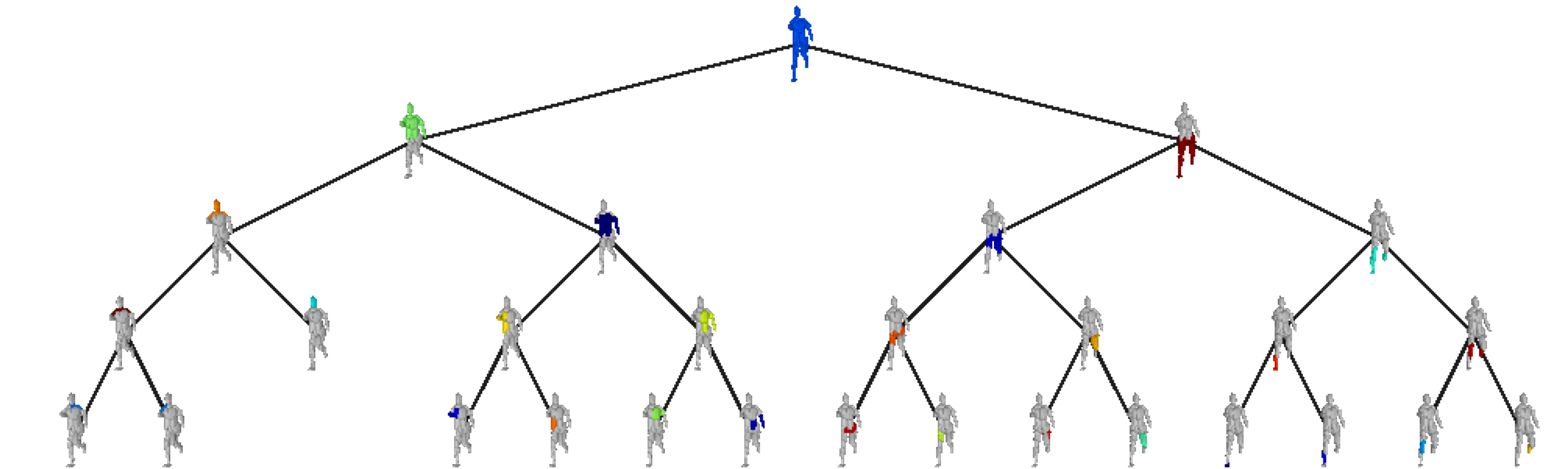}
    \end{subfigure}
    \hfill
    \begin{subfigure}[b]{0.49\linewidth}
    	\centering
        \includegraphics[width=\linewidth]{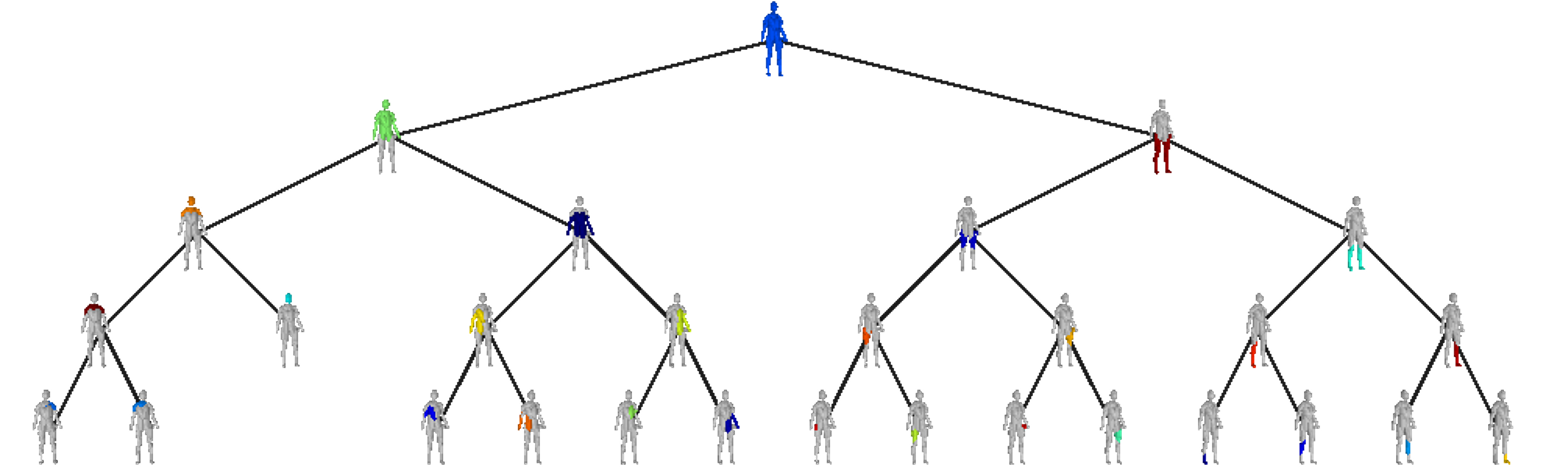}
    \end{subfigure}
    \vskip\baselineskip    \vspace{-1.2em}
    \begin{subfigure}[b]{0.49\linewidth}
		\centering
        \caption{\textbf{Predicted Hierarchy}}
    \end{subfigure}
    \hfill
    \begin{subfigure}[b]{0.49\linewidth}
		\centering
        \caption{\textbf{Predicted Hierarchy}}
    \end{subfigure}
    \vskip\baselineskip	\begin{subfigure}[b]{0.15\linewidth}
    	\centering
        \includegraphics[width=1.2\linewidth]{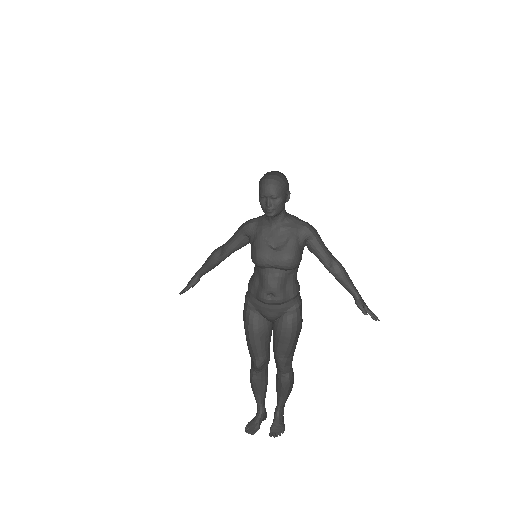}
    \end{subfigure}
    \hfill
    \begin{subfigure}[b]{0.15\linewidth}
    	\centering
        \includegraphics[width=\linewidth]{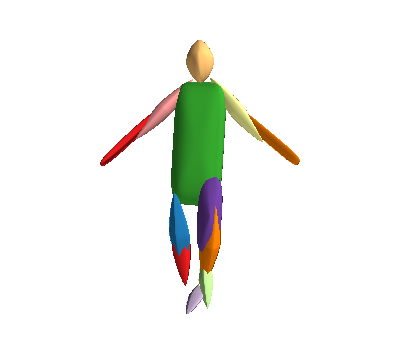}
    \end{subfigure}
    \hfill
    \begin{subfigure}[b]{0.15\linewidth}
    	\centering
        \includegraphics[width=\linewidth]{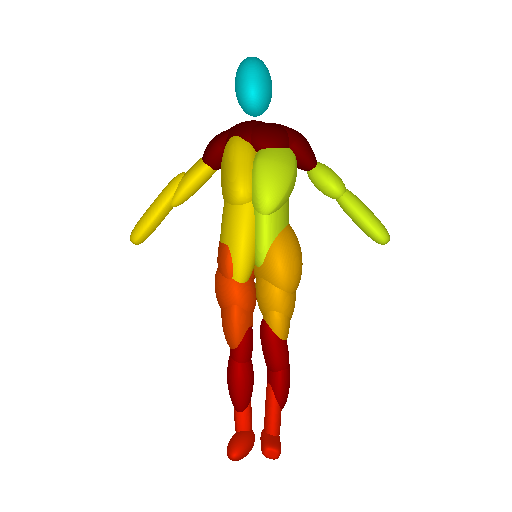}
    \end{subfigure}
    \hfill
    \begin{subfigure}[b]{0.15\linewidth}
    	\centering
        \includegraphics[width=1.2\linewidth]{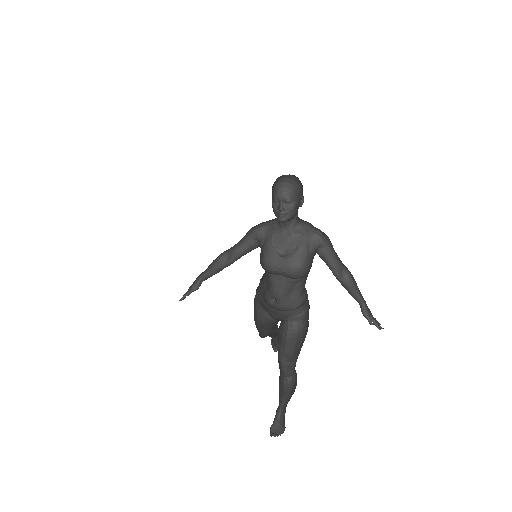}
    \end{subfigure}
    \hfill
    \begin{subfigure}[b]{0.15\linewidth}
    	\centering
        \includegraphics[width=1.0\linewidth]{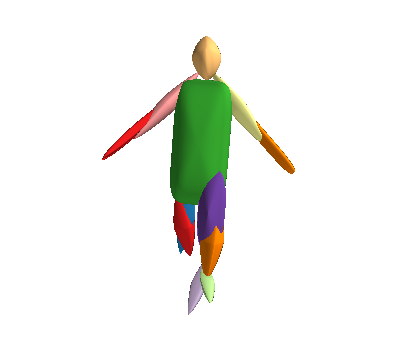}
    \end{subfigure}
    \hfill
    \begin{subfigure}[b]{0.15\linewidth}
    	\centering
        \includegraphics[width=\linewidth]{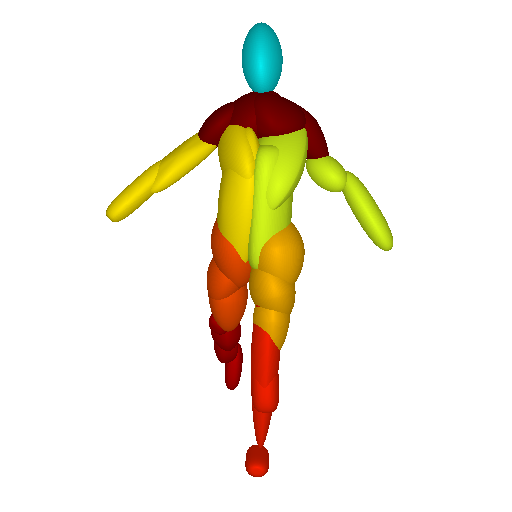}
    \end{subfigure}

    \vskip\baselineskip    \vspace{-1.2em}
    \begin{subfigure}[b]{0.15\linewidth}
		\centering
        \caption{\textbf{Input}}
    \end{subfigure}
    \hfill
    \begin{subfigure}[b]{0.15\linewidth}
		\centering
        \caption{\bf{SQs}\cite{Paschalidou2019CVPR}}
    \end{subfigure}
    \hfill
    \begin{subfigure}[b]{0.15\linewidth}
		\centering
        \caption{\textbf{Ours}}
    \end{subfigure}
    \hfill
    \begin{subfigure}[b]{0.15\linewidth}
		\centering
        \caption{\textbf{Input}}
    \end{subfigure}
    \hfill
    \begin{subfigure}[b]{0.15\linewidth}
		\centering
        \caption{\bf{SQs}\cite{Paschalidou2019CVPR}}
    \end{subfigure}
    \hfill
    \begin{subfigure}[b]{0.15\linewidth}
		\centering
        \caption{\textbf{Ours}}
    \end{subfigure}

    \vskip\baselineskip
    \begin{subfigure}[b]{0.49\linewidth}
    	\centering
        \includegraphics[width=\linewidth]{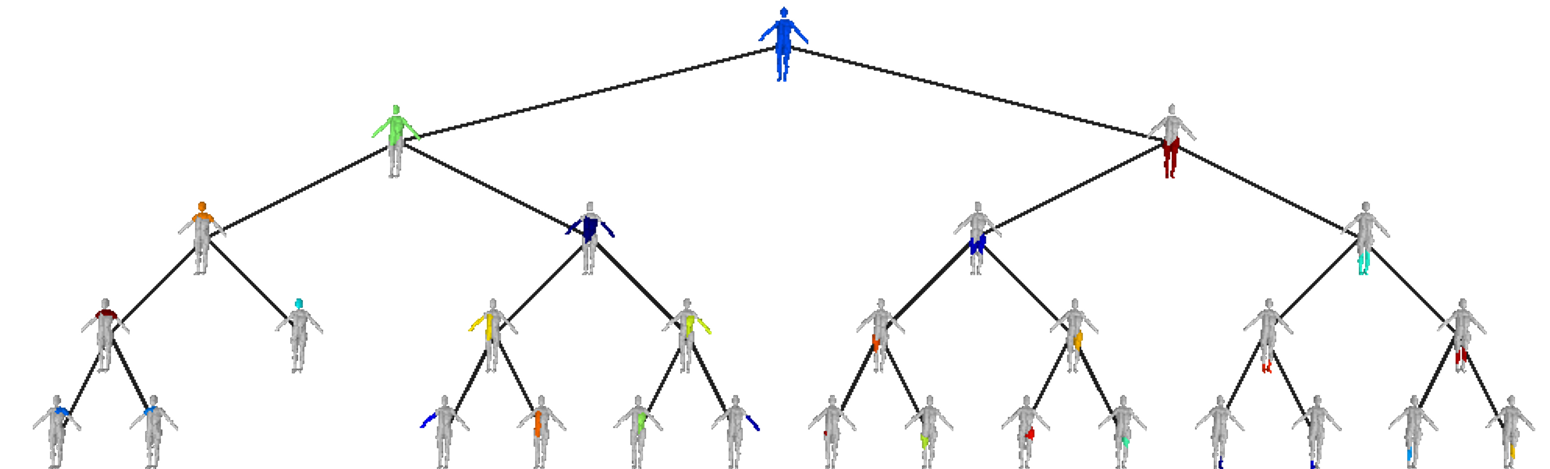}
    \end{subfigure}
    \hfill
    \begin{subfigure}[b]{0.49\linewidth}
    	\centering
        \includegraphics[width=\linewidth]{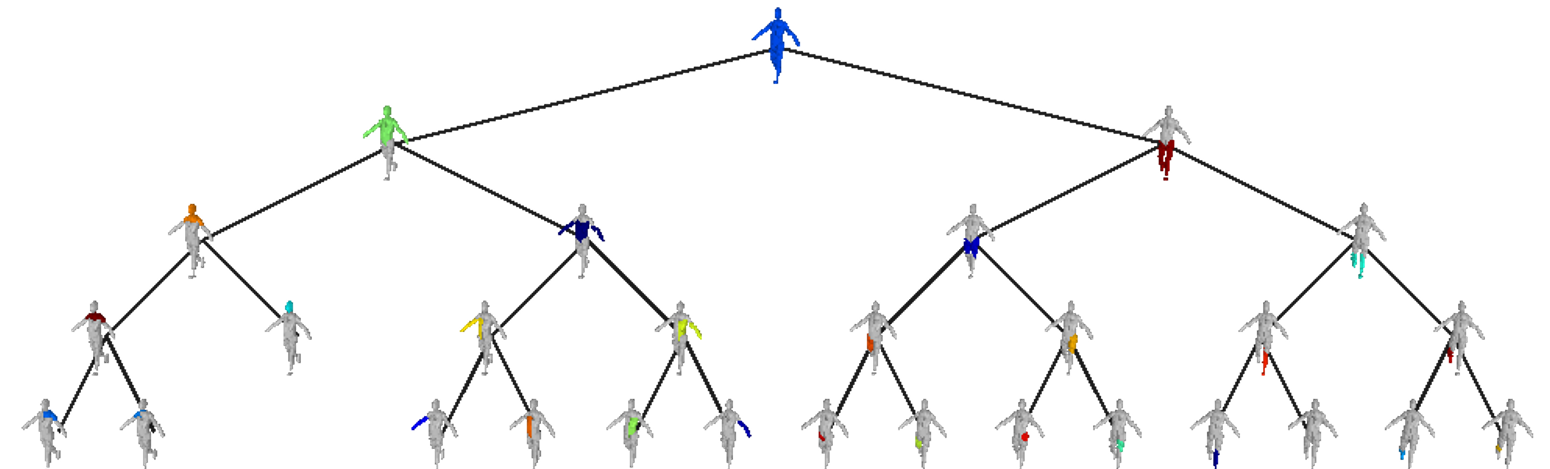}
    \end{subfigure}
    \vskip\baselineskip    \vspace{-1.2em}
    \begin{subfigure}[b]{0.49\linewidth}
		\centering
        \caption{\textbf{Predicted Hierarchy}}
    \end{subfigure}
    \hfill
    \begin{subfigure}[b]{0.49\linewidth}
		\centering
        \caption{\textbf{Predicted Hierarchy}}
    \end{subfigure}
    \vskip\baselineskip    \vspace{-1em}
    \caption{\textbf{Qualitative Results on D-FAUST.} Our network learns semantic mappings of body parts across different body shapes and articulations. Note that the network predicts less primitives for modelling the upper part of the human body.}
    \label{fig:dfaust_supp_2}
\end{figure}

In \figref{fig:dfaust_supp}$+$\ref{fig:dfaust_supp_2}, we qualitatively compare our predictions with \cite{Paschalidou2019CVPR}. We remark that even though \cite{Paschalidou2019CVPR} is more parsimonious, our predictions are more accurate. For example, we note that our shape reconstructions capture the details of the muscles of the legs that are not captured in \cite{Paschalidou2019CVPR}. For completeness, we also visualize the predicted hierarchy up to the fourth depth level. Another interesting aspect of our model, which is also observed in \cite{Paschalidou2019CVPR, Tulsiani2017CVPRa} is related to the semanticness of the learned hierarchies. We note that our model consistently uses the same node for representing the same part of the human body. For instance, node $(4, 15)$, namely the $15$-th node at the $4$-th depth level, consistently represents the right foot, whereas, node $(4, 12)$ represents the left foot. This is better illustrated in \figref{fig:dfaust_supp_semantic_parts}. In this figure, we only color the primitive associated with a particular node, for various humans, and we remark that the same primitive is used for representing the same body part.

Finally, another interesting characteristic of our model is related to its ability to use less primitives for reconstructing humans, with smaller bodies. In particular, while the lower part of the human body is consistently represented with the same set of primitives, the upper part can be represented with less depending on the size and the articulation of the human body.
This is illustrated in \figref{fig:dfaust_supp_2}, where we visualize the predictions of our model for such scenarios.

Below, we provide the full hierarchies of the results on D-FAUST from our main submission.

\begin{figure}
    \centering
    \includegraphics[width=1.0\linewidth]{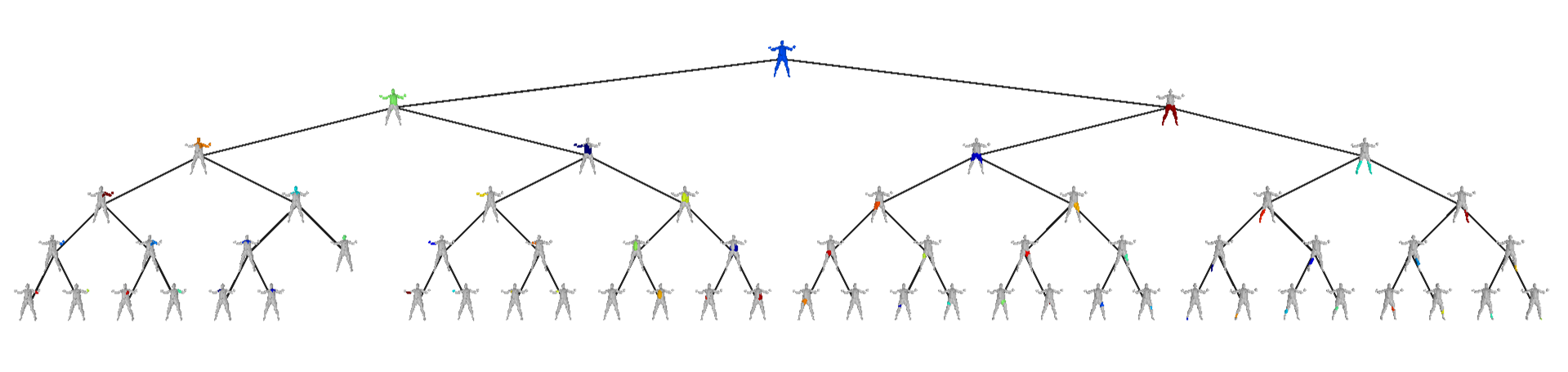}
    \includegraphics[width=1.0\linewidth]{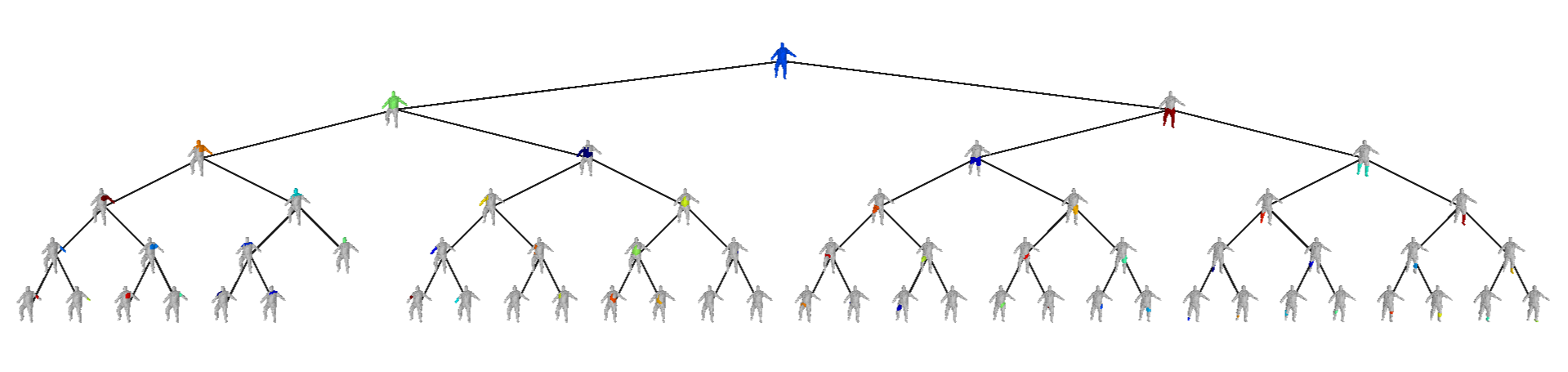}
    \caption{{\bf{Full hierarchies of Figure 6 in our main submission.}} Please zoom-in for details}
    \vspace{-1em}
\end{figure}

\begin{figure}
    \centering
    \includegraphics[width=1.0\linewidth]{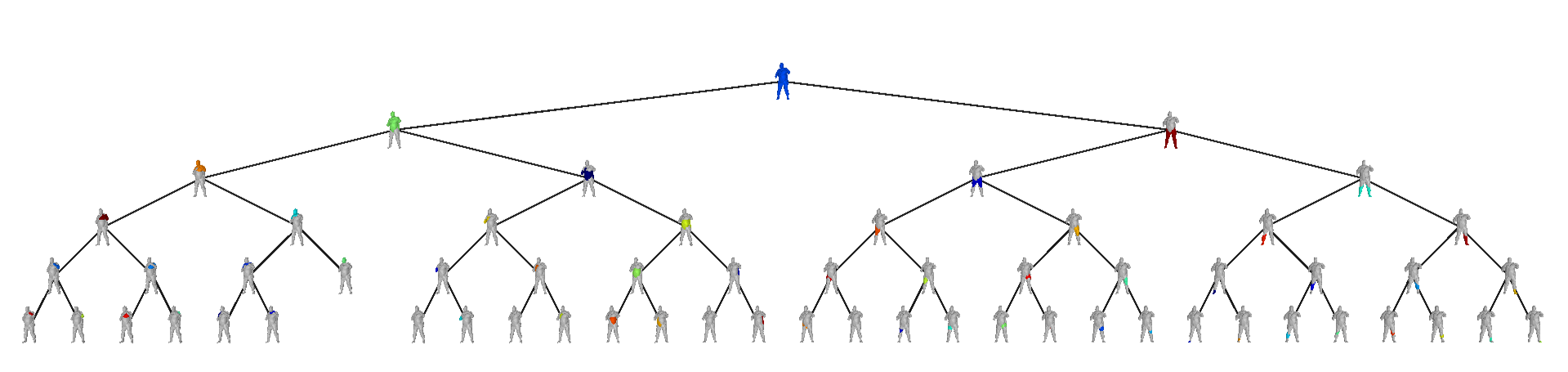}
    \includegraphics[width=1.0\linewidth]{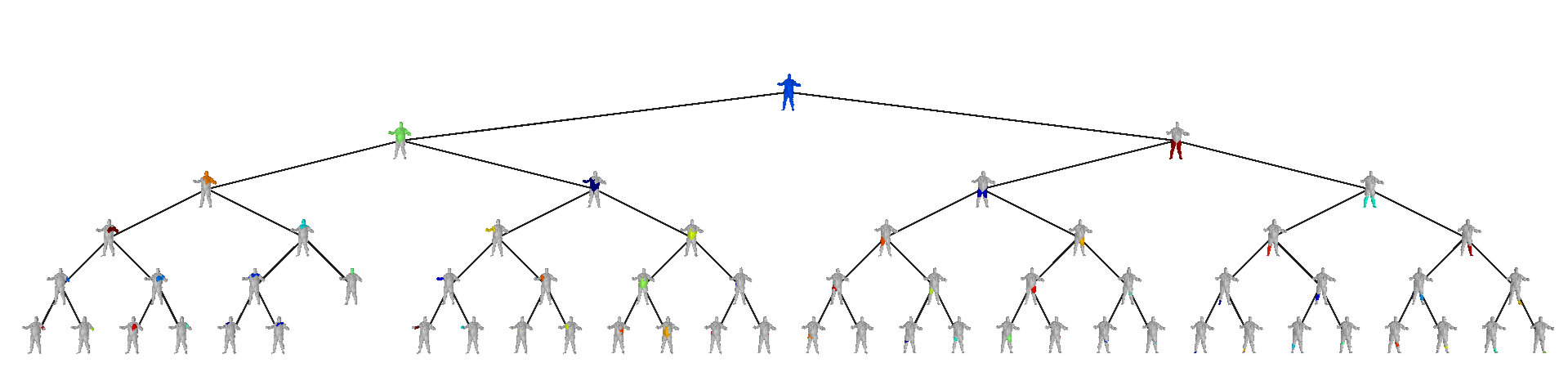}
    \includegraphics[width=1.0\linewidth]{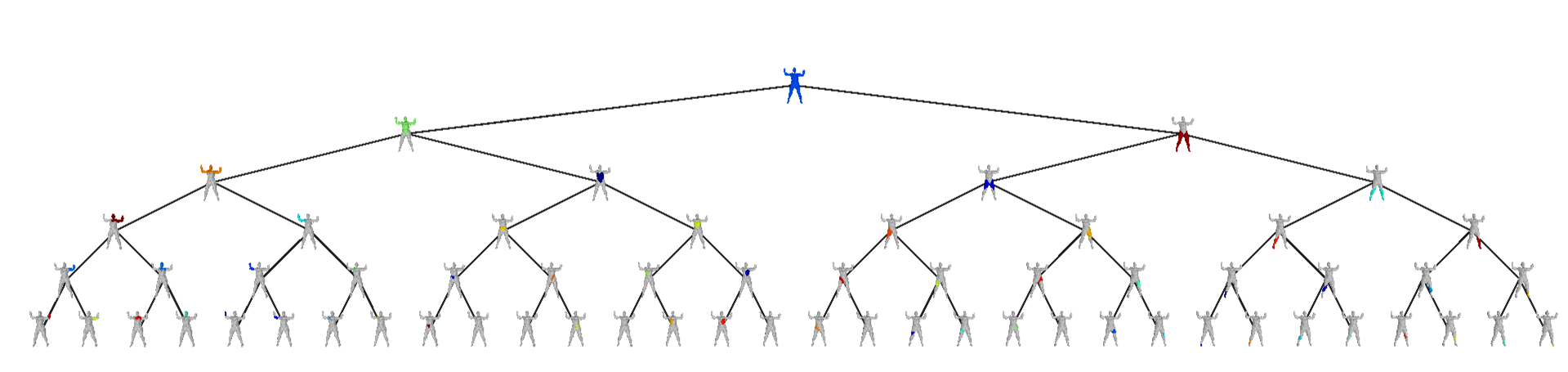}
    \includegraphics[width=1.0\linewidth]{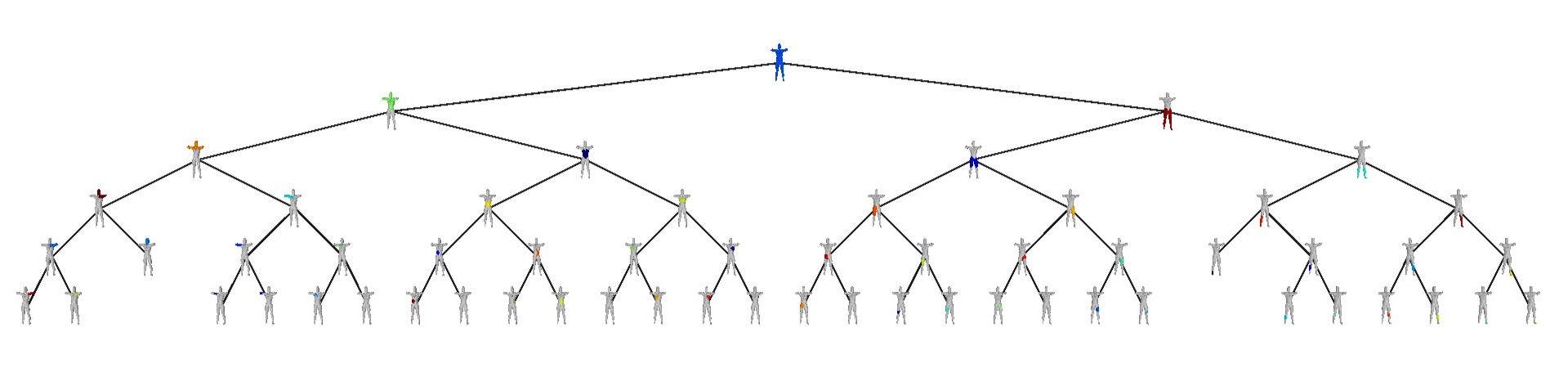}
    \caption{{\bf{Full hierarchies of Figure 8 in our main submission.}} Please zoom-in for details.}
    \vspace{-1em}
\end{figure}

\end{document}